\documentclass[twoside,11pt]{article}
\pdfoutput=1
\usepackage{jair, theapa, rawfonts}

\usepackage{balance} 
\usepackage{subfigure}
\usepackage{wrapfig}        
\usepackage{algorithm}
\usepackage{algorithmic}
\usepackage{amsmath, amsthm, paralist, amssymb}
\usepackage{graphicx}
\usepackage{booktabs}
\usepackage{xcolor}

\newtheorem{theorem}{Theorem}
\newtheorem{lemma}{Lemma}

\newtheorem{corollary}{Corollary}
\newtheorem{assumption}{Assumption}

\jairheading{80}{2024}{1-42}{04/2024}{09/2024}
\ShortHeadings{Understanding the Generalization Gap in Visual RL}
{Lyu, Wan, Li \& Lu}
\firstpageno{1}

\begin{document}

\title{Understanding What Affects the Generalization \\
 Gap in Visual Reinforcement Learning: \\ Theory and Empirical Evidence}

\author{\name Jiafei Lyu \email lvjf20@mails.tsinghua.edu.cn \\
        \addr Tsinghua Shenzhen International Graduate School\\
        Tsinghua University, Shenzhen, China
        \AND
       \name Le Wan \email vinowan@tencent.com \\
       \addr IEG, Tencent\\
       Shenzhen, China 
       \AND
       \name Xiu Li \email li.xiu@sz.tsinghua.edu.cn \\
       \addr (Corresponding Author) \\
       \addr Tsinghua Shenzhen International Graduate School \\
        Tsinghua University, Shenzhen, China
       \AND
       \name Zongqing Lu  \email zongqing.lu@pku.edu.cn \\
       \addr School of Computer Science, Peking University\\
       Beijing, China}


\maketitle

\begin{abstract}
Recently, there are many efforts attempting to learn useful policies for continuous control in visual reinforcement learning (RL). In this scenario, it is important to learn a \emph{generalizable} policy, as the testing environment may differ from the training environment, e.g., there exist distractors during deployment. Many practical algorithms are proposed to handle this problem. However, to the best of our knowledge, none of them provide a theoretical understanding of what affects the generalization gap and why their proposed methods work. In this paper, we bridge this issue by theoretically answering the key factors that contribute to the generalization gap when the testing environment has distractors. Our theories indicate that minimizing the representation distance between training and testing environments, which aligns with human intuition, is the most critical for the benefit of reducing the generalization gap. Our theoretical results are supported by the empirical evidence in the DMControl Generalization Benchmark (DMC-GB). 
\end{abstract}

\section{Introduction}

Visual reinforcement learning (RL) \shortcite{Nair2018VisualRL,yarats2022mastering} has aroused much attention from the community due to its success in handling complex control tasks solely from visual inputs. The advances in visual RL are promising for deploying RL algorithms in real-world applications since many physical devices (e.g., robotics) receive image observations. The progress in visual RL also has the potential to expedite the emergence of the large models in RL, or big decision models, since the visual inputs can be easily aligned with pre-processing and we only need to align the action space, whereas in state-based tasks, we need to handle the alignment of both state space and action space.

As a small step towards this goal, it is critical for the policies learned by the visual RL algorithms to be able to \emph{generalize} to unseen scenarios, like human beings. Unfortunately, it is a challenging problem at the current stage as the difference between the clean training environment and the unseen environments is not predictable. For example, we train an RL agent in simulated environments from image observations, while need to deploy it in real-world tasks, where the evaluated scene may quite differ from the trained ones (e.g., it may rain). Meanwhile, the image observations are complex and fragile to be attacked, e.g., noises may be included in the testing environment.

Existing methods remedy the mismatch between the training and testing environment by leveraging data augmentation \shortcite{Yarats2021ReinforcementLW,yarats2022mastering,Hansen2020GeneralizationIR,Raileanu2021AutomaticDA,Hansen2021StabilizingDQ}, self-supervision methods \shortcite{agarwal2021contrastive,zhang2021learning,hansen2021selfsupervised} or pre-trained image encoders \shortcite{Yuan2022PreTrainedIE,Ze2022VisualRL,dittadi2021role}, etc. Despite their success in achieving good performance during generalization, to the best of our knowledge, \emph{none of them actively explain why their methods work in practice from a theoretical perspective}. In this paper, we aim at bridging this gap by theoretically analyzing what affects the generalization gap. We believe the theoretical understanding of how to reduce the generalization gap is important for guiding the design of good and generalizable algorithms in the future, instead of blindly adding new modules to pray for a `lucky' trial.

We focus on the very specific generalization setting: the algorithm is trained in a clean environment with visual input, while deployed in an unseen testing environment with distractors. We note here that the distractor indicates that the observations in testing environment contain some diversions that may affect the performance of the agent, e.g., the color of the controlled agent or the background of the agent changes (the controlled agent remains unchanged). We also allow a slight difference of dynamics between training and testing environments. Nevertheless, it is very challenging to directly analyze the generalization gap between the training environment and the testing environment, since the policy keeps evolving during the training process. 

To tackle this challenge, we resort to \emph{reparameterization trick} to decouple the randomness in the environment from the evolving policy, the transition dynamics, and the initial state distribution. As a consequence, we present the reparameterizable visual RL framework. Under some mild assumptions, we establish concrete theoretical bounds on the generalization gap when deploying the visual policies in testing environments with distractors. Our results suggest that the most crucial factor that influences the test performance is the representation deviation before and after adding the distractor. Note that we focus on \emph{on-policy} RL (where the episodes are sampled using the current policy during training) instead of off-policy RL, since the data-collecting policy keeps evolving in the context of visual RL.


We also examine the rationality of the assumptions we made, and the theoretical conclusions we achieved by conducting experiments of different algorithms in DMControl Generalization Benchmark (DMC-GB) \shortcite{Hansen2020GeneralizationIR}. In this benchmark, the testing environments involve distractions like color changing or background changing of the controlled agent (no extra object/hindrance are included). It turns out that the empirical evidence is consistent with the theoretical insights.

\section{Related Work}
\label{sec:relatedwork}

\noindent \textbf{Visual reinforcement learning.} The success of learning visual representation in computer vision \shortcite{Vincent2008ExtractingAC,Doersch2015UnsupervisedVR,Kolesnikov2019RevisitingSV,Cole2021WhenDC,Kolesnikov2019BigT} has inspired the development of image-based RL. Many methods \shortcite{Hafner2018LearningLD,Majumdar2023WhereAW,Lee2019StochasticLA,Srinivas2020CURLCU,Schwarzer2020DataEfficientRL} have shown the benefits of auto-encoders \shortcite{Finn2015LearningVF,Tschannen2018RecentAI} to visual RL. The potential advantages of auxiliary tasks or objectives \shortcite{jaderberg2017reinforcement,Stooke2020DecouplingRL,Lin2019AdaptiveAT,Kulhnek2019VisionbasedNU} and data augmentations \shortcite{yarats2022mastering,Yarats2021ReinforcementLW,Laskin2020ReinforcementLW} are also widely explored. There are also some advances on explicit representation in visual RL like object-centric RL that aim at increasing the interpretability of decision-making \shortcite{Wu2023ReadAR,Delfosse2023InterpretableAE,Delfosse2024InterpretableCB,Luo2024INSIGHTEN,Zheng2022SymbolicVR}.

\noindent \textbf{Generalization in visual RL.} Generalization is a central challenge in RL \shortcite{Cobbe2018QuantifyingGI,Packer2018AssessingGI}, which is strongly correlated with overfitting issue \shortcite{Cobbe2018QuantifyingGI,Zhang2018ADO,Zhang2018ASO,Song2020Observational} (it is customary that one uses the same environment for both training and testing). In visual RL, it is vital for the policy learned from pixels to be able to generalize to unseen scenarios due to the variability of image observations in many real-world tasks. Domain randomization \shortcite{Chebotar2018ClosingTS,Peng2017SimtoRealTO,Pinto2017AsymmetricAC,Dai2019AnalysingDR,Polvara2020SimtoRealQL,Slaoui2019RobustDR,Yue2019DomainRA} and data augmentation \shortcite{Hansen2021StabilizingDQ,Hansen2020GeneralizationIR,Raileanu2021AutomaticDA,Fan2021SECANTSC,Lee2019NetworkRA,Mitrano2022DataAF,Huang2022SpectrumRM,Maksymets2021THDATH} have been proven to be effective in terms of generalizing to visually different scenes. Furthermore, utilizing self-supervision methods for improving generalization gains much interest \shortcite{agarwal2021contrastive,zhang2021learning,Hansen2020GeneralizationIR,hansen2021selfsupervised,Sun2019TestTimeTW,Wang2021UnsupervisedVA}. Some researchers attempt to improve the generalization capability of RL agent via symbolic rules or structures \shortcite{Zambaldi2018RelationalDR,Jiang2019NeuralLR,Cao2022GALOISBD} There are also many other interesting perspectives that encourage generalization of visual policies. Notably, \shortciteA{Yuan2022PreTrainedIE} leverage pre-trained image encoder for universal representations in the training and testing environments; \shortciteA{Hansen2021StabilizingDQ} uncover the pitfalls of data augmentation in deep $Q$-learning, and mitigate them by enforcing regularization between the augmented and unaugmented data; \shortciteA{Bertoin2022LookWY} mask the input image pixel-by-pixel based on the calculated attribution to extract and highlight the important regions (similar to noise handling methods like \shortcite{Grooten2023AutomaticNF}), and regularize the value function to ensure the consistency between the $Q$-values of input and masked images. \shortciteA{Li2023NormalizationEG} find that the generalization performance of the agent can be improved with some proper normalization techniques.

In spite of the huge success of the aforementioned methods in generalizing to previously unseen testing environments, as far as we can tell, none of them provide explanations of why their method improves generalization performance from a theoretical perspective. In this paper, we bridge this gap via providing theoretical bounds on the generalization gap in visual RL. Prior theoretical results of generalization in RL mainly lie in bandits \shortcite{Agarwal2014TamingTM,slivkins2011contextual,Jaksch2008NearoptimalRB,Li2022InstanceoptimalPA,Agrawal2012ThompsonSF}, and most of them analyze the regret, i.e., the deviation between the expected value and the optimal return. However, studies on the generalization gap in general RL settings are few. \shortciteA{lan22onthegeneralization} theoretically characterize how and when the state representation generalizes based on the notion of effective dimension. \shortciteA{Bertrn2020InstanceBG} treat the dynamics of the training level as instances and analyze the generalization bound of the value gap between the training and testing environments based on the training instances. The most relevant to our work is \shortcite{Wang2019OnTG}, which unpacks the generalization error of on-policy RL under reparameterization. However, their results do not apply to the existence of an encoder in the agent and distractors during testing. Our results also clearly anatomize how to reduce the generalization gap in visual RL.

\noindent \textbf{Reparameterization in RL.} The reparameterization trick has been widely adopted in deep neural networks, for the benefit of optimization efficiency \shortcite{Kingma2015VariationalDA,jang2017categorical,Maddison2016TheCD,Salimans2016WeightNA,Mostafa2019ParameterET,kingma2013auto}. A distribution can be reparameterized \shortcite{kingma2013auto} if (1) it has a tractable inverse CDF (cumulative distribution function); (2) it is a composition of reparameterizable distributions (e.g., Gaussian distribution); (3) we combine partial reparameterization with score function estimators \shortcite{Ruiz2016TheGR,Bauer2021GeneralizedDR}. In the context of RL, reparameterization trick is also very popular. For example, PGPE \shortcite{Sehnke2008PolicyGW} and SAC-style maximum entropy RL algorithms \shortcite{Haarnoja2018SoftAO,Haarnoja2018SoftAA} perform policy reparameterization; EPG \shortcite{ciosek2020expected} reparameterizes both the policy and the critic; SVG \shortcite{Heess2015LearningCC}, instead, reparameterizes the policy and the environmental transition probability. Other interesting attempts include reparameterizing the action space \shortcite{Wei2018HierarchicalAF,Bahl2020NeuralDP,yang2022trail}, the weight vectors in the neural network \shortcite{Salimans2016WeightNA}, a distribution over sample space via the learned quantile function \shortcite{Dabney2018ImplicitQN}, etc. In this work, we leverage the reparameterization trick for quantifying the generalization gap in visual RL. Despite that some prior work \cite{Heess2015LearningCC,Wang2019OnTG} explore the reparameterization trick in RL, as far as we can tell, we are the first that apply the reparameterization trick on the transition function in the context of visual RL (where distractors exist in the testing environment) to facilitate theoretical derivations of generalization bounds.

\section{Preliminaries}
\label{sec:preliminaries}
\textbf{Reinforcement Learning.} Reinforcement learning (RL) deals with sequential decision-making problems, and it can be specified by a Markov Decision Process (MDP) $\mathcal{M}=\{\mathcal{S},\mathcal{A}, r, p, p_0, \gamma\}$. $\mathcal{S}$ and $\mathcal{A}$ are state space and action space, respectively, $p(s,a,s^\prime):\mathcal{S}\times\mathcal{A}\times\mathcal{S}\mapsto [0,1]$ is the transition probability from $s$ to $s^\prime$ after taking action $a$, $p_0(s):\mathcal{S}\mapsto [0,1]$ is the initial state distribution, $r(s,a):\mathcal{S}\times\mathcal{A}\mapsto \mathbb{R}$ is the scalar reward function, and $\gamma\in [0,1)$ is the discount factor. The policy $\pi_\theta(s)$ parameterized by $\theta$ is a mapping from state space to action space. We denote the policy class as $\Pi$, then we have $\pi_\theta(s)\in\Pi$. The goal of reinforcement learning is to find a policy that can maximize the expected long-term discounted return $J(\theta) = \mathbb{E}_\pi[\sum_{t=0}^\infty \gamma^t r(s_t,a_t)]$.

Since in practice, it is usually infeasible for infinite interactions, we consider episodic MDPs with a finite horizon in this work. We denote the length of the episode (i.e., the horizon) as $T+1$ and the corresponding trajectory as $\tau$, i.e., $\tau = \{s_0, s_1,\ldots,s_T\}$. Denote the joint distribution of the trajectories in an episode $\tau=\{s_0,\ldots,s_T\}$ as $\mathcal{D}_{\pi,p,p_0}$, which is jointly determined by the transition probability $p$, initial state distribution $p_0$, and the learned policy $\pi$. For simplicity, we assume both $p$ and $p_0$ are fixed. Then, $\mathcal{D}_{\pi,p,p_0}$ degenerates into $\mathcal{D}_\pi$. Under finite horizon, our goal is $\max_{\pi\in\Pi} \mathbb{E}_{\tau\sim\mathcal{D}_\pi}[J(\tau;\theta)] = \mathbb{E}_{\tau\sim\mathcal{D}_\pi}[\sum_{t=0}^T\gamma^t r(s_t,\pi(s_t))]$.

\textbf{Generalization Gap.} Generalization is a crucial problem and has been widely studied in the context of supervised learning \shortcite{Motiian2017UnifiedDS,Li2018DomainGW,Zhou2021DomainGA,shalev2014understanding}. In supervised learning, we usually have access to a dataset $\mathcal{D}$ containing i.i.d. (independent and identically distributed) samples $\{(x_i,y_i)\}_{i=1}^n$. The generalization gap in this scenario is $\| \mathbb{E}[l(k,x,y)] - \frac{1}{n}\sum_{i=1}^n l(k,x_i,y_i) \|_2^2$, where $k$ is the learned prediction function, $l$ is the user-defined loss function. The generalization gap here measures the deviation between the expected loss and the empirical average loss. Following a similar formulation, the generalization gap in reinforcement learning can be defined as: $\| \mathbb{E}_{\tau\sim\mathcal{D}^\prime_{\hat{\pi}}}[J(\tau)] - \frac{1}{n}\sum_{i=1}^n J(\tau_i) \|_2^2$, where $\mathcal{D}^\prime_{\hat{\pi}}$ is the state sequence distribution in the testing environment, $\hat{\pi}=\arg\max_{\pi\in\Pi,\tau_i\in\mathcal{D}_{\pi}}\frac{1}{n}\sum_{i=1}^n J(\tau_i)$, $n$ is the number of training episodes. It is difficult to quantify the generalization gap in RL due to the fact that the underlying sample distribution in the training environment $\mathcal{D}_\pi$ changes as the policy evolves, whereas the sample distribution in the supervised learning is kept fixed. Meanwhile, $\mathcal{D}^\prime_{\hat{\pi}}$ may differ from the sample distribution in the training environment since there may exist a distribution shift of the transition dynamics and initial state distribution between the training and testing environments.

\textbf{Visual RL Setting.} In this paper, we consider visual RL where the agent receives image observations and executes actions based on them. It is generally formulated by a Partial Observable Markov Decision Process (POMDP) $\mathcal{M}=\{\mathcal{S},\mathcal{O},\mathcal{A}, r, p, p_0, \gamma\}$, where $\mathcal{O}$ is the observation space. We assume direct access to $\mathcal{S}$, while our analysis also applies when involving $\mathcal{O}$. This is valid since under the setting we consider, generalizing to environments with distractors, $\mathcal{O}=\mathcal{S}$. We denote the visual input at timestep $t$ as $s_t$, which can be high-dimensional. In visual RL, it is a common practice that we adopt an encoder $\phi(\cdot):\mathcal{S}\mapsto\Phi$ for extracting knowledge from complex image inputs, $\Phi$ is the representation space. The underlying RL algorithm accepts the representations outputted by the encoder and optimizes itself accordingly. The reward function gives $r(s,\pi(\phi(s)))$ and the policy is $\pi(\phi(s))$. Therefore, the randomness in the policy is deeply influenced by the encoder. During testing, we assume there exists the distractor $f(\cdot)\in \mathcal{F}$ that transforms the vanilla image observation into a new image, where $\mathcal{F}$ is the function class that contains all possible distractors during deployment. We name $f(\cdot)$ as the \emph{transpose function} (or \emph{distraction function}), which can take an arbitrary form and can even have no explicit expression. However, we do not allow it to modify the physical structure of the target agent, i.e., the color, camera pose, or background of the agent can be changed. After adding the distractor, the state space is augmented, and we assume that the reward function and the policy are well-defined in this augmented state space. This can also be formulated as the following setting: we have $N$ MDPs, $\{\mathcal{M}_1, \mathcal{M}_2,\ldots,\mathcal{M}_N\}$ that share the same structure, while we only have access to $\mathcal{M}_1$ during training and aim at learning a policy that can achieve good performance in other MDPs in a zero-shot manner. Without loss of generality, we present the detailed objective function in the testing and training environment below,
\begin{equation*}
    {\rm Testing\, environment\, objective:}\qquad \mathbb{E}_{\tau\sim\mathcal{D}^\prime_{\hat{\pi}}}[J(\overbrace{\phi(\underbrace{f(\tau)}_{\rm transposed\, trajectory})}^{\rm representation})],
\end{equation*}
\begin{equation*}
    {\rm Training\, environment\, objective:}\qquad \mathbb{E}_{\tau\sim\mathcal{D}_{\pi}}[J(\overbrace{\phi(\tau)}^{\rm representation})].
\end{equation*}

Regardless of whether there exist distractors during deployment, the reward is decided by the action at state $s$, hence the objective function in the testing environment can be expressed as
\begin{equation}
    \mathbb{E}_{\tau\sim\mathcal{D}^\prime_{\hat{\pi}}}[J(\phi(f(\tau)))]=\mathbb{E}_{\tau\sim\mathcal{D}^\prime_{\hat{\pi}}}\left[\sum_{t=0}^T\gamma^t r(s_t, \pi(\phi(f(s_t))))\right],
\end{equation}
and similarly, we have the objective function in the training environment,
\begin{equation}
    \mathbb{E}_{\tau\sim\mathcal{D}_{\hat{\pi}}}[J(\phi(\tau))]=\mathbb{E}_{\tau\sim\mathcal{D}_{\hat{\pi}}}\left[\sum_{t=0}^T\gamma^t r(s_t, \pi(\phi(s_t)))\right].
\end{equation}
Throughout this paper, we denote $\|\cdot\|$ as the $L_2$-norm.

\section{Reparameterizable Visual RL}

\begin{figure}
    \centering
    \includegraphics[width=0.5\linewidth]{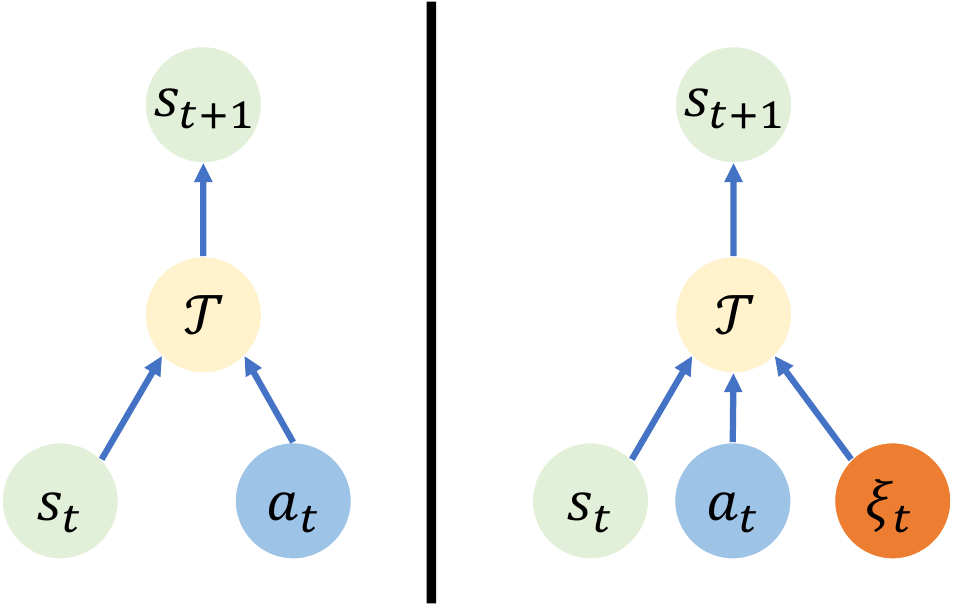}
    \caption{Comparison of state transition before (left) and after (right) reparameterization.}
    \label{fig:comparison}
\end{figure}

As we discuss in Section \ref{sec:preliminaries}, it is hard to quantify the generalization gap in RL since the policy can evolve and different trajectories can be gathered in $\mathcal{D}_\pi$, resulting in a changing sample distribution $\mathcal{D}_\pi$. \textbf{It incurs challenges when we directly apply the theoretical results from supervised learning} because the theoretical results from supervised learning require a fixed sample distribution (the loss function can change) while in RL, the reward function is fixed but the sample distribution is not. It then necessities to decouple the randomness of the policy from the expected return. We resort to the reparameterization tool to address this issue. The goal of reparameterization is to find a way to recast a statistical expression in a different way while preserving its meaning. By reparameterizing the transition dynamics, we can represent the randomness of the sample distribution with a random variable. We show in Figure \ref{fig:comparison} the comparison of how state transfers before and after reparameterization. We assume that both the transition dynamics and the state initialization process can be reparameterized, then by using the reparameterization trick \shortcite{jang2017categorical,Kingma2015VariationalDA,Figurnov2018ImplicitRG,Xu2018VarianceRP}, we can rewrite the objective function in the training environment as follows:
\begin{equation}
    \label{eq:reparameterizationrl}
    \mathbb{E}_{\tau\sim\mathcal{D}_\pi}[J(\phi(\tau))] = \mathbb{E}_{\xi\sim q(\xi)}\left[ J(\phi(\tau(g(\xi;\pi_\theta)))) \right],
\end{equation}
where $g$ is a deterministic function of parameter $\theta$, $q(\xi)$ is the distribution of the random variable $\xi$. Since we do not care much about the form of $g$, we \emph{absorb} $g$ into $\tau$ (as SVG \shortcite{Heess2015LearningCC} does), yielding $\tau(\xi;\pi_\theta)$. This is valid because $g$ shares the parameter with the policy, and the underlying meaning of $s_t$ does not change by doing so. Observing Equation \ref{eq:reparameterizationrl}, we find that now the objective function no longer depends on the sample distribution $\mathcal{D}_\pi$ and the policy $\pi$. That is, we isolate the randomness of the policy $\pi$ from the expected return, and the policy $\pi$ can now only affect the reward signal $r(s_t,\pi(\phi(s_t)))$ through the representation $\phi(s_t)$. Note that the state is transitioned with the reparametrized transition dynamics $\mathcal{T}$ (if the state is transitioned with the original dynamics 
function, then the resulting distribution $\mathcal{D}_\pi$ still relies on the policy). The introduced random variable $\xi$ can be sampled before the start of the episode and the distribution $q(\xi)$ does not evolve. Then, we could leverage the conventional generalization theory to derive the generalization gap.

It is interesting to note here that if we assume the states in the MDP are discrete and $p_0(s), p(s,a)$ are multinomial distributions, then we can reparameterize it with Gumbel distribution \shortcite{Gumbel1954StatisticalTO}, which are extensively adopted and studied in \shortcite{jang2017categorical,Maddison2016TheCD,Potapczynski2019InvertibleGR,Huijben2021ARO,Lorberbom2021LearningGG,Joo2020GeneralizedGG}. Denote $\mathcal{G}^{|\mathcal{S}|}$ as the $|\mathcal{S}|$-dimensional standard Gumbel distribution, then Gumbel random variable $\xi_0,\xi_1,\ldots,\xi_T$ can be sampled from $\mathcal{G}^{|\mathcal{S}|}$ (i.e., $\xi_i\sim \exp(-\xi - \exp(-\xi))$), and the sampling procedure in the MDP can be acquired by utilizing the Gumbel-max trick \shortcite{Huijben2021ARO,Wang2019OnTG,jang2017categorical,Oberst2019CounterfactualOE}:
\begin{equation}
    \label{eq:gumbelmax}
    s_{t+1}=\arg\max(\xi_t + \log p(s_t,\pi(s_t))), t = 0,1,\ldots,T.
\end{equation}
There is also an interesting connection between $\arg\max$ operator and Gumbel-softmax as illustrated in \shortcite{jang2017categorical,Maddison2016TheCD,Gumbel1954StatisticalTO}, if some relaxations are adopted for the state space (e.g., one-hot).

\begin{algorithm}[H]
    \centering
    \caption{Reparameterizable Visual RL}\label{alg:reparameterizablerl}
    \begin{algorithmic}[1]
        \STATE Sample $\xi_0,\xi_1,\ldots,\xi_T$
        \STATE Get $s_0=\mathcal{I}(\xi_0)$
        \STATE Initialize $R=0$
        \STATE Set encoder $\phi(\cdot)$, policy $\pi(\cdot)$
        \FOR{$t$ = 0 to $T$}
        \STATE $R=R+\gamma^t r(s_t,\pi(\phi(s_t)))$
        \STATE $s_{t+1} = \mathcal{T}(s_t, \pi(\phi(s_t)), \xi_t)$
        \ENDFOR
    \end{algorithmic}
\end{algorithm}

We denote $\mathcal{T}(s,\pi(s))=p(s,\pi(s),s^\prime)$ as the state transition probability of the system, and $\mathcal{I}:\Xi\mapsto\mathcal{S}$ is the initialization function, where $\Xi$ is the space of the random variable $\xi$s. We then formally present the pseudo code of reparameterizable visual RL in Algorithm \ref{alg:reparameterizablerl}, where we reparameterize the \emph{transition dynamics} of the system to fulfill Equation \ref{eq:reparameterizationrl}. Notably, we do not require the dimension of $\xi$s to be the same as state $s$ in Algorithm \ref{alg:reparameterizablerl}, and the random variables $\xi_0,\xi_1,\ldots,\xi_T$ can be sampled from different distributions. The initialization function takes $\xi_0$ as input to produce $s_0$, and the transition probability $\mathcal{T}(s_t,\pi(\phi(s_t)),\xi_t)$ takes the current state $s_t$, the action induced by the policy $\pi(\phi(s_t))$, and $\xi_t$ to produce next state $s_{t+1}, \forall\, t$. 

Note that the random variables $\xi_0,\xi_1,\ldots,\xi_T$ can be drawn from some distributions before the episode starts, hence isolating the randomness of the policy. The above formulation also applies when the policy evolves during training, since the trajectory can be decided deterministically by executing reparameterized transition function $\mathcal{T}(s_t,\pi(\phi(s_t)),\xi_t)$ repeatedly. Intuitively, the randomness from the encoder is also decoupled, and its influence can be included in that of the policy. Importantly, we do not require the encoder parameters to be fixed across the budget of $n$ episodes.


\section{Theoretical Analysis on the Generalization Error}
\label{sec:theory}
In this section, we formally present our theoretical results on the generalization gap in visual RL, with the aid of the reparameterization tool. First, we impose some Lipschitz assumptions on the transition probability, the policy, and the reward function, which are fundamental for the following analysis. Most of these assumptions are reasonable and can be satisfied in practice. We discuss the rationality of these assumptions in Section \ref{sec:rationality}.

\begin{assumption}
\label{ass:trans}
The transition dynamics $\mathcal{T}(s,a,\xi):\mathcal{S}\times\mathcal{A}\times\Xi\mapsto \mathcal{S}$ is $L_{t_1}$-Lipschitz in terms of state $s$ and $L_{t_2}$-Lipschitz in terms of action $a$, i.e., $\forall s,s^\prime,a,a^\prime,\xi$,
\begin{align*}
    &\|\mathcal{T}(s,a,\xi) - \mathcal{T}(s^\prime,a,\xi)\|\le L_{t_1}\|s - s^\prime\|, \\
    &\|\mathcal{T}(s,a,\xi) - \mathcal{T}(s,a^\prime,\xi)\|\le L_{t_2}\|a - a^\prime\|.
\end{align*}
\end{assumption}

\begin{assumption}
\label{ass:policy}
    The policy $\pi(\phi;\theta)$ is $L_{\pi_1}$-Lipschitz in terms of the variable $\phi$ and $L_{\pi_2}$-Lipschitz in terms of the parameter $\theta$, i.e., $\forall \phi,\phi^\prime,\theta,\theta^\prime$,
    \begin{align*}
        &\|\pi(\phi;\theta) - \pi(\phi^\prime;\theta)\|\le L_{\pi_1}\|\phi - \phi^\prime\| \\
        &\|\pi(\phi;\theta) - \pi(\phi;\theta^\prime)\|\le L_{\pi_2}\|\theta - \theta^\prime\|.
    \end{align*}
\end{assumption}

\begin{assumption}
\label{ass:rew}
    The reward function $r(s,a)$ is bounded, i.e., $\forall s, a$, $|r(s,a)|\le r_{\rm max}$, and is $L_{r_1}$-Lipschitz in terms of state $s$, $L_{r_2}$-Lipschitz in terms of action $a$, i.e., $\forall s,s^\prime$ and $\forall\, a,a^\prime$,
    \begin{align*}
        &|r(s,a)-r(s^\prime,a)|\le L_{r_1} \|s-s^\prime\|,\\
        &|r(s,a)-r(s,a^\prime)|\le L_{r_2} \|a-a^\prime\|
    \end{align*}
\end{assumption}

Based on these assumptions, we first can derive how the output of the policy changes using different inputs and parameters.

\begin{lemma}[Policy deviation]
\label{lemma:policydev}
Assume that Assumption \ref{ass:policy} hold, denote $\phi(\cdot)$ as the encoder. Then under the transition $\mathcal{T}(s,a,\xi)$, at step $t\in\{0,1,\ldots,T\}$ in an episode of length $T+1$, we have
\begin{equation}
\begin{gathered}
    \|\pi(\phi(s_t^\prime);\theta^\prime) - \pi(\phi(s_t);\theta)\| \le L_{\pi_1}\|\phi(s_t^\prime) - \phi(s_t)\|  + L_{\pi_2}\|\theta^\prime - \theta\|, \forall s_t,s_t^\prime\in\mathcal{S}, \forall \theta,\theta^\prime.
\end{gathered}    
\end{equation}
\end{lemma}

\begin{proof}
    This lemma is a direct product of Assumption \ref{ass:policy}.
    \begin{align*}
        &\|\pi(\phi(s_t^\prime);\theta^\prime) - \pi(\phi(s_t);\theta)\| \\
        &= \| \pi(\phi(s_t^\prime);\theta^\prime) - \pi(\phi(s_t^\prime);\theta) + \pi(\phi(s_t^\prime);\theta) - \pi(\phi(s_t);\theta) \| \\
        &\le \|\pi(\phi(s_t^\prime);\theta^\prime) - \pi(\phi(s_t^\prime);\theta)\| + \|\pi(\phi(s_t^\prime);\theta) - \pi(\phi(s_t);\theta)\| \\
        &\le L_{\pi_1}\|\phi(s_t^\prime) - \phi(s_t)\| + L_{\pi_2}\|\theta - \theta^\prime\|.
    \end{align*}
\end{proof}

Note that the triangle inequality is used in this proof, and it is also widely used for the following proofs. Furthermore, we can derive the relationship between the deviations of two states at timestep $t$ and timestep $t-1$ as depicted in the following lemma.

\begin{lemma}[State deviation]
\label{lemma:statedev}
Assume that Assumption \ref{ass:trans} and \ref{ass:policy} hold, denote $\phi(\cdot)$ as the encoder, then under the transition $\mathcal{T}(s,a,\xi)$ and the policy $\pi(\cdot;\theta)$, at step $t$ in an episode of length $T+1$, we have $\forall s_t,s_t^\prime\in\mathcal{S}, t\in\{1,\ldots,T\}$,
\begin{equation}
    \|s_t - s_t^\prime\| \le L_{t_1}\|s_{t-1}^\prime - s_{t-1}\| + L_{t_2}L_{\pi_1}\|\phi(s_{t-1}^\prime) - \phi(s_{t-1})\|.
\end{equation}
\end{lemma}

\begin{proof}
    This lemma is also quite straightforward. By using the reparameterized visual RL framework, we have
    \begin{align*}
        \|s_t - s_t^\prime\|
        &= \|\mathcal{T}(s_{t-1},\pi(\phi(s_{t-1});\theta), \xi_{t-1}) - \mathcal{T}(s_{t-1}^\prime,\pi(\phi(s_{t-1}^\prime);\theta), \xi_{t-1})\| \\
        &= \|\mathcal{T}(s_{t-1},\pi(\phi(s_{t-1});\theta), \xi_{t-1}) - \mathcal{T}(s_{t-1}^\prime,\pi(\phi(s_{t-1});\theta), \xi_{t-1}) \\
        & \qquad \quad + \mathcal{T}(s_{t-1}^\prime,\pi(\phi(s_{t-1});\theta), \xi_{t-1}) - \mathcal{T}(s_{t-1}^\prime,\pi(\phi(s_{t-1}^\prime);\theta), \xi_{t-1})\| \\
        &\le \|\mathcal{T}(s_{t-1},\pi(\phi(s_{t-1});\theta), \xi_{t-1}) - \mathcal{T}(s_{t-1}^\prime,\pi(\phi(s_{t-1});\theta), \xi_{t-1})\| \\
        & \qquad \quad + \|\mathcal{T}(s_{t-1}^\prime,\pi(\phi(s_{t-1});\theta), \xi_{t-1}) - \mathcal{T}(s_{t-1}^\prime,\pi(\phi(s_{t-1}^\prime);\theta), \xi_{t-1})\| \\
        &\le L_{t_1}\|s_{t-1}-s_{t-1}^\prime\| + L_{t_2}\|\pi(\phi(s_{t-1});\theta) - \pi(\phi(s_{t-1}^\prime);\theta)\| \\
        &\le L_{t_1}\|s_{t-1}-s_{t-1}^\prime\| + L_{t_2}L_{\pi_1}\|\phi(s_{t-1}) - \phi(s_{t-1}^\prime)\|,
    \end{align*}
    where we use the Lemma \ref{lemma:policydev} by setting $\theta = \theta^\prime$.
\end{proof}

Similarly, we can derive how reward function changes with different input states. The resulting lemma can be found below.

\begin{lemma}[Reward deviation]
    \label{lemma:rewdev}
    Assume that Assumption \ref{ass:policy}, \ref{ass:rew} hold, denote $\phi(\cdot)$ as the encoder, then under the transition $\mathcal{T}(s,a,\xi)$ and the policy $\pi(\cdot;\theta)$, at step $t$ in an episode of length $T+1$, we have $\forall s_t,s_t^\prime\in\mathcal{S}$,
    \begin{equation}
    \begin{gathered}
        |r(s_t,\pi(\phi(s_t))) - r(s_t^\prime,\pi(\phi(s_t^\prime)))| \le L_{r_1} \|s_t-s_t^\prime\|  + L_{r_2} L_{\pi_1}\|\phi(s_t^\prime) - \phi(s_t)\|.
    \end{gathered}
    \end{equation}
\end{lemma}

\begin{proof}
    It is easy to find that,
    \begin{align*}
        &|r(s_t,\pi(\phi(s_t))) - r(s_t^\prime, \pi(\phi(s_t^\prime)))| \\
        &= |r(s_t,\pi(\phi(s_t))) - r(s_t^\prime,\pi(\phi(s_t))) +  r(s_t^\prime,\pi(\phi(s_t))) - r(s_t^\prime, \pi(\phi(s_t^\prime)))| \\
        &\le |r(s_t,\pi(\phi(s_t))) - r(s_t^\prime,\pi(\phi(s_t)))| + |r(s_t^\prime,\pi(\phi(s_t))) - r(s_t^\prime, \pi(\phi(s_t^\prime)))| \\
        &\le L_{r_1}\|s_t - s_t^\prime\| + L_{r_2} \|\pi(\phi(s_t)) - \pi(\phi(s_t^\prime))\| \\
        &\le L_{r_1}\|s_t - s_t^\prime\| + L_{r_2} L_{\pi_1} \|\phi(s_t) - \phi(s_t^\prime)\|,
    \end{align*}
    where we also use the Lemma \ref{lemma:policydev} by setting $\theta = \theta^\prime$.
\end{proof}

Lemma \ref{lemma:policydev}, \ref{lemma:statedev} and \ref{lemma:rewdev} indicate that \textit{representation matters} in reparameterized visual RL, as state deviation, policy deviation, and reward deviation are all highly correlated with the representation deviation. For example, if $s_t$ is close to $s_t^\prime$, their representation distance also ought to be small such that the agent executes similar actions upon them. These lemmas pave the way for Theorem \ref{theo:fixedpolicy}, where we show that given a (fixed) policy $\pi$, and the trajectories $\tau(\xi)$, the performance difference between adding the distractor (i.e., the transpose function) or without the distractor is determined by the difference between the transposed state $f(s_t)$ and the original state $s_t,\forall\, t\in[0,T]$.

\begin{theorem}[Fixed policy shift error]
    \label{theo:fixedpolicy}
    Assume that Assumption \ref{ass:trans}, \ref{ass:policy}, \ref{ass:rew} hold and the encoder $\phi(\cdot)$ is $L_\phi$-Lipschitz, i.e., $\forall s,s^\prime, \|\phi(s) - \phi(s^\prime)\|\le L_\phi \|s - s^\prime\|$, then we have
    \begin{equation}
    \begin{gathered}
        \left\|\mathbb{E}_\xi [J(\phi(f(\tau(\xi;\pi,\mathcal{T},\mathcal{I}))))] - \mathbb{E}_\xi [J(\phi(\tau(\xi;\pi,\mathcal{T},\mathcal{I})))] \right\|
        \le L_{r_2}L_{\pi_1}L_\phi \sum_{t=0}^T \gamma^t \mathbb{E}_\xi\left[\|f(s_t) - s_t\|\right].
    \end{gathered}
    \end{equation}
\end{theorem}

\begin{proof}
    Since the transition dynamics, initialization function, and policy are kept unchanged, we have
    \begin{align*}
        &\left\|\mathbb{E}_\xi [J(\phi(f(\tau(\xi;\pi,\mathcal{T},\mathcal{I}))))] - \mathbb{E}_\xi [J(\phi(\tau(\xi;\pi,\mathcal{T},\mathcal{I})))] \right\| \\
         &\qquad\qquad = \left\|\int_\xi [J(\phi(f(\tau(\xi;\pi,\mathcal{T},\mathcal{I}))))]d \xi - \int_\xi [J(\phi(\tau(\xi;\pi,\mathcal{T},\mathcal{I})))] d\xi \right\| \\
         &\qquad\qquad \le \int_\xi \left\|[J(\phi(f(\tau(\xi;\pi,\mathcal{T},\mathcal{I}))))] -  [J(\phi(\tau(\xi;\pi,\mathcal{T},\mathcal{I})))] \right\| d\xi.
    \end{align*}
    Note that the only difference between the two trajectories lies in whether it involves the transpose function. Therefore, we have
    \begin{align*}
        &\quad \left\|[J(\phi(f(\tau(\xi;\pi,\mathcal{T},\mathcal{I}))))] -  [J(\phi(\tau(\xi;\pi,\mathcal{T},\mathcal{I})))] \right\| \\
        &= \left\|\sum_{t=0}^T \gamma^t r(s_t, \pi(\phi(f(s_t)))) - \sum_{t=0}^T \gamma^t r(s_t,\pi(\phi(s_t)))\right\| \\
        &\le \sum_{t=0}^T \gamma^t \left\| r(s_t, \pi(\phi(f(s_t)))) - r(s_t,\pi(\phi(s_t))) \right\| \\
        &\le \sum_{t=0}^T \gamma^t L_{r_2} \|\pi(\phi(f(s_t))) - \pi(\phi(s_t))\| \\
        &\le \sum_{t=0}^T \gamma^t L_{r_2} L_{\pi_1} \|\phi(f(s_t)) - \phi(s_t)\| \\
        &\le L_{r_2} L_{\pi_1} L_\phi \sum_{t=0}^T \gamma^t \|f(s_t) - s_t\|.
    \end{align*}
    Then, it is natural to derive that
    \begin{align*}
        &\left\|\mathbb{E}_\xi [J(\phi(f(\tau(\xi;\pi,\mathcal{T},\mathcal{I}))))] - \mathbb{E}_\xi [J(\phi(\tau(\xi;\pi,\mathcal{T},\mathcal{I})))] \right\| \\
        &\qquad\qquad \le \int_\xi \left\|[J(\phi(f(\tau(\xi;\pi,\mathcal{T},\mathcal{I}))))] -  [J(\phi(\tau(\xi;\pi,\mathcal{T},\mathcal{I})))] \right\| d\xi \\
        &\qquad\qquad \le L_{r_2} L_{\pi_1} L_\phi \int_\xi \sum_{t=0}^T \gamma^t \|f(s_t) - s_t\| d\xi  = L_{r_2} L_{\pi_1} L_\phi\sum_{t=0}^T \gamma^t \mathbb{E}_\xi\left[ \|f(s_t) - s_t\| \right].
    \end{align*}
\end{proof}

Furthermore, if the transpose function $f(s)$ represents a linear noise distractor, the above bound can be simplified, as illustrated in the following corollary. Note that it is common that the noise is involved in a linear way upon the visual observations in real-world applications \shortcite{Stone2021TheDC}.

\begin{corollary}
\label{coro:transposeerror}
If the transpose function satisfies $f(s_t) = s_t + \epsilon_t$ where $\epsilon_t$ is time-dependent bounded transpose term, i.e., $\forall t, \|\epsilon_t\|\le \eta<\infty$, then based on the assumptions in Theorem \ref{theo:fixedpolicy}, we have:
\begin{equation}
\begin{gathered}
    \|\mathbb{E}_\xi [J(\phi(f(\tau(\xi;\pi,\mathcal{T},\mathcal{I}))))] - \mathbb{E}_\xi [J(\phi(\tau(\xi;\pi,\mathcal{T},\mathcal{I})))]\|
    \le L_{r_2}L_{\pi_1}L_\phi \eta \dfrac{1-\gamma^{T+1}}{1-\gamma}.
\end{gathered}
\end{equation}
\end{corollary}

\begin{proof}
    The proof is quite straightforward. With the conclusion in Theorem \ref{theo:fixedpolicy}, we have
    \begin{align*}
        &\left\|\mathbb{E}_\xi [J(\phi(f(\tau(\xi;\pi,\mathcal{T},\mathcal{I}))))] - \mathbb{E}_\xi [J(\phi(\tau(\xi;\pi,\mathcal{T},\mathcal{I})))] \right\|\le L_{r_2} L_{\pi_1} L_\phi\sum_{t=0}^T \gamma^t \mathbb{E}_\xi\left[ \|f(s_t) - s_t\| \right].
    \end{align*}
    Then by using that $f(s_t) = s_t + \epsilon_t$, and $\|\epsilon_t\|\le\eta,\forall t$, we have
    \begin{align*}
        \left\|\mathbb{E}_\xi [J(\phi(f(\tau(\xi;\pi,\mathcal{T},\mathcal{I}))))] - \mathbb{E}_\xi [J(\phi(\tau(\xi;\pi,\mathcal{T},\mathcal{I})))] \right\|
        &\le L_{r_2} L_{\pi_1} L_\phi\sum_{t=0}^T \gamma^t \mathbb{E}_\xi\left[ \|f(s_t) - s_t\| \right]\\
        &=L_{r_2} L_{\pi_1} L_\phi\sum_{t=0}^T \gamma^t \mathbb{E}_\xi\left[ \|\epsilon_t\| \right]\\
        &\le L_{r_2} L_{\pi_1} L_\phi\sum_{t=0}^T \gamma^t \mathbb{E}_\xi\left[ \eta \right] \\
        &= L_{r_2} L_{\pi_1} L_\phi\eta \dfrac{1-\gamma^{T+1}}{1-\gamma}.
    \end{align*}
    This concludes the proof.
\end{proof}

The above corollary requires that the noises are linear, which might be restrictive in practice. We then loose this requirement and even allow that the transpose function does not necessarily to be deterministic. Instead, the transposed state is sampled from an unknown distribution. We do not make any assumption on this distribution. For the benefit of theoretical derivation, we assume that the average effect of the distraction (i.e., transpose function $f$) is bounded, and the variance of $f(s)$ gives $\sigma^2$. Then, we can still derive a similar bound as Corollary \ref{coro:transposeerror} and show that the policy performance shift after adding $f(\cdot)$ can be large even on the identical trajectory.

\begin{corollary}
\label{coro:unknowndistribution}
If the transpose function $f(s_t)$ is sampled from a distribution and satisfies $\|\mathbb{E}[f(s_t)] - s_t\|\le \eta<\infty$, i.e., the average effect of the distraction is bounded, and suppose that the variance of $f(s)$ is $\sigma^2$ which is fixed, then based on the assumptions in Theorem \ref{theo:fixedpolicy}, we have with probability at least $1-\delta$:
\begin{equation}
\begin{gathered}
    \|\mathbb{E}_\xi [J(\phi(f(\tau(\xi;\pi,\mathcal{T},\mathcal{I}))))] - \mathbb{E}_\xi [J(\phi(\tau(\xi;\pi,\mathcal{T},\mathcal{I})))]\|
    \le L_{r_2}L_{\pi_1}L_\phi \dfrac{1-\gamma^{T+1}}{1-\gamma}\left( \eta + \sqrt{\dfrac{\sigma^2}{\delta}} \right).
\end{gathered}
\end{equation}
\end{corollary}

\begin{proof}
    Based on the conclusion in Theorem \ref{theo:fixedpolicy}, we have
    \begin{align*}
        &\left\|\mathbb{E}_\xi [J(\phi(f(\tau(\xi;\pi,\mathcal{T},\mathcal{I}))))] - \mathbb{E}_\xi [J(\phi(\tau(\xi;\pi,\mathcal{T},\mathcal{I})))] \right\|\\
        &\qquad\qquad \le L_{r_2} L_{\pi_1} L_\phi\sum_{t=0}^T \gamma^t \mathbb{E}_\xi\left[ \|f(s_t) - s_t\| \right] \\
        &\qquad\qquad \le L_{r_2} L_{\pi_1} L_\phi\sum_{t=0}^T \gamma^t \mathbb{E}_\xi\left[ \|f(s_t) - \mathbb{E}[f(s_t)]\| + \| \mathbb{E}[f(s_t)] - s_t\| \right] \\
        &\qquad\qquad \le L_{r_2} L_{\pi_1} L_\phi\sum_{t=0}^T \gamma^t \mathbb{E}_\xi\left[\eta + \|f(s_t) - \mathbb{E}[f(s_t)]\| \right].
    \end{align*}
    Then by Chebyshev's inequality, we have with probability at least $1-\delta$, $\|f(s_t) - \mathbb{E}[f(s_t)]\| \le \sqrt{\frac{\sigma^2}{\delta}}$. It is then easy to find that,
    \begin{align*}
        &\left\|\mathbb{E}_\xi [J(\phi(f(\tau(\xi;\pi,\mathcal{T},\mathcal{I}))))] - \mathbb{E}_\xi [J(\phi(\tau(\xi;\pi,\mathcal{T},\mathcal{I})))] \right\| \\
        &\qquad\qquad\le L_{r_2} L_{\pi_1} L_\phi\sum_{t=0}^T \gamma^t \mathbb{E}_\xi\left[\eta + \|f(s_t) - \mathbb{E}[f(s_t)]\| \right]\\
        &\qquad\qquad\le L_{r_2} L_{\pi_1} L_\phi\sum_{t=0}^T \gamma^t \mathbb{E}_\xi\left[ \eta + \sqrt{\dfrac{\sigma^2}{\delta}} \right] \\
        &\qquad\qquad \le L_{r_2}L_{\pi_1}L_\phi \dfrac{1-\gamma^{T+1}}{1-\gamma}\left( \eta + \sqrt{\dfrac{\sigma^2}{\delta}} \right).
    \end{align*}
    This concludes the proof.
\end{proof}

Theorem \ref{theo:fixedpolicy} demonstrate that given a fixed policy, the return deviation before and after adding the distractor in the trajectory is determined by the deviation between the state and the transposed state. Corollary \ref{coro:transposeerror} and Corollary \ref{coro:unknowndistribution} indicate that the policy performance shift after involving the transpose function $f(\cdot)$ can be large (e.g., $\eta$ is large) and the error accumulates with larger $T$, i.e., the upper bound becomes larger if $T$ increases. In practice, the testing environment may differ from the training environment, e.g., the initial state distribution, the transition probability. We allow a slight mismatch between the training and testing environments, as shown below.

\begin{assumption}
\label{ass:train-test}
    The transition $\mathcal{T}$, initialization function $\mathcal{I}$ in the training environment and $\mathcal{T}^\prime, \mathcal{I}^\prime$ in the testing environment satisfy: $\forall s,a,\xi,\|(\mathcal{T}-\mathcal{T}^\prime)(s,a,\xi)\|\le\zeta$, $\|(\mathcal{I}-\mathcal{I}^\prime)(\xi)\|\le \epsilon$.
\end{assumption}

We then investigate the theoretical bound when involving the above assumption. It is challenging because the transition dynamics and the initialization function between the training and testing environments can be different. We need to derive how state under the initialization $\mathcal{I}$ and state under $\mathcal{I}^\prime$ deviates each other at timestep $t$ because It is necessary to get the bound for returns. We show the corresponding lemma below.


\begin{lemma}
    \label{lemma:sametransitiondifferentinitial}
    Assume that Assumptions \ref{ass:trans}, \ref{ass:policy}, \ref{ass:rew} hold, the encoder $\phi(\cdot)$ is $L_\phi$-Lipschitz, the initialization function $\mathcal{I^\prime}$ in the testing environment satisfies $\forall \xi, \|(\mathcal{I}-\mathcal{I}^\prime)(\xi)\|\le\epsilon$, and the transition dynamics is the same for the training and testing environments. We further assume that the testing environment has no distractors, then under policy $\pi(\cdot;\theta)$, we have
    \begin{align}
        \|s_t - s_t^\prime\| \le (L_{t_1} + L_{t_2}L_{\pi_1}L_\phi)^t\epsilon := \nu^t\epsilon, \forall s_t,s_t^\prime.
    \end{align}
\end{lemma}
\begin{proof}
    With the framework of reparameterized visual RL, we have
    \begin{align*}
        \|s_t - s_t^\prime\|
        &= \|\mathcal{T}(s_{t-1},\pi(\phi(s_{t-1});\theta),\xi_{t-1}) - \mathcal{T}(s_{t-1}^\prime,\pi(\phi(s_{t-1}^\prime);\theta),\xi_{t-1})\| \\
        &\le \|\mathcal{T}(s_{t-1},\pi(\phi(s_{t-1});\theta),\xi_{t-1}) - \mathcal{T}(s_{t-1},\pi(\phi(s_{t-1}^\prime);\theta),\xi_{t-1})\| \\
        &\qquad  + \|\mathcal{T}(s_{t-1},\pi(\phi(s_{t-1}^\prime);\theta),\xi_{t-1}) - \mathcal{T}(s_{t-1}^\prime,\pi(\phi(s_{t-1}^\prime);\theta),\xi_{t-1})\| \\
        &\le L_{t_1}\|s_{t-1} - s_{t-1}^\prime\| + L_{t_2} \| \pi(\phi(s_{t-1});\theta) - \pi(\phi(s_{t-1}^\prime);\theta) \| \\
        &\le L_{t_1}\|s_{t-1} - s_{t-1}^\prime\| + L_{t_2} L_{\pi_1} \|\phi(s_{t-1}) - \phi(s_{t-1}^\prime)\| \\
        &\le L_{t_1}\|s_{t-1} - s_{t-1}^\prime\| + L_{t_2} L_{\pi_1} L_\phi \|s_{t-1} - s_{t-1}^\prime\| \\
        &= (L_{t_1} + L_{t_2}L_{\pi_1}L_\phi)\|s_{t-1} - s_{t-1}^\prime\|.
    \end{align*}
    By denoting $\nu = L_{t_1} + L_{t_2}L_{\pi_1}L_\phi$, and doing iteration, we have
    \begin{align*}
        \|s_t - s_t^\prime\| &\le \nu^t\|s_0-s_0^\prime\| = \nu^t\epsilon,
    \end{align*}
    where the last equality is due to the fact that $\|s_0-s_0^\prime\| = \|(\mathcal{I}-\mathcal{I}^\prime)(\xi)\|\le\epsilon$.
\end{proof}

We also need the relationship between $s_t$ under training environment dynamics $\mathcal{T}$ and $s_t^\prime$ testing environment dynamics $\mathcal{T}^\prime$. The resulting bound is shown below.

\begin{lemma}
    \label{lemma:sameintialdifferenttransition}
    Assume that Assumptions \ref{ass:trans}, \ref{ass:policy}, \ref{ass:rew} hold, the encoder $\phi(\cdot)$ is $L_\phi$-Lipschitz, the transition dynamics $\mathcal{T^\prime}$ in the testing environment satisfies $\forall s,a,\xi, \|(\mathcal{T}-\mathcal{T}^\prime)(\xi)\|\le\zeta$, and the initialization function is the same for the training and testing environments. We further assume that the testing environment has no distractors. Suppose $s$ comes from the training environment, and $s^\prime$ from the testing environment, then under policy $\pi(\cdot;\theta)$, we have
    \begin{align}
        \|s_t - s_t^\prime\| \le \zeta \dfrac{1-\nu^t}{1-\nu},
    \end{align}
    where $\nu = L_{t_1} + L_{t_2}L_{\pi_1}L_\phi$.
\end{lemma}

\begin{proof}
    Since $s_t$ is from the training environment, and $s_t^\prime$ is from the testing environment, it is easy to find that
    \begin{align*}
        \|s_t - s_t^\prime\|
        &= \|\mathcal{T}(s_{t-1},\pi(\phi(s_{t-1});\theta),\xi_{t-1}) - \mathcal{T}^\prime(s_{t-1}^\prime, \pi(\phi(s_{t-1}^\prime);\theta),\xi_{t-1})\| \\
        &\le \|\mathcal{T}(s_{t-1},\pi(\phi(s_{t-1});\theta),\xi_{t-1}) - \mathcal{T}(s_{t-1}^\prime,\pi(\phi(s_{t-1}^\prime);\theta),\xi_{t-1})\| \\
        &\qquad \|\mathcal{T}(s_{t-1}^\prime,\pi(\phi(s_{t-1}^\prime);\theta),\xi_{t-1}) - \mathcal{T}^\prime(s_{t-1}^\prime, \pi(\phi(s_{t-1}^\prime);\theta),\xi_{t-1})\| \\
        &\le \zeta + \|\mathcal{T}(s_{t-1},\pi(\phi(s_{t-1});\theta),\xi_{t-1}) - \mathcal{T}(s_{t-1}^\prime,\pi(\phi(s_{t-1}^\prime);\theta),\xi_{t-1})\|.
    \end{align*}
    By following the same procedure in the proof of Lemma \ref{lemma:sametransitiondifferentinitial}, we have
    \begin{align*}
        &\|\mathcal{T}(s_{t-1},\pi(\phi(s_{t-1});\theta),\xi_{t-1}) - \mathcal{T}(s_{t-1}^\prime,\pi(\phi(s_{t-1}^\prime);\theta),\xi_{t-1})\|\le \nu\|s_{t-1}-s_{t-1}^\prime\|.
    \end{align*}
    Combining these results, we have
    \begin{align*}
        \|s_t - s_t^\prime\| \le \zeta + \nu \|s_{t-1}-s_{t-1}^\prime\|.
    \end{align*}
    By recursion, we have
    \begin{align*}
        \|s_t - s_t^\prime\| \le \zeta (1 + \nu + \ldots + \nu^{t-1}) + \nu^t \|s_0-s_0^\prime\| = \zeta\dfrac{1-\nu^t}{1-\nu}.
    \end{align*}
    The last equality is due to the fact that the initialization function is identical for the training and testing environments, and hence $s_0 = s_0^\prime$.
\end{proof}

Now, we are ready to bound the returns in the training environment (without distractor) and testing environment (with distractors). The theorem below tells us that the performance difference in the training and testing environments are jointly influenced by (1) the initialization difference $\epsilon$, (2) transition dynamics difference $\zeta$, and (3) the representation difference $\varrho$.

\begin{theorem}
    \label{theo:traintestfixpolicy}
    Assume that Assumptions \ref{ass:trans}, \ref{ass:policy}, \ref{ass:rew}, \ref{ass:train-test} hold. We further assume that (1) the encoder $\phi(\cdot)$ is $L_\phi$-Lipschitz, i.e., $\forall s,s^\prime, \|\phi(s) - \phi(s^\prime)\|\le L_\phi \|s - s^\prime\|$, and (2) $\|\phi(f(s)) - \phi(s)\|\le \varrho,\forall \, s$. $f(\cdot)$ is the transpose function in the testing environment. Denote $\nu = L_{t_1}+L_{t_2}L_{\pi_1}L_\phi, \lambda = L_{r_1}+L_{r_2}L_{\pi_1}L_\phi$, then we have,
    \begin{align*}
        \| \underbrace{\mathbb{E}_\xi [J(\phi(f(\tau(\xi;\pi,\mathcal{T}^\prime,\mathcal{I}^\prime))))]}_{\rm performance\, in\, the\, testing\, env} &- \underbrace{\mathbb{E}_\xi [J(\phi(\tau(\xi;\pi,\mathcal{T},\mathcal{I})))]}_{\rm performance \, in \, the \, training \, env}\| \\
        &\le \lambda \zeta \sum_{t=0}^T \gamma^t \dfrac{\nu^t-1}{\nu-1} + \lambda\epsilon\sum_{t=0}^T \gamma^t \nu^t + \dfrac{L_{r_2}L_{\pi_1} \varrho }{1-\gamma}(1-\gamma^{T+1}).
    \end{align*}
\end{theorem}

\begin{proof}
    We decompose the performance difference in the training and testing environments as follows:
    \begin{align*}
        \| \mathbb{E}_\xi [J(\phi(f(\tau(\xi;\pi,\mathcal{T}^\prime,\mathcal{I}^\prime))))] &- \mathbb{E}_\xi [J(\phi(\tau(\xi;\pi,\mathcal{T},\mathcal{I})))]\| \\
        &\le 
        \underbrace{\| \mathbb{E}_\xi [J(\phi(f(\tau(\xi;\pi,\mathcal{T}^\prime,\mathcal{I}^\prime))))] - \mathbb{E}_\xi [J(\phi(\tau(\xi;\pi,\mathcal{T}^\prime,\mathcal{I}^\prime)))]\|}_{\rm (I)} \\
        & + \underbrace{\| \mathbb{E}_\xi [J(\phi(\tau(\xi;\pi,\mathcal{T}^\prime,\mathcal{I}^\prime)))] - \mathbb{E}_\xi [J(\phi(\tau(\xi;\pi,\mathcal{T}^\prime,\mathcal{I})))]\|}_{\rm (II)} \\
        & + \underbrace{\| \mathbb{E}_\xi [J(\phi(\tau(\xi;\pi,\mathcal{T}^\prime,\mathcal{I})))] - \mathbb{E}_\xi [J(\phi(\tau(\xi;\pi,\mathcal{T},\mathcal{I})))]\|}_{\rm (III)}.
    \end{align*}
    We then bound each term separately. For term (I), we have
    \begin{align*}
        {\rm (I)} &= \|\mathbb{E}_\xi [J(\phi(f(\tau(\xi;\pi,\mathcal{T}^\prime,\mathcal{I}^\prime))))] - \mathbb{E}_\xi [J(\phi(\tau(\xi;\pi,\mathcal{T}^\prime,\mathcal{I}^\prime)))]\| \\
        &\le \int_\xi\left\|  \sum_{t=0}^T \gamma^t\left( r(s_t,\pi(\phi(f(s_t)))) - r(s_t,\pi(\phi(s_t))) \right) \right\|d\xi \\
        &\le \int_\xi\sum_{t=0}^T \gamma^t\left\|  \left( r(s_t,\pi(\phi(f(s_t)))) - r(s_t,\pi(\phi(s_t))) \right) \right\|d\xi \\
        &\le \int_\xi\sum_{t=0}^T \gamma^t L_{r_2}\left\|  \pi(\phi(f(s_t))) - \pi(\phi(s_t)) \right\|d\xi \\
        &\le \int_\xi\sum_{t=0}^T \gamma^t L_{r_2}L_{\pi_1}\left\|  \phi(f(s_t)) - \phi(s_t) \right\|d\xi \\
        &\le \int_\xi\sum_{t=0}^T \gamma^t L_{r_2}L_{\pi_1}\varrho d\xi = L_{r_2}L_{\pi_1}\varrho\dfrac{1-\gamma^{T+1}}{1-\gamma}.
    \end{align*}
    For term (II), we have
    \begin{align*}
        {\rm (II)} &= \|\mathbb{E}_\xi [J(\phi(\tau(\xi;\pi,\mathcal{T}^\prime,\mathcal{I}^\prime)))] - \mathbb{E}_\xi [J(\phi(\tau(\xi;\pi,\mathcal{T}^\prime,\mathcal{I})))]\| \\
        &\le \int_\xi\left\| \sum_{t=0}^T \gamma^t \left( r(s_t,\pi(\phi(s_t);\theta)) - r(s_t^\prime,\pi(\phi(s_t^\prime);\theta)) \right) \right\|d\xi \\
        &\le \int_\xi\sum_{t=0}^T \gamma^t \|r(s_t,\pi(\phi(s_t);\theta)) - r(s_t^\prime,\pi(\phi(s_t^\prime);\theta))\| d\xi.
    \end{align*}
    By using Lemma \ref{lemma:rewdev}, and that the encoder is $L_\phi$-Lipschitz, we have
    \begin{align*}
        {\rm (II)} &\le \int_\xi\sum_{t=0}^T \gamma^t \left( L_{r_1}\|s_t - s_t^\prime\| + L_{r_2} L_{\pi_1} \|\phi(s_t) - \phi(s_t^\prime)\| \right) d\xi \\
        &\le \int_\xi\sum_{t=0}^T \gamma^t \underbrace{(L_{r_1} + L_{r_2} L_{\pi_1}L_\phi)}_{:=\lambda} \|s_t - s_t^\prime\| d\xi
    \end{align*}
    Term (II) has the identical transition dynamics but different initialization functions, then by using Lemma \ref{lemma:sametransitiondifferentinitial}, we have
    \begin{align*}
        {\rm (II)} &\le \lambda \int_\xi \sum_{t=0}^T\gamma^t \nu^t\epsilon d\xi = \lambda\epsilon \sum_{t=0}^T \gamma^t\nu^t.
    \end{align*}
    For term (III), we follow the same way of bounding term (II) and have
    \begin{align*}
        {\rm (III)} &= \| \mathbb{E}_\xi [J(\phi(\tau(\xi;\pi,\mathcal{T}^\prime,\mathcal{I})))] - \mathbb{E}_\xi [J(\phi(\tau(\xi;\pi,\mathcal{T},\mathcal{I})))]\| \le \int_\xi\sum_{t=0}^T \gamma^t \lambda \|s_t - s_t^\prime\| d\xi.
    \end{align*}
    Term (III) has the same initialization function, but different transition dynamics. By using Lemma \ref{lemma:sameintialdifferenttransition}, we have
    \begin{align*}
        {\rm (III)} &\le \int_\xi\sum_{t=0}^T \gamma^t \lambda \|s_t - s_t^\prime\| d\xi \le \int_\xi \sum_{t=0}^T \gamma^t \lambda\zeta \dfrac{1-\nu^t}{1-\nu} d\xi = \lambda\zeta\sum_{t=0}^T \gamma^t\dfrac{1-\nu^t}{1-\nu}.
    \end{align*}
    Finally, Theorem \ref{theo:traintestfixpolicy} holds by combining the bounds of term (I), (II), and (III).
\end{proof}

\textbf{Remark 1:} By assuming $\|\phi(f(s)) - \phi(s)\|\le \varrho,\forall \, s$, we actually are enforcing a `local' regularization on the encoder, i.e., the representation distance upon the captured state between the training environment $s$ and the testing environment $f(s)$ ought to be bounded. This naturally leads to yet another requirement on the Lipschitz constant $L_\phi$, i.e., it has to satisfy $L_\phi \le \frac{\varrho}{\max_{s\in\mathcal{S}}\| f(s) - s \|}$.

\textbf{Remark 2:} The above theorem does not involve the process of policy improvement (which is why we emphasize a fixed policy), and is simply the performance of \emph{any} policy in the training and testing environments. However, the result is still meaningful as it meets the most common way of evaluating RL policies, i.e., evaluate the agent with the fixed policy after certain environmental steps.

We note that among the all critical factors, the initialization difference $\epsilon$ and the transition dynamics difference $\zeta$ are uncontrollable, and are strongly correlated with the physical systems. \textit{It turns out that the most critical way of reducing the performance shift is to regularize the representation difference $\varrho$}. Before moving to our most critical conclusion, we need to present the formal generalization error of visual RL when the testing environment has distractors. Recall that the generalization error has the form of $\| \mathbb{E}_{\tau\sim\mathcal{D}^\prime_{\hat{\pi}}}[J(\tau)] - \frac{1}{n}\sum_{i=1}^n J(\tau_i) \|_2^2$ (as illustrated in Section \ref{sec:preliminaries}). To that end, we first present the following lemma, which is a direct product of classical learning theory (i.e., Theorem 3.3 in \shortcite{mohri2018foundations}).

\begin{lemma}
    \label{lemma:bound}
    For reparameterized visual RL, if the rewards are bounded, $|r(s,a)|\le r_{\rm max}$, $\forall \,s,a, r_{\rm max}>0$. Assume that the sampled peripheral random variables $\xi$s are i.i.d. for each episode. Then with probability at least $1-\delta$, for any policy $\pi\in\Pi$, the following holds,
    \small
    \begin{align*}
        &\left\|\mathbb{E}_\xi \left[J\left(\phi\left(\tau\left(\xi; \pi, \mathcal{T},\mathcal{I}\right)\right)\right)\right] - \dfrac{1}{n} \sum_{i=1}^n J(\phi(\tau(\xi_i;\pi,\mathcal{T},\mathcal{I}))) \right\| \le 2Rad(J_{\pi,\mathcal{T},\mathcal{I}}) + \mathcal{O}\left( r_{\rm max} \sqrt{\frac{\log (1/\delta)}{n}}\right),
    \end{align*}
    where $Rad(J_{\pi,\mathcal{T},\mathcal{I}})=\mathbb{E}_{\xi}\mathbb{E}_\sigma [\sup_{\pi} \frac{1}{n}\sum_{i=1}^n \sigma_i J(\phi(\tau(\xi_i;\pi,\mathcal{T},\mathcal{I})))]$ is the Rademacher complexity, $\sigma_i$ is the Rademacher variable, $n$ is the number of training episodes.
\end{lemma}

\begin{proof}
    Please refer to the proof of Theorem 3.3 in \shortcite{mohri2018foundations}.
\end{proof}

The above bound holds uniformly for all policies $\pi\in\Pi$. Then, we formally present the generalization gap bound of the policy in the testing environment with distractors and the clean training environment.

\begin{theorem}[Generalization error]
\label{theo:generalizationerror}
    Assume that the assumptions made in Theorem \ref{theo:traintestfixpolicy} hold and $\xi_i$ are i.i.d.. Denote $f(\cdot)$ as the transpose function in the testing environment. Then with probability at least $1-\delta$, the generalization gap between the training environment and testing environment satisfies:
    \small
    \begin{align*}
        &\left\|\mathbb{E}_\xi \left[J\left(\phi\left(f\left(\tau\left(\xi; \pi, \mathcal{T}^\prime,\mathcal{I}^\prime\right)\right)\right)\right)\right] - \dfrac{1}{n} \sum_{i=1}^n J(\phi(\tau(\xi_i;\pi,\mathcal{T},\mathcal{I}))) \right\| \\
        &\le 2 Rad(J_{\pi,\mathcal{T},\mathcal{I}}) + \lambda \zeta \sum_{t=0}^T \gamma^t \dfrac{\nu^t-1}{\nu-1} + \lambda\epsilon\sum_{t=0}^T \gamma^t \nu^t + \dfrac{L_{r_2}L_{\pi_1} \varrho (1-\gamma^{T+1})}{1-\gamma} +\mathcal{O}\left( r_{\rm max} \sqrt{\dfrac{\log (1/\delta)}{n}} \right),
    \end{align*}
    where $Rad(J_{\pi,\mathcal{T},\mathcal{I}})=\mathbb{E}_{\xi}\mathbb{E}_\sigma [\sup_{\pi} \frac{1}{n}\sum_{i=1}^n \sigma_i J(\phi(\tau(\xi_i;\pi,\mathcal{T},\mathcal{I})))]$ is the Rademacher complexity, $\nu = L_{t_1}+L_{t_2}L_{\pi_1}L_\phi, \lambda = L_{r_1}+L_{r_2}L_{\pi_1}L_\phi$, and $n$ is the number of training episodes.
\end{theorem}

\begin{proof}
    We decompose the target as follows:
    \begin{align*}
        &\left\|\mathbb{E}_\xi \left[J\left(\phi\left(f\left(\tau\left(\xi; \pi, \mathcal{T}^\prime,\mathcal{I}^\prime\right)\right)\right)\right)\right] - \dfrac{1}{n} \sum_{i=1}^n J(\phi(\tau(\xi_i;\pi,\mathcal{T},\mathcal{I}))) \right\| \\
        &\le \underbrace{\left\|\mathbb{E}_\xi \left[J\left(\phi\left(f\left(\tau\left(\xi; \pi, \mathcal{T}^\prime,\mathcal{I}^\prime\right)\right)\right)\right)\right] - \mathbb{E}_\xi \left[J\left(\phi\left(\tau\left(\xi; \pi, \mathcal{T},\mathcal{I}\right)\right)\right)\right] \right\|}_{\rm (I)} \\
        &\qquad\qquad + \underbrace{\left\| \mathbb{E}_\xi \left[J\left(\phi\left(\tau\left(\xi; \pi, \mathcal{T},\mathcal{I}\right)\right)\right)\right] - \dfrac{1}{n} \sum_{i=1}^n J(\phi(\tau(\xi_i;\pi,\mathcal{T},\mathcal{I}))) \right\|}_{\rm (II)}.
    \end{align*}
    Then, the conclusion holds by bounding term (I) with Theorem \ref{theo:traintestfixpolicy}, and term (II) with Lemma \ref{lemma:bound}.
\end{proof}

The remaining procedure is to bound the Rademacher complexity. To fulfill end, we first introduce the following lemma which shows that the empirical return $J$ also satisfies the Lipschitz condition.

\begin{lemma}
    \label{lemma:returnlipschitz}
    Assume that Assumptions \ref{ass:trans}, \ref{ass:policy}, \ref{ass:rew} hold, and the encoder is $L_\phi$-Lipschitz, then the empirical return $J(\phi(\tau(\cdot; \theta)))$, as a function of $\theta$, is $L_J$-Lipschitz, i.e.,
    \begin{equation}
    \begin{gathered}
        \|J(\phi(\tau(\cdot;\theta))) - J(\phi(\tau(\cdot;\theta^\prime)))\| \le L_J \|\theta - \theta^\prime\|, \\
        \mathrm{where}\, L_J = \sum_{t=0}^T \gamma^t \left( \lambda L_{t_2}L_{\pi_2}\dfrac{\nu^t-1}{\nu-1} + L_{r_2}L_{\pi_2} \right).
    \end{gathered}
    \end{equation}
\end{lemma}

\begin{proof}
    Under policies $\pi(\cdot;\theta), \pi(\cdot;\theta^\prime)$ with parameters $\theta,\theta^\prime$, different trajectories will be collected. However, the initialization process is identical (i.e., $s_0=s_0^\prime$) as it does not depend on $\theta$. Suppose $s\sim\tau(\cdot;\theta)$ and $s^\prime\sim\tau(\cdot;\theta^\prime)$, then we have
    \begin{align*}
        \|J(\phi(\tau(\cdot;\theta))) - J(\phi(\tau(\cdot;\theta^\prime)))\| 
        &=\left\| \sum_{t=0}^T\gamma^t r(s_t,\pi(\phi(s_t);\theta)) - \sum_{t=0}^T\gamma^t r(s_t^\prime,\pi(\phi(s_t^\prime);\theta^\prime)) \right\| \\
        &\le \sum_{t=0}^T \gamma^t |r(s_t,\pi(\phi(s_t);\theta)) - r(s_t^\prime,\pi(\phi(s_t^\prime);\theta^\prime))|.
    \end{align*}
    By following a similar procedure in proving Lemma \ref{lemma:rewdev}, we have
    \begin{align*}
        |r(s_t,\pi(\phi(s_t);\theta)) - r(s_t^\prime,\pi(\phi(s_t^\prime);\theta^\prime))| 
        &\le L_{r_1} \|s_t - s_t^\prime\| + L_{r_2} \|\pi(\phi(s_t);\theta) - \pi(\phi(s_t^\prime);\theta^\prime)\|.
    \end{align*}
    Furthermore, by using Lemma \ref{lemma:policydev}, we have
    \begin{align*}
        &|r(s_t,\pi(\phi(s_t);\theta)) - r(s_t^\prime,\pi(\phi(s_t^\prime);\theta^\prime))| \\
        &\le L_{r_1} \|s_t - s_t^\prime\| + L_{r_2} (L_{\pi_1}\|\phi(s_t^\prime) - \phi(s_t)\| + L_{\pi_2}\|\theta - \theta^\prime\|) \\
        &\le L_{r_1} \|s_t - s_t^\prime\| + L_{r_2} (L_{\pi_1} L_\phi\|s_t^\prime - s_t\| + L_{\pi_2}\|\theta - \theta^\prime\|) \\
        &= \underbrace{(L_{r_1}+L_{r_2}L_{\pi_1}L_\phi)}_{:=\lambda} \|s_t - s_t^\prime\| + L_{r_2}L_{\pi_2} \|\theta - \theta^\prime\|.
    \end{align*}
    We now investigate $\|s_t - s_t^\prime\|$, by following a similar procedure we use in proving Lemma \ref{lemma:statedev}, we have
    \begin{align*}
        \|s_t - s_t^\prime\|
        &= \|\mathcal{T}(s_{t-1},\pi(\phi(s_{t-1});\theta),\xi_{t-1}) - \mathcal{T}(s_{t-1}^\prime,\pi(\phi(s_{t-1}^\prime);\theta^\prime),\xi_{t-1})\| \\
        &\le \|\mathcal{T}(s_{t-1},\pi(\phi(s_{t-1});\theta),\xi_{t-1}) - \mathcal{T}(s_{t-1},\pi(\phi(s_{t-1}^\prime);\theta^\prime),\xi_{t-1})\| \\
        &\qquad + \|\mathcal{T}(s_{t-1},\pi(\phi(s_{t-1}^\prime);\theta^\prime),\xi_{t-1}) - \mathcal{T}(s_{t-1}^\prime,\pi(\phi(s_{t-1}^\prime);\theta^\prime),\xi_{t-1})\| \\
        &\le L_{t_1}\|s_{t-1} - s_{t-1}^\prime\| + L_{t_2} \| \pi(\phi(s_{t-1});\theta) - \pi(\phi(s_{t-1}^\prime);\theta^\prime) \|.
    \end{align*}
    Again, we use Lemma \ref{lemma:policydev}, and can derive
    \begin{align*}
        \|s_t - s_t^\prime\| 
        &\le L_{t_1}\|s_{t-1} - s_{t-1}^\prime\| + L_{t_2} (L_{\pi_1}\|\phi(s_{t-1}^\prime) - \phi(s_{t-1})\| + L_{\pi_2}\|\theta - \theta^\prime\|) \\
        &\le L_{t_1}\|s_{t-1} - s_{t-1}^\prime\| + L_{t_2} (L_{\pi_1} L_\phi \|s_{t-1}^\prime - s_{t-1}\| + L_{\pi_2}\|\theta - \theta^\prime\|) \\
        &= (L_{t_1} + L_{t_2} L_{\pi_1} L_\phi)\|s_{t-1} - s_{t-1}^\prime\| + L_{t_2}L_{\pi_2}\|\theta - \theta^\prime\| \\
        &= \nu\|s_{t-1} - s_{t-1}^\prime\| + L_{t_2}L_{\pi_2}\|\theta - \theta^\prime\|.
    \end{align*}
    By recursion, we can further have
    \begin{align*}
        \|s_t - s_t^\prime\| &\le \nu^t\|s_0 - s_0^\prime\| + L_{t_2}L_{\pi_2}\dfrac{1-\nu^t}{1-\nu}\|\theta - \theta^\prime\| = L_{t_2}L_{\pi_2}\dfrac{1-\nu^t}{1-\nu}\|\theta - \theta^\prime\|.
    \end{align*}
    The last equality is because $s_0 = s_0^\prime$. Combining the above results, and we have
    \begin{align*}
        \|J(\phi(\tau(\cdot;\theta))) - J(\phi(\tau(\cdot;\theta^\prime)))\|
        &=\left\| \sum_{t=0}^T\gamma^t r(s_t,\pi(\phi(s_t);\theta)) - \sum_{t=0}^T\gamma^t r(s_t^\prime,\pi(\phi(s_t^\prime);\theta^\prime)) \right\| \\
        &\le \sum_{t=0}^T \gamma^t |r(s_t,\pi(\phi(s_t);\theta)) - r(s_t^\prime,\pi(\phi(s_t^\prime);\theta^\prime))| \\
        &\le \sum_{t=0}^T \gamma^t (\lambda \|s_t - s_t^\prime\| + L_{r_2}L_{\pi_2}\|\theta - \theta^\prime\|) \\
        &\le \sum_{t=0}^T \gamma^t \left( \lambda L_{t_2}L_{\pi_2}\dfrac{1-\nu^t}{1-\nu}\|\theta - \theta^\prime\| + L_{r_2}L_{\pi_2}\|\theta - \theta^\prime\| \right) \\
        &= \underbrace{\sum_{t=0}^T \gamma^t \left( \lambda L_{t_2}L_{\pi_2}\dfrac{1-\nu^t}{1-\nu} + L_{r_2}L_{\pi_2} \right)}_{:=L_J} \|\theta - \theta^\prime\|.
    \end{align*}
    We thus can conclude the proof.
\end{proof}

Furthermore, if the number of parameters $m$ in the policy $\pi(\cdot;\theta)$ is bounded, the Rademacher complexity can be controlled, as desired.

\begin{lemma}
    \label{lemma:rademacherbound}
    Given Assumptions \ref{ass:trans}, \ref{ass:policy}, \ref{ass:rew}, and suppose the parameters of the policy $\theta\in\mathbb{R}^m$ is bounded such that $\|\theta\|\le K$, $K\in\mathbb{R}^+$, then the Rademacher complexity $Rad(J_{\pi,\mathcal{T},\mathcal{I}})$ is bounded, i.e.,
    \begin{equation}
        Rad(J_{\pi,\mathcal{T},\mathcal{I}}) = \mathcal{O}\left( L_J K \sqrt{\dfrac{m}{n}} \right).
    \end{equation}
\end{lemma}

\begin{proof}
    Note that we have a budget of $n$ episodes. Denote $A = ({\bf a}_1, {\bf a}_2,\ldots, {\bf a}_N)$ is a finite set of vectors, with each element containing $(J(\phi(\tau(\xi_1))), J(\phi(\tau(\xi_2))), \ldots, J(\phi(\tau(\xi_n))))$. For simplicity, we write $J_i = J(\phi(\tau(\xi_i)))$. One can see that $Rad(A) = Rad(J_{\pi,\mathcal{T},\mathcal{I}})$. Denote ${\bf \sigma}$ as a $n$-dimensional Rademacher variable. Denote $\langle f,g \rangle$ as the inner product of two vectors $f,g$. We define $\bar{\bf a} = \frac{1}{N}\sum_{i=1}^N {\bf a}_i$. Let $\lambda>0$ and $A^\prime = (\lambda a_1, \lambda a_2, \ldots, \lambda a_n)$. The proof is quite similar to that of the well-known Massart Lemma. We have
    \begin{align*}
        nRad(A^\prime) &= \mathbb{E}_\xi \mathbb{E}_{\bf \sigma} \left[ \max_{{\bf a}\in A} \langle {\bf \sigma}, \lambda {\bf a}\rangle \right] = \mathbb{E}_\xi \mathbb{E}_{\bf \sigma} \left[ \log \left( \max_{{\bf a}\in A} e^{\lambda \langle {\bf \sigma}, {\bf a} \rangle} \right) \right] \le \mathbb{E}_\xi \mathbb{E}_{\bf \sigma} \left[ \log \left( \sum_{{\bf a}\in A} e^{\lambda \langle {\bf \sigma}, {\bf a} \rangle} \right) \right].
    \end{align*}
    By using the Jensen's inequality, we have
    \small
    \begin{align*}
        nRad(A^\prime) \le \mathbb{E}_\xi \mathbb{E}_{\bf \sigma} \left[ \log \left( \sum_{{\bf a}\in A} e^{\lambda \langle {\bf \sigma}, {\bf a} \rangle} \right) \right] &\le \log \left( \mathbb{E}_\xi \mathbb{E}_{\bf \sigma} \left[ \sum_{{\bf a}\in A} e^{\lambda \langle {\bf \sigma}, {\bf a} \rangle} \right] \right) = \log \left( \sum_{{\bf a}\in A} \mathbb{E}_\xi \mathbb{E}_{\bf \sigma} \left[ e^{\lambda \langle {\bf \sigma}, {\bf a} \rangle} \right] \right). 
    \end{align*}
    Expanding the above formula, and we have
    \begin{align*}
        nRad(A^\prime) \le \log \left( \sum_{{\bf a}\in A} \mathbb{E}_\xi \mathbb{E}_{\bf \sigma} \left[ e^{\lambda \langle {\bf \sigma}, {\bf a} \rangle} \right] \right) =  \log \left( \sum_{{\bf a}\in A} \prod_{i=1}^n \mathbb{E}_\xi \mathbb{E}_{\bf \sigma} \left[ e^{\lambda \sigma_i J_i} \right] \right). 
    \end{align*}
    We now claim that $\dfrac{e^a + e^{-a}}{2}\le e^{\frac{a^2}{2}}, \forall a\in\mathbb{R}$. Notice that $e^a = \sum_{n=0}^\infty \dfrac{a^n}{n!}$, hence $\dfrac{e^a + e^{-a}}{2}=\sum_{n=0}^\infty \dfrac{a^{2n}}{(2n)!}$. Furthermore, we have $e^{\frac{a^2}{2}} = \dfrac{a^{2n}}{n! 2^n}$. The claim holds since $(2n)!\ge 2^n n!$.

    Therefore, $\mathbb{E}_{\sigma_i}[e^{\sigma_i J_i}] = \dfrac{e^{J_i} + e^{-J_i}}{2}\le e^{J_i^2/2}$, and we have
    \begin{align*}
        nRad(A^\prime) = n\lambda Rad(J_{\pi,\mathcal{T},\mathcal{I}}) &\le \log \left( \sum_{{\bf a}\in A} \prod_{i=1}^n \mathbb{E}_\xi \mathbb{E}_{\bf \sigma} \left[ e^{\lambda \sigma_i J_i} \right] \right) \\
        &\le \log \left( \sum_{{\bf a}\in A} \prod_{i=1}^n e^{\lambda^2J_i^2/2} \right) \\
        &=  \log \left( \sum_{{\bf a}\in A} e^{\lambda^2\|{\bf a}\|^2_2/2} \right) \\
        & \le \log \left( |A| \max_{{\bf a}\in A} e^{\lambda^2\|{\bf a}\|^2_2/2} \right) \\
        &= \log \left( |A| \right) + \max_{{\bf a}\in A} \frac{\lambda^2\|{\bf a}\|^2_2}{2}.
    \end{align*}
    Hence, we have
    \begin{align*}
        R(J_{\pi,\mathcal{T},\mathcal{I}}) &\le \dfrac{\log \left( |A| \right) + \max_{{\bf a}\in A} \frac{\lambda^2\|{\bf a}\|^2_2}{2}}{\lambda n}.
    \end{align*}
    Then, by setting $\lambda = \sqrt{\frac{2\log(|A|)}{\|{\bf a}\|^2_2}}$, we have
    \begin{align*}
        Rad(J_{\pi,\mathcal{T},\mathcal{I}}) &\le \|{\bf a}\|\dfrac{\sqrt{2\log(|A|)}}{n}.
    \end{align*}
    Based on Lemma 26.6 in \shortcite{shalev2014understanding}, the Rademacher complexity satisfies $Rad(\{{\bf a}+{\bf a}_0: {\bf a}\sim A\})=Rad(A)$. We can hence rewrite the above inequality into $Rad(J_{\pi,\mathcal{T},\mathcal{I}}) \le \|{\bf a} - \bar{\bf a}\|\dfrac{\sqrt{2\log(|A|)}}{n}$. Note that by definition, ${\bf a} = (J_1, \ldots, J_n)$. Therefore by using the $L_J$-Lipschitz property of the return, we have $\|{\bf a} - \bar{\bf a}\| \le \sqrt{n} L_J \|\theta - \theta^\prime\|\le \sqrt{n} L_J (\|\theta\| + \|\theta^\prime\|) \le 2K \sqrt{n} L_J$. We then have
    \begin{align*}
        Rad(J_{\pi,\mathcal{T},\mathcal{I}}) &\le 2K \sqrt{n} L_J\dfrac{\sqrt{2\log(|A|)}}{n}.
    \end{align*}
    It remains to bound $\log(|A|)$. Since the policy parameter $\theta\in\mathbb{R}^m$, and is bounded, then $\log (|A|)$ generally scales with $\mathcal{O}(d^m)$ \shortcite{shalev2014understanding} where $d$ is some bounded constant number. Finally, we have
    \begin{align*}
        Rad(J_{\pi,\mathcal{T},\mathcal{I}}) &\le \mathcal{O}\left(K L_J\sqrt{\dfrac{m}{n}}\right).
    \end{align*}
\end{proof}

Finally, by combining previous results, we get the following generalization bound.

\begin{theorem}
\label{theo:final}
    Suppose that the assumptions made in Theorem \ref{theo:generalizationerror} and Lemma \ref{lemma:rademacherbound} hold. Then we have with probability at least $1-\delta$, the generalization error gives,
    \small
    \begin{align*}
        &\left\|\mathbb{E}_\xi \left[J\left(\phi\left(f\left(\tau\left(\xi; \pi, \mathcal{T}^\prime,\mathcal{I}^\prime\right)\right)\right)\right)\right] - \dfrac{1}{n} \sum_{i=1}^n J(\phi(\tau(\xi_i;\pi,\mathcal{T},\mathcal{I}))) \right\| \\
        &\le \lambda \zeta \sum_{t=0}^T \gamma^t \dfrac{\nu^t-1}{\nu-1} + \lambda\epsilon\sum_{t=0}^T \gamma^t \nu^t + \dfrac{L_{r_2}L_{\pi_1} \varrho }{1-\gamma}(1-\gamma^{T+1}) +\mathcal{O}\left( L_J K \sqrt{\dfrac{m}{n}} \right) + \mathcal{O}\left(r_{\rm max} \sqrt{\dfrac{\log (1/\delta)}{n}} \right).
    \end{align*}
\end{theorem}

\begin{proof}
  It follows by combining Theorem \ref{theo:generalizationerror} and Lemma \ref{lemma:rademacherbound}.
\end{proof}

\textbf{Remark:} We summarize a key insight based on the above bound, \textit{the generalization gap can only be small if the representation distance between the training and testing environments is small}, due to the fact that $\varrho$ is the only factor that one can control in the bound. This observation is somewhat consistent with a human's intuition: the representations before and after involving distractors are similar and hence the policy can retrieve good behaviors it learned in the training environment.

Note that the Assumption \ref{ass:trans} requires that the environment dynamics is \emph{smooth}, i.e., no sudden change occurs. Our theory still holds if the Lipschitz continuity in terms of transition dynamics is slightly violated, e.g., the transition dynamics satisfies $\|\mathcal{T}(s,a,\xi) - \mathcal{T}(s^\prime,a,\xi)\|\le L_{t_1}\|s - s^\prime\| + B$, where $B\in\mathbb{R}, B< \infty$ is a bounded constant number. Then extra constant terms will be included in the bounds. Our core conclusion also holds since $B$ is a constant, i.e., under this condition, minimizing the representation deviation is still the key way of reducing the generalization error. However, we have to admit that if $B$ is uncontrollable, or no similar constraints like $\|\mathcal{T}(s,a,\xi) - \mathcal{T}(s^\prime,a,\xi)\|\le L_{t_1}\|s - s^\prime\| + B$ can be satisfied, our theories will fail and cannot be used to explain the generalization capability then. We stress that many real-world (or simulation) dynamics do not often encounter abrupt changes (if abrupt changes occur, those are often referred to as emergent conditions, e.g., the robot dog fall). It is hard to present a general theoretical framework that fits all of the conditions. We do not view this as a drawback of our theory.

\section{Experimental Support}
In this section, we resolve the following concerns: (1) are the assumptions we made in Section \ref{sec:theory} reasonable? (Section \ref{sec:rationality}) (2) do existing algorithms agree with our theoretical insight? (Section \ref{sec:align}). 

\subsection{The rationality of the assumptions}
\label{sec:rationality}

\begin{figure*}[htb]
    \centering
    \includegraphics[width=0.25\textwidth]{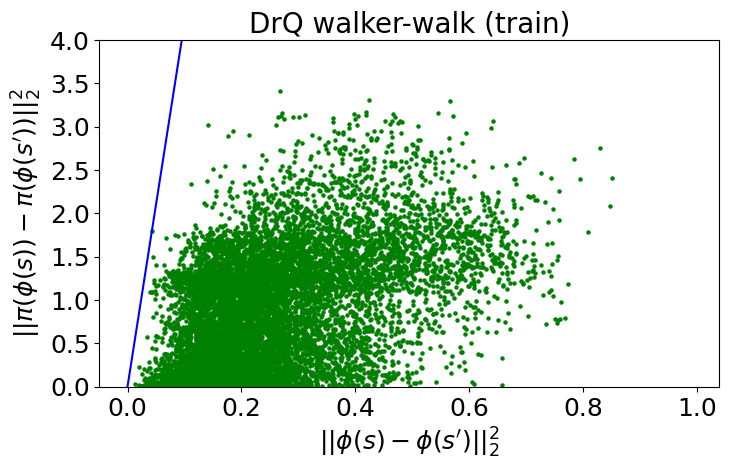}\hspace{-1mm}
    \includegraphics[width=0.25\textwidth]{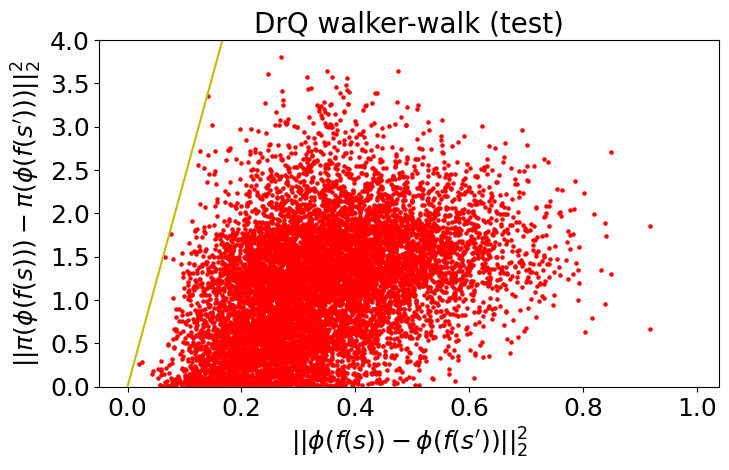}\hspace{-1mm}
    \includegraphics[width=0.25\textwidth]{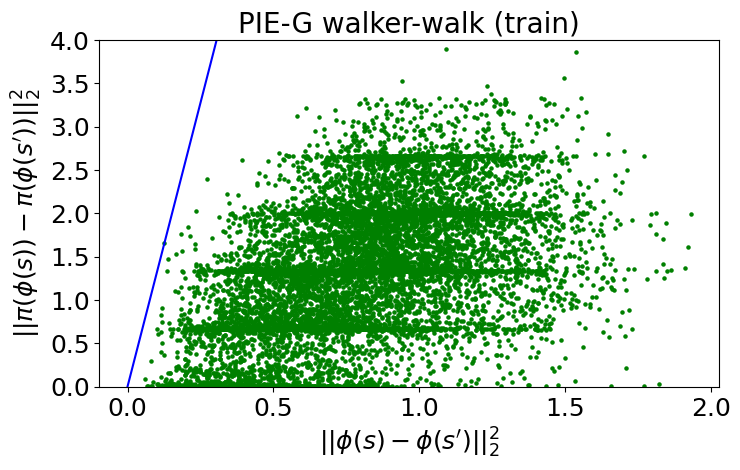}\hspace{-1mm}
    \includegraphics[width=0.25\textwidth]{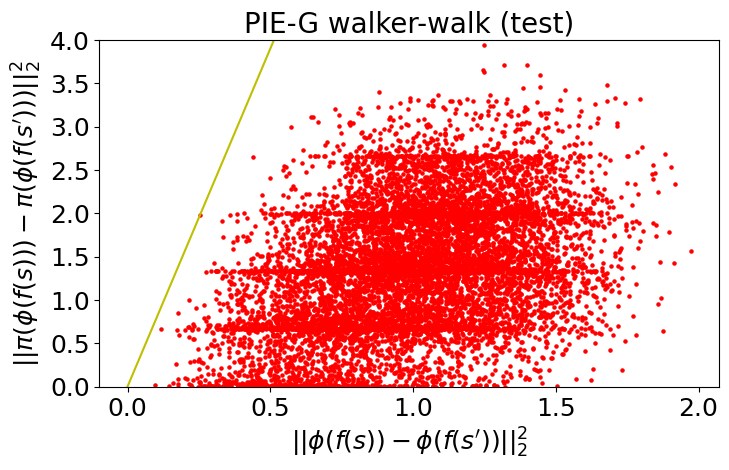}\hspace{-1mm}
    \caption{\textbf{Evidence that the Lipschitz condition of the policy holds before and after adding distractors.} We present the scatter plot of the policy deviation against the representation deviation of DrQ and PIE-G on walker-walk \emph{video-easy} task with and without distractors (i.e., the transpose function $f(\cdot)$). The solid line denotes the maximum slope in the batch, i.e., $y=kx$ where $k = \max\frac{\|\pi(\phi(s)) - \pi(\phi(s^\prime))\|}{\|\phi(s) - \phi(s^\prime)\|}$ for the training environment, and $k=\max\frac{\|\pi(\phi(f(s))) - \pi(\phi(f(s^\prime)))\|}{\|\phi(f(s)) - \phi(f(s^\prime))\|}$ for the testing environment. Since there always exist a $k$ such that the policy deviation of all samples can be bounded, the Lipschitz condition for the policy holds naturally.}
    \label{fig:rationality}
\end{figure*}

The main assumptions we made for reaching the theoretical bound are the \textit{smoothness} on the reward, policy, and transition dynamics. Note that such smoothness assumptions are widely adopted in prior work \shortcite{Liu2022LearningSN,Miyato2018SpectralNF,Scaman2018LipschitzRO,Lecarpentier2020LipschitzLR}. Since the reward functions are mostly designed manually, it is easy to satisfy the Lipschitz condition. Assuming the Lipschitz continuity in the environmental dynamics is also valid and it is satisfied if the state space is bounded and no sudden change in this physical system occurs. The Lipschitz assumption for the policy $\pi$ can be easily satisfied for many policy classes in practice, e.g., neural networks \shortcite{Fazlyab2019EfficientAA,DOro2020HowTL}. We empirically examine whether such property still holds with distractors. We compare the policy learned by DrQ \shortcite{Yarats2021ReinforcementLW}, which uses plain data augmentation, and PIE-G \shortcite{Yuan2022PreTrainedIE}, which leverages the pre-trained (on ImageNet) image encoder \shortcite{He2015DeepRL}, on \texttt{walker-walk} task from DMControl Generalization Benchmark (DMC-GB) \shortcite{Hansen2020GeneralizationIR}. To that end, we first gather 10 trajectories $\tau$ in the clean training environment with the learned DrQ agent, and draw the scatter plot of $\|\pi(\phi(s)) - \pi(\phi(s^\prime))\|_2^2$ against $\|\phi(s) - \phi(s^\prime)\|_2^2$ via sampling $s,s^\prime\sim\tau$ in a bootstrapping way for $10^4$ times. We then add distractors in $\tau$ by replacing the background of the agent with playing videos, and plot the corresponding scatter plot. The results are shown in Figure \ref{fig:rationality}, and one can find that the Lipschitz condition for the policy network is generally satisfied with or without distractors.

\subsection{Do existing methods align with the theoretical results?}
\label{sec:align}


We then examine whether our theory applies to existing algorithms and explains why they work in practice. Since our contribution mainly lies in the theoretical side, we do not include extensive studies on all recent algorithms and can only conduct experiments with some of them. We examine the following six algorithms: DrQ \shortcite{Yarats2021ReinforcementLW}, PAD \shortcite{hansen2021selfsupervised}, CURL \shortcite{Srinivas2020CURLCU} SODA \shortcite{Hansen2020GeneralizationIR}, SVEA \shortcite{Hansen2021StabilizingDQ}, and PIE-G \shortcite{Yuan2022PreTrainedIE}. These methods are chosen because of the fact that they typically represent different categories of methods that handling the generalization problems in visual RL, and that their reported results can be reproduced. DrQ explores the data augmentation trick to enhance visual RL training; PAD utilizes a self-supervised objective to adapt to the testing environment during deployment; CURL leverages the contrastive learning objective to capture better representations; SODA reformulates the problem of generalization as a representational consistency learning problem and enables the encoder to map different views of the same state to similar representations; SVEA uses strong data augmentation (e.g., augmentation with random convolution networks) and regularizes the way $Q$ value updates; PIE-G is built upon DrQ-v2 \shortcite{yarats2022mastering} while replaces its encoder with the pre-trained image encoder.

It is known that DrQ often fails when generalizing to unseen test environments, while PIE-G and SVEA are among the strongest generalization visual RL algorithms. In order to show that our theory holds in practice, we expect that the representation deviation $\|\phi(f(s))-\phi(s)\|$ (as well as the policy deviation $\|\pi(\phi(f(s))) - \pi(\phi(s))\|$) of a stronger generalization algorithm ought to be smaller than weak ones. We verify this by conducting experiments on two environments from DMC-GB, \texttt{walker-walk} and \texttt{finger-spin} (results on wider environments can be found in Appendix \ref{sec:broaderexperiments}). We note that the policy deviation counts when comparing different generalization visual RL algorithms because different algorithms train their policies in different ways, leading to different $L_{\pi_1}$ that may affect the term $\frac{L_{r_2}L_{\pi_1}\varrho}{1-\gamma}(1-\gamma^{T+1})$. The policy deviation is an important metric that directly indicate \emph{how well the agent can recover its behavior learned in the clean train environment when deployed in the test environment with distractors}. We run these algorithms under their default hyperparameters on the clean training environment first and then replace the background with playing videos or change the color of the agent (e.g., \emph{video-easy} setting in DMC-GB contains 10 different video backgrounds). We adopt three generalization scenarios for empirical evaluation: color-hard, video-easy, and video-hard. Our experimental setting is, the trajectory remains the same, and only backgrounds or the colors are changed. This generally meets our formulation, i.e., we have $s$ in the training environment while also needing $f(s)$ in the testing environment, upon the identical state $s$. We then evaluate the representation deviation using the learned encoder and the policy deviation with the policy network of each algorithm on the clean training trajectory and the testing trajectories with distractors for 100 episodes and 5 different random seeds.

\begin{figure*}[!htb]
    \centering
    \includegraphics[width=0.44\linewidth]{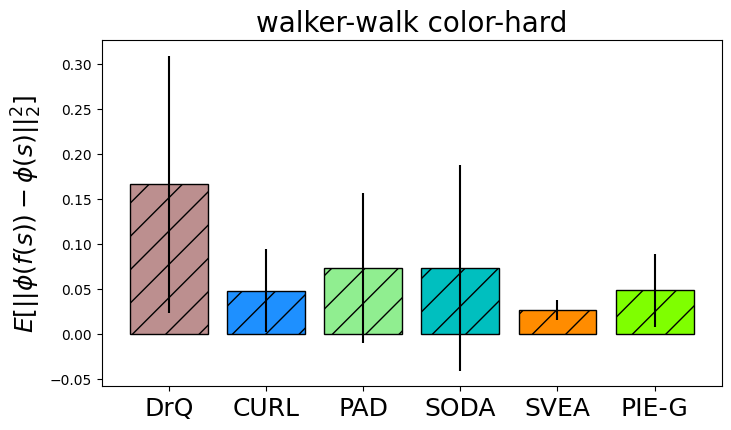}\hspace{-1mm}
    \includegraphics[width=0.44\linewidth]{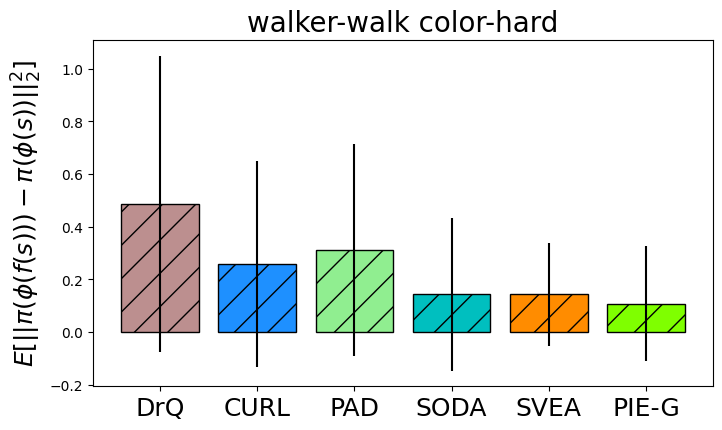}\hspace{-1mm}
    \includegraphics[width=0.44\linewidth]{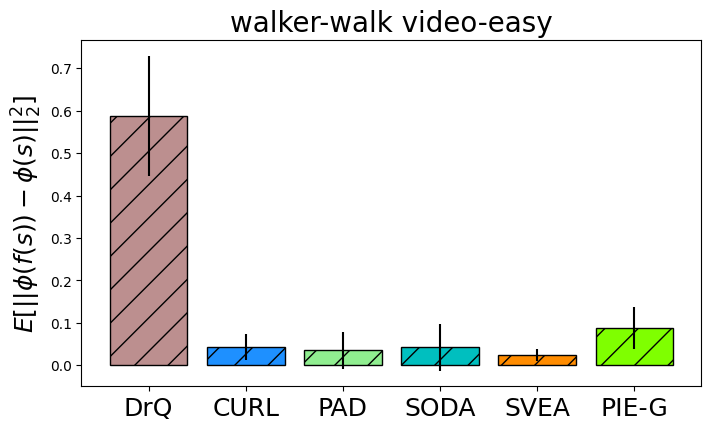}\hspace{-1mm}
    \includegraphics[width=0.44\linewidth]{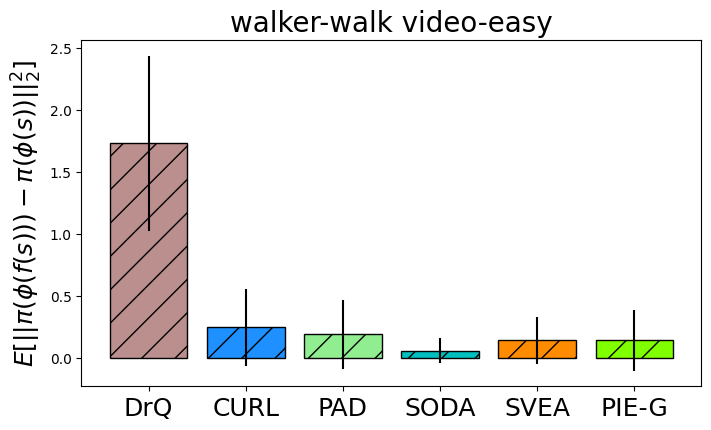}\hspace{-1mm}
    \includegraphics[width=0.44\linewidth]{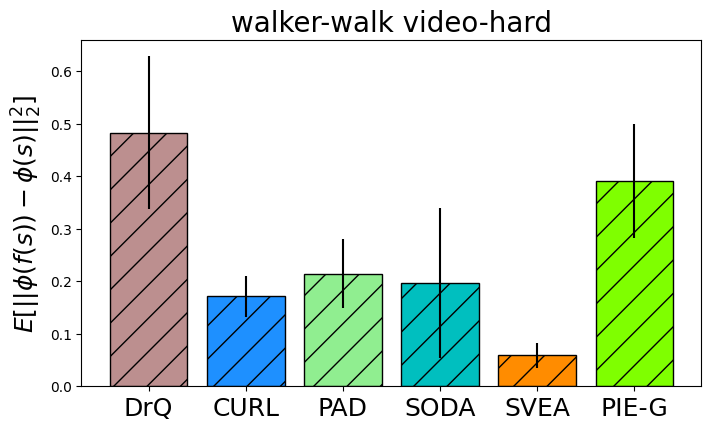}\hspace{-1mm}
    \includegraphics[width=0.44\linewidth]{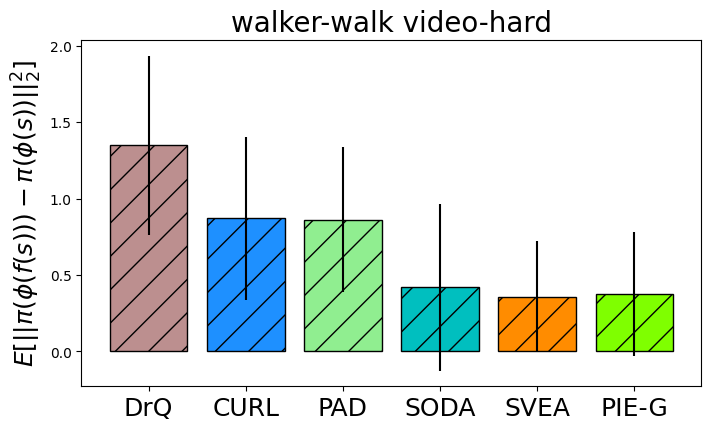}
    \caption{\textbf{Evidence that our theoretical results can explain empirical algorithms.} We present comparison of average representation deviation ($\mathbb{E}[\|\phi(s) - \phi(s^\prime)\|_2^2]$, left column) and average policy deviation ($\mathbb{E}[\|\pi(\phi(s)) - \pi(\phi(s^\prime))\|_2^2]$, right column) of 6 typical methods on color-hard, video-easy, and video-hard settings of walker-walk task from DMC-GB. The results are averaged over the trajectory and across 5 varied random seeds. The error bar denotes the average standard deviation along the trajectory and 5 seeds.}
    \label{fig:barplotwalkerwalk}
\end{figure*}

\begin{figure*}[!htb]
    \centering
    \includegraphics[width=0.44\linewidth]{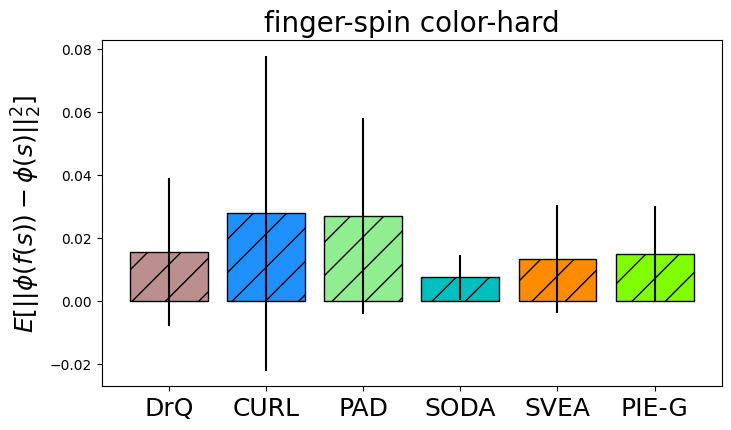}\hspace{-1mm}
    \includegraphics[width=0.44\linewidth]{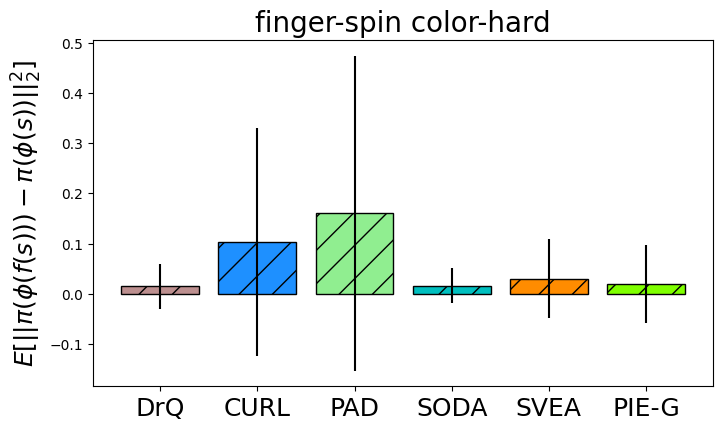}\hspace{-1mm}
    \includegraphics[width=0.44\linewidth]{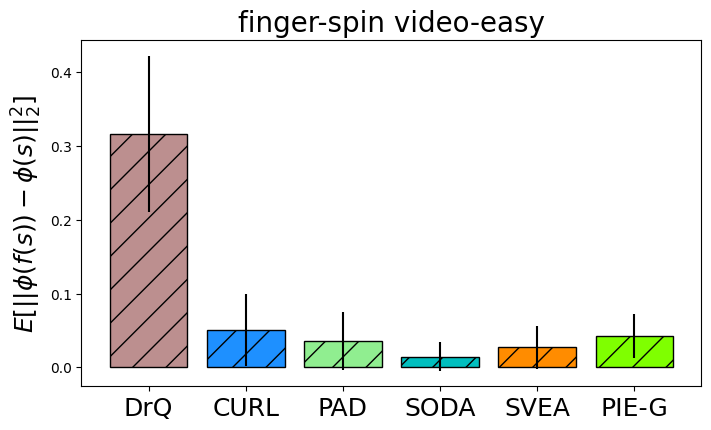}\hspace{-1mm}
    \includegraphics[width=0.44\linewidth]{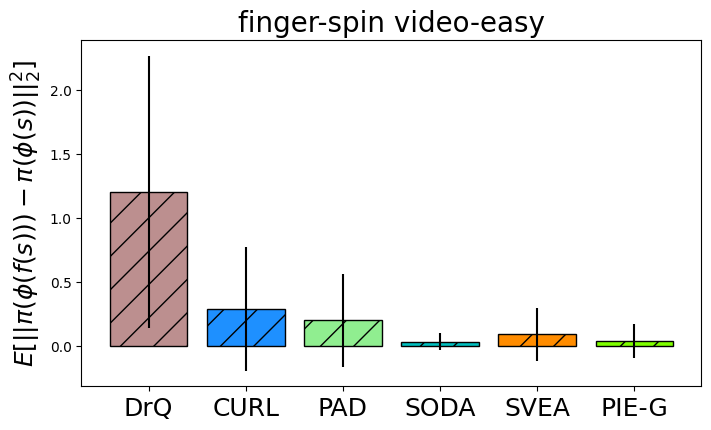}\hspace{-1mm}
    \includegraphics[width=0.44\linewidth]{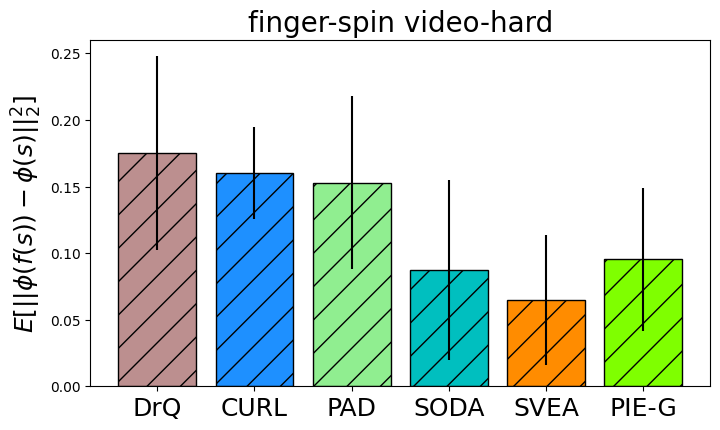}\hspace{-1mm}
    \includegraphics[width=0.44\linewidth]{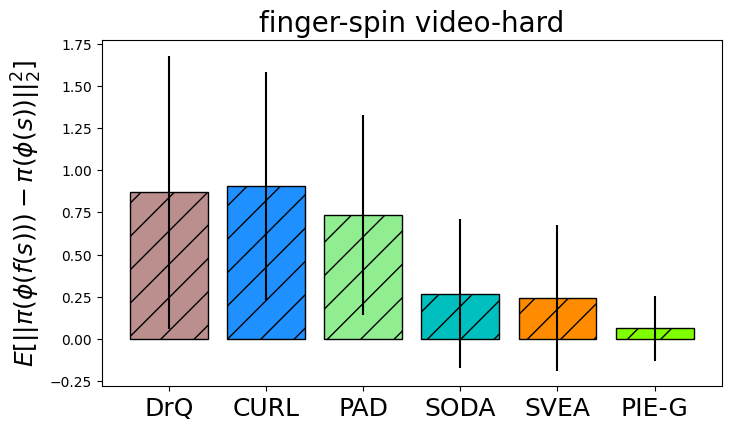}
    \caption{\textbf{Evidence that our theoretical results can explain empirical algorithms.} We demonstrate comparison of average representation deviation ($\mathbb{E}[\|\phi(s) - \phi(s^\prime)\|_2^2]$, left column) and average policy deviation ($\mathbb{E}[\|\pi(\phi(s)) - \pi(\phi(s^\prime))\|_2^2]$, right column) of 6 typical methods on color-hard, video-easy, and video-hard settings of finger-spin task from DMC-GB. The results are averaged over the trajectory and 5 different random seeds. The error bar represents the average standard deviation along the trajectory and 5 seeds.}
    \label{fig:barplotfingerspin}
\end{figure*}

All algorithms are run for 500K environmental steps. The evaluated trajectories are made up of 1000 transitions. We measure the average representation deviation $\mathbb{E}_{s\sim\tau}[\|\phi(f(s))-\phi(s)\|]$ and the policy deviation $\mathbb{E}_{s\sim\tau}[\|\pi(\phi(f(s))) - \pi(\phi(s))\|]$ along the trajectory. We summarize the results of walker-walk task in Figure \ref{fig:barplotwalkerwalk} and the results of finger-spin task in Figure \ref{fig:barplotfingerspin}. 

It turns out that DrQ generally has the largest policy deviation and representation deviation among all of the methods (and its generalization performance is generally the worst). The representation deviation and policy deviation of CURL and PAD are generally smaller than DrQ (but the deviations are larger than DrQ on a limited number of tasks). Based on \shortcite{hansen2021selfsupervised,Yuan2022PreTrainedIE}, CURL often does not exhibit satisfying generalization performance and we observe that its policy deviation and representation deviation are much larger than methods like SODA and SVEA. PAD also underperforms methods like SVEA and PIE-G in many generalization tasks. It is understandable because many studies \shortcite{jaderberg2017reinforcement,hansen2021selfsupervised,Lin2019AdaptiveAT,Lyle2021OnTE} have shown that the best choice of auxiliary tasks in self-supervision methods highly depends on the specific RL task, and negative impacts can incur if we choose sub-optimal auxiliary tasks. The strong generalization visual RL algorithms like SODA, SVEA and PIE-G have small policy deviation and representation deviation on most of the tasks. Notably, the standard deviation of DrQ is also the largest, while algorithms like SODA, SVEA and PIE-G enjoy much smaller standard deviations, indicating that they are more stable than DrQ. The empirical results are unanimously in line with our expectations. Hence, we believe our theory explains in part why these algorithms work in practice.

For readers of interest, we include the results of average policy deviation and representation deviation along the trajectory (that is, the plots of average policy deviation and representation deviation against the steps) on some tasks in Appendix \ref{sec:appmissingalongthetrajectory}, where we find that methods like DrQ has large representation deviation and policy deviation on almost every transition in the trajectory while strong generalization methods like SVEA and PIE-G enjoy much smaller deviations. We believe these can further validate the rationality of our theoretical analysis.

\section{Discussions}
\label{sec:discussion}
In this section, we discuss some issues that readers of interest may wonder, e.g., guidance on improving test performance. Furthermore, we discuss some possibly promising ways for developing future advanced generalization visual RL algorithms and the theoretical bounds when different reward functions are considered in the training and testing environments.

\textbf{How reparameterization can be applied in RL?} Our core contribution in this paper is the construction of theoretical bounds on the generalization gap in visual RL, via the reparameterization trick. As we highlight in Section \ref{sec:relatedwork}, the community has already witnessed many trials in terms of leveraging reparameterization in RL, with some of them being quite popular and widely used (e.g., the SAC algorithm). Apart from the aforementioned applications of reparameterization, one can employ it in linear quadratic regulator (LQR) \shortcite{Bradtke1992ReinforcementLA,Cao2010OptimalLA,Simchowitz2020NaiveEI,recht2019tour}. For instance, \shortciteA{jost2017accelerating} reparameterize the input variables with an LQR controller to accelerate linear model predictive control. In practice, the reparameterization can also be applied in the (real-world) physical systems where the dynamics are formulated by stochastic partial differential equations that contain reparameterizable parameters or components over continuous state-action space \shortcite{Heess2015LearningCC,Perkins2001LyapunovConstrainedAS}, as we do in Algorithm \ref{alg:reparameterizablerl}. 

\textbf{How to improve test performance?} Our theoretical result in Theorem \ref{theo:final} indicates that if one wants good performance in the testing environment, then one should (1) minimize the representation deviation between training and testing environments, such that the generalization gap is small; (2) improve the performance in the training environment, since only if the agent achieves good performance during training (e.g., collected $n$ trajectories are all of the high return) can it possibly acquire satisfying performance in the testing environment. The second part is closely correlated to the \textit{sample efficiency} in visual RL. Though many works try to attain some progress, including model-based \shortcite{Hansen2022TemporalDL,Ye2021MasteringAG} and model-free \shortcite{yarats2022mastering,Hansen2022TemporalDL,lyu2024off} methods, it still remains a central challenge.

\textbf{Possible directions of developing RL algorithms for generalization with image input.} We believe the most critical issue for enhancing generalization in visual RL is how to get robust and generalizable representations. Studying how and when the image representations can generalize to unseen scenarios both empirically and theoretically may be vital, and \shortcite{lan22onthegeneralization} serves as a good primary work. Extracting important regions in the visual input to dismiss the influence of distractors from the testing environment is also promising. \shortciteA{Bertoin2022LookWY} realize it by building an attribution map. Moreover, this can be done with the aid of foundation models, e.g., SAM \shortcite{Kirillov2023SegmentA}. It is straightforward that the generalization performance can be improved if the influence of distractors in the testing environments is cleared. It is equivalent to directly minimizing $\|f(s)-s\|$ (our theories still apply here). Meanwhile, it is interesting to see whether large language models like ChatGPT can aid generalization in visual RL, e.g., ask the ChatGPT to summarize the image and tell us what action we should take, such that we use the text embedding as an extra input in the policy. By asking large language models (LLM), one may extract some key features that are not affected by the distractors. For example, the LLM may summarize that ``there is a running robot in the figure", and this may be vital for the agent to execute suitable actions. By incorporating such information as an additional input, we may minimize the representation deviation between the training and testing environments. From another perspective, enhancing the robustness of the policy is promising and encouraging for improving the generalization ability, i.e., controlling $L_{\pi_1},L_{\pi_2}$, and we can borrow some ideas from the field of robust RL \shortcite{morimoto2001robust,Tessler2019ActionRR,zhang2021robust,derman2020bayesian}.

\textbf{More general theoretical bounds when the reward functions are different.} In our theoretical analysis, we assume that the training environment and the testing environment can have slight different transition dynamics and initialization function, but with the same reward function. It is then interesting to investigate the generalization gap of visual RL agent when different reward functions are adopted in training and testing environments. We denote the reward function in the training environment as $r(s,a)$ and the reward function in the testing environment as $r^\prime(s,a)$. We denote the return that leverages the testing rewards $r^\prime$ as $J^\prime$, i.e., $\mathbb{E}_{\tau}[J^\prime(\phi(f(\tau)))]=\mathbb{E}_{\tau}\left[\sum_{t=0}^T\gamma^t r^\prime(s_t, \pi(\phi(f(s_t))))\right]$. Then with similar assumptions and some additional assumption on the rewards in the training and testing environments, we can derive a similar generalization bound as Theorem \ref{theo:final}, where our core insight in the previous analysis still holds, i.e., the core factor that impact the generalization error is the representation deviation of the state before and after adding the distractors.

\begin{theorem}
\label{theo:rewarddifferent}
    Suppose that the assumptions made in Theorem \ref{theo:generalizationerror} and Lemma \ref{lemma:rademacherbound} hold. We suppose the testing environment has a different reward function $r^\prime(s,a)$ from the training environment. We assume $|r(s,a) - r^\prime(s,a)|\le \epsilon_r< \infty, \epsilon_r\in\mathbb{R},\forall s,a$, i.e., given any state and action, the rewards in the training environment and the testing environments are bounded. Then we have with probability at least $1-\delta$, the generalization error gives,
    \small
    \begin{align*}
        &\left\|\mathbb{E}_\xi \left[J^\prime\left(\phi\left(f\left(\tau\left(\xi; \pi, \mathcal{T}^\prime,\mathcal{I}^\prime\right)\right)\right)\right)\right] - \dfrac{1}{n} \sum_{i=1}^n J(\phi(\tau(\xi_i;\pi,\mathcal{T},\mathcal{I}))) \right\| \le \lambda \zeta \sum_{t=0}^T \gamma^t \dfrac{\nu^t-1}{\nu-1} + \lambda\epsilon\sum_{t=0}^T \gamma^t \nu^t \\
        & + \dfrac{1-\gamma^{T+1}}{1-\gamma}(L_{r_2}L_{\pi_1} \varrho + \epsilon_r) +\mathcal{O}\left( L_J K \sqrt{\dfrac{m}{n}} \right) + \mathcal{O}\left(r_{\rm max} \sqrt{\dfrac{\log (1/\delta)}{n}} \right).
    \end{align*}
\end{theorem}

\begin{proof}
    We decompose the target as follows:
    \begin{align*}
        &\left\|\mathbb{E}_\xi \left[J^\prime\left(\phi\left(f\left(\tau\left(\xi; \pi, \mathcal{T}^\prime,\mathcal{I}^\prime\right)\right)\right)\right)\right] - \dfrac{1}{n} \sum_{i=1}^n J(\phi(\tau(\xi_i;\pi,\mathcal{T},\mathcal{I}))) \right\| \\
        &\le \underbrace{\left\|\mathbb{E}_\xi \left[J^\prime\left(\phi\left(f\left(\tau\left(\xi; \pi, \mathcal{T}^\prime,\mathcal{I}^\prime\right)\right)\right)\right)\right] - \mathbb{E}_\xi \left[J\left(\phi\left(f\left(\tau\left(\xi; \pi, \mathcal{T}^\prime,\mathcal{I}^\prime\right)\right)\right)\right)\right] \right\|}_{\rm (I)} \\
        &\qquad\qquad + \underbrace{\left\| \mathbb{E}_\xi \left[J\left(\phi\left( f\left(\tau\left(\xi; \pi, \mathcal{T}^\prime,\mathcal{I}^\prime\right)\right)\right)\right)\right] - \dfrac{1}{n} \sum_{i=1}^n J(\phi(\tau(\xi_i;\pi,\mathcal{T},\mathcal{I}))) \right\|}_{\rm (II)}.
    \end{align*}
    For term (I), we have
    \begin{align*}
        {\rm (I)} &= \left\|\mathbb{E}_\xi \left[J^\prime\left(\phi\left(f\left(\tau\left(\xi; \pi, \mathcal{T}^\prime,\mathcal{I}^\prime\right)\right)\right)\right)\right] - \mathbb{E}_\xi \left[J\left(\phi\left(f\left(\tau\left(\xi; \pi, \mathcal{T}^\prime,\mathcal{I}^\prime\right)\right)\right)\right)\right] \right\| \\
        &\le \int_\xi\left\|  \sum_{t=0}^T \gamma^t\left( r^\prime(s_t,\pi(\phi(f(s_t)))) - r(s_t,\pi(\phi(s_t))) \right) \right\|d\xi \\
        &\le \int_\xi\sum_{t=0}^T \gamma^t\epsilon_r = \dfrac{1-\gamma^{T+1}}{1-\gamma}\epsilon_r.
    \end{align*}
    For term (II), we directly bound it with Theorem \ref{theo:final}, and then the conclusion holds accordingly.
\end{proof}

\section{Conclusions and Limitations}
Despite the emergence of many practical and promising algorithms for enhancing the generalization capability of visual RL policies, a clear and instructive theoretical analysis on the generalization gap, and how to minimize the generalization gap are absent. The main purpose of this work is to provide a theoretical bound on the generalization gap in visual RL when there exist distractors in the testing environment, and try to explain why previous methods work based on the derived bounds. However, directly analyzing the generalization gap is difficult since the policy keeps evolving. We isolate the randomness from the policy by resorting to the reparameterization trick. Our bound indicates that the key to reducing the generalization gap is to minimize the representation deviation between the training and testing environments. We further provide empirical evidence on the validity of the assumptions and conclusion, which we find is consistent with the theoretical results. An interesting future work is to study the generalization gap of off-policy visual RL and characterize what matters under this setting, e.g., derive the generalization gap bound of \emph{offline} visual RL algorithms.

We believe the limitations of this work mainly lie in the following aspects: (1) the reparameterization trick cannot be experimentally applied in the simulated environments, e.g., DMC-GB, and some real-world scenarios, because they often do not meet the reparameterzable conditions; (2) some of the assumptions we made (e.g., the smoothness in dynamics) may not necessarily hold in general cases.

\acks{Work done while Jiafei Lyu was an intern at Tencent IEG. This work was supported by the STI 2030-Major Projects under Grant 2021ZD0201404. The authors would like to thank the anonymous reviewers for their valuable comments and suggestions.
}

\appendix

\section{Missing Example Trajectories on DMC-GB}
\label{sec:exampletraj}
In this section, we provide the example trajectories of the visual RL agent in the training environment, and the corresponding trajectories of DrQ, SVEA, and PIE-G in the testing environment. We choose walker-walk video-easy task as the testbed.

\begin{figure*}[htb]
    \centering
    \includegraphics[width=0.09\linewidth]{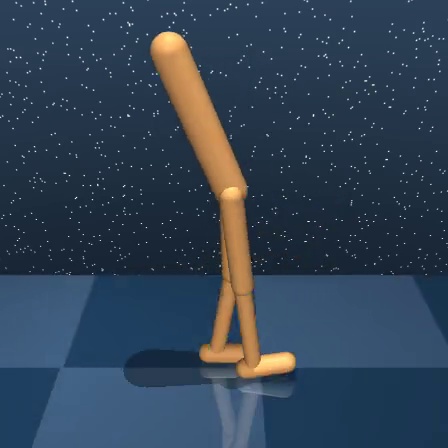}\hspace{-1mm}
    \includegraphics[width=0.09\linewidth]{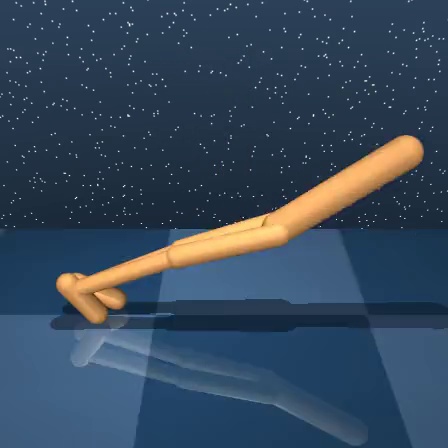}\hspace{-1mm}
    \includegraphics[width=0.09\linewidth]{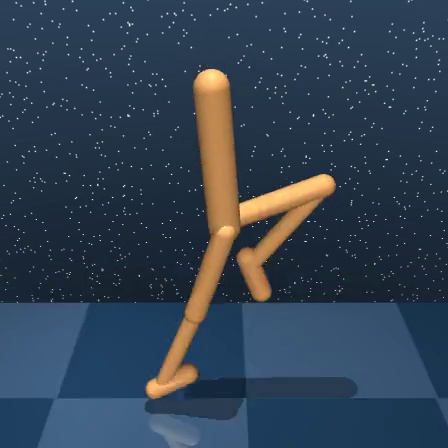}\hspace{-1mm}
    \includegraphics[width=0.09\linewidth]{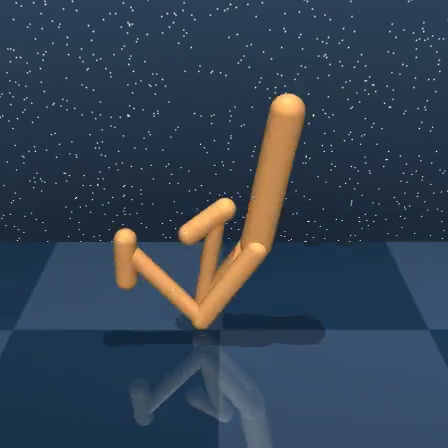}\hspace{-1mm}
    \includegraphics[width=0.09\linewidth]{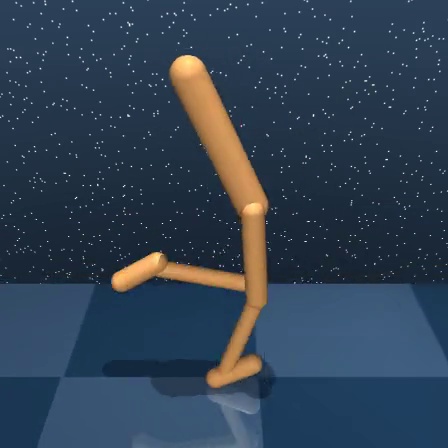}\hspace{-1mm}
    \includegraphics[width=0.09\linewidth]{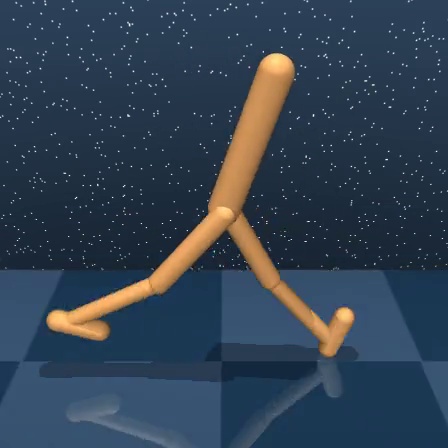}\hspace{-1mm}
    \includegraphics[width=0.09\linewidth]{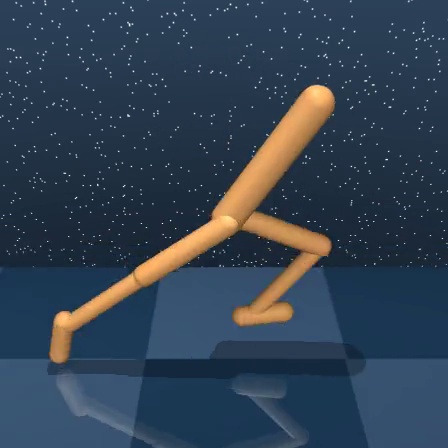}\hspace{-1mm}
    \includegraphics[width=0.09\linewidth]{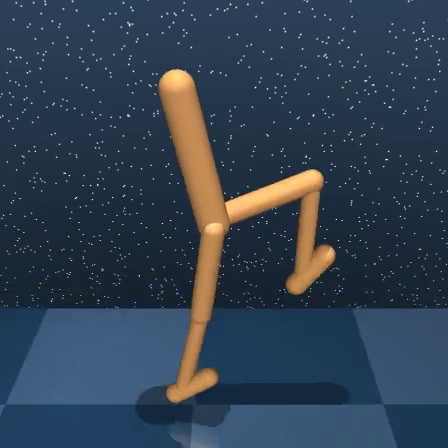}\hspace{-1mm}
    \includegraphics[width=0.09\linewidth]{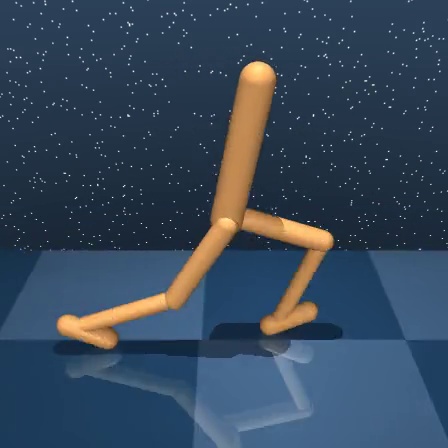}\hspace{-1mm}
    \includegraphics[width=0.09\linewidth]{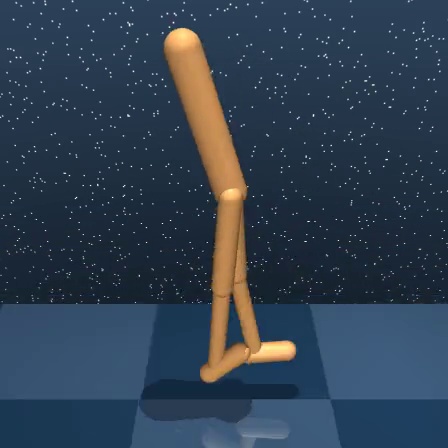}\hspace{-1mm}
    \includegraphics[width=0.09\linewidth]{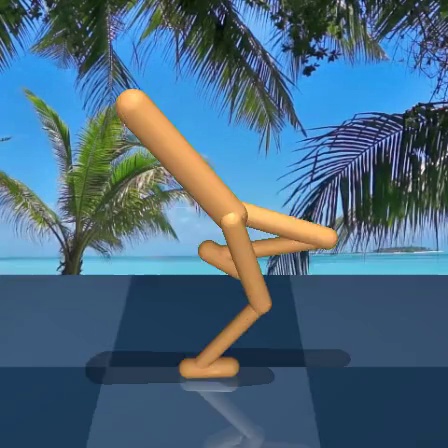}\hspace{-1mm}
    \includegraphics[width=0.09\linewidth]{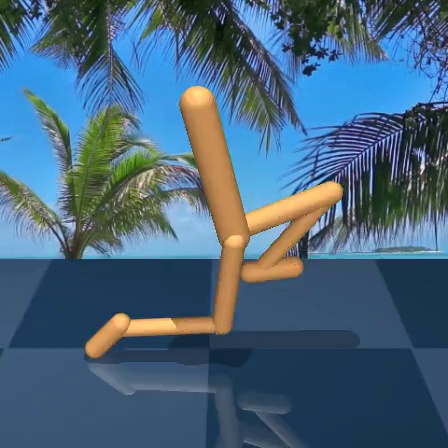}\hspace{-1mm}
    \includegraphics[width=0.09\linewidth]{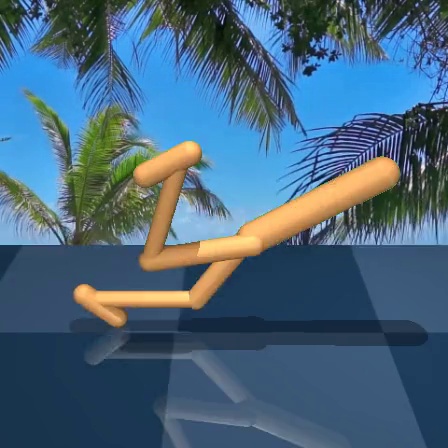}\hspace{-1mm}
    \includegraphics[width=0.09\linewidth]{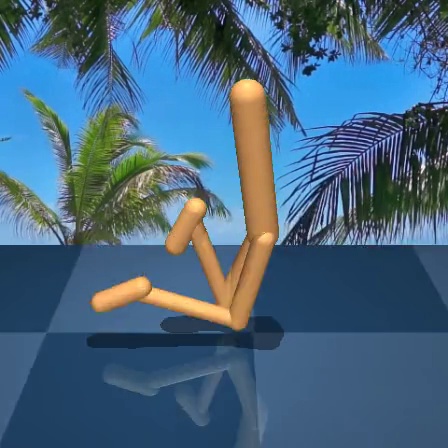}\hspace{-1mm}
    \includegraphics[width=0.09\linewidth]{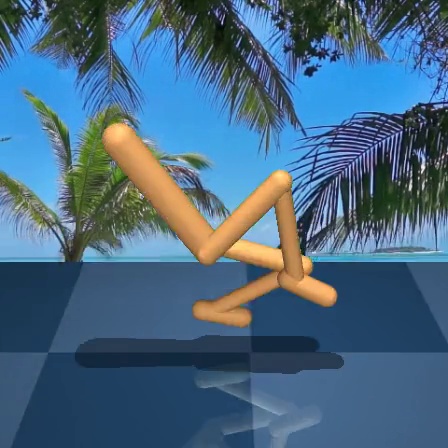}\hspace{-1mm}
    \includegraphics[width=0.09\linewidth]{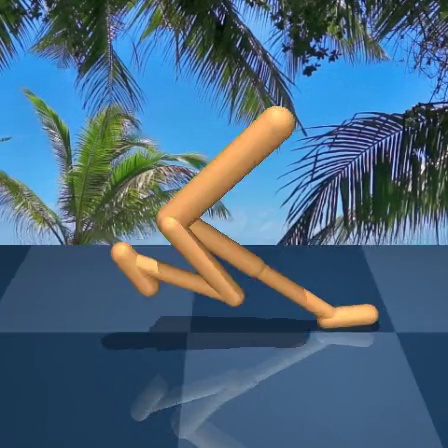}\hspace{-1mm}
    \includegraphics[width=0.09\linewidth]{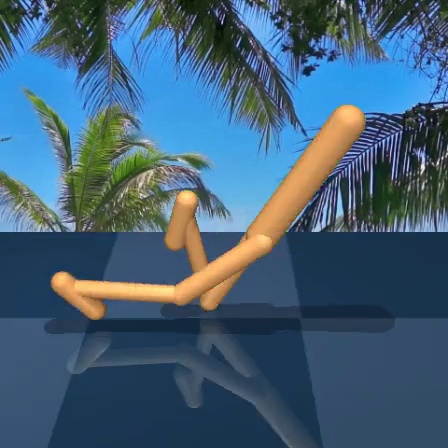}\hspace{-1mm}
    \includegraphics[width=0.09\linewidth]{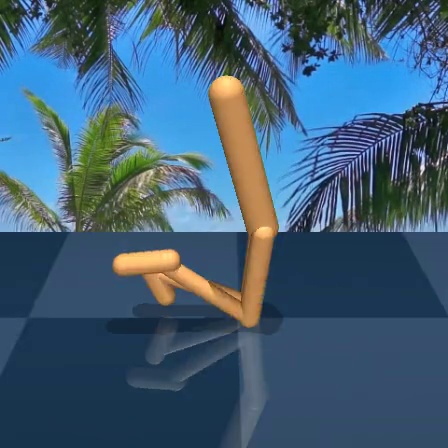}\hspace{-1mm}
    \includegraphics[width=0.09\linewidth]{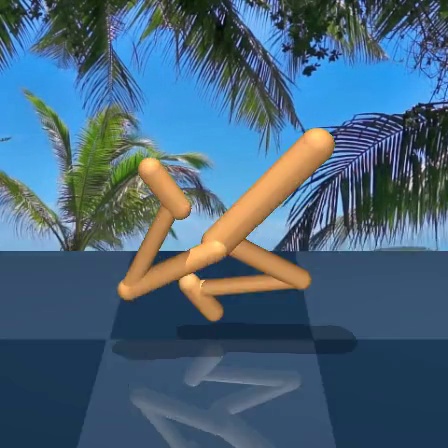}\hspace{-1mm}
    \includegraphics[width=0.09\linewidth]{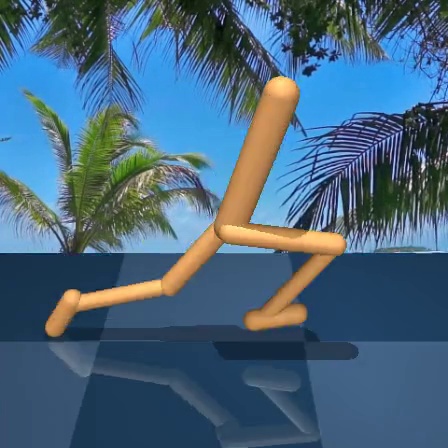}\hspace{-1mm}
    \includegraphics[width=0.09\linewidth]{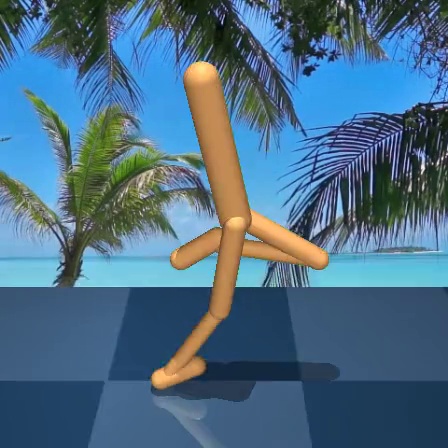}\hspace{-1mm}
    \includegraphics[width=0.09\linewidth]{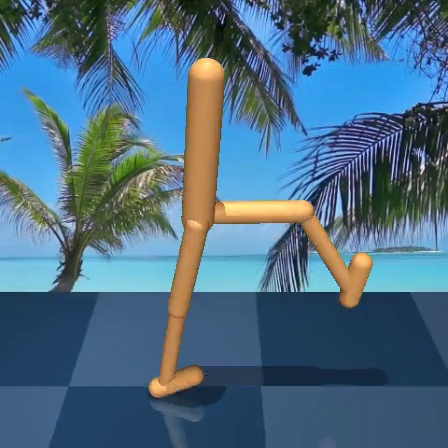}\hspace{-1mm}
    \includegraphics[width=0.09\linewidth]{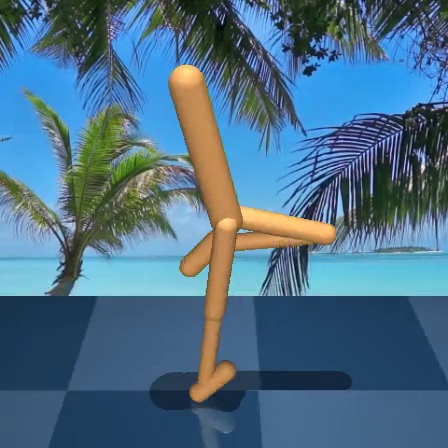}\hspace{-1mm}
    \includegraphics[width=0.09\linewidth]{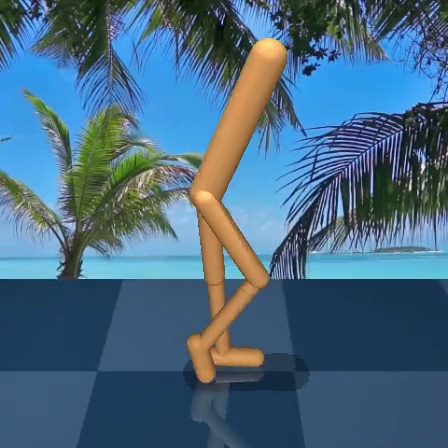}\hspace{-1mm}
    \includegraphics[width=0.09\linewidth]{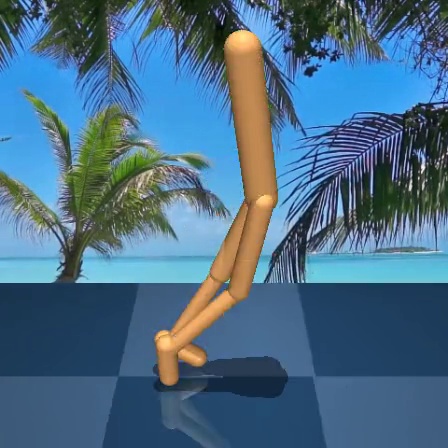}\hspace{-1mm}
    \includegraphics[width=0.09\linewidth]{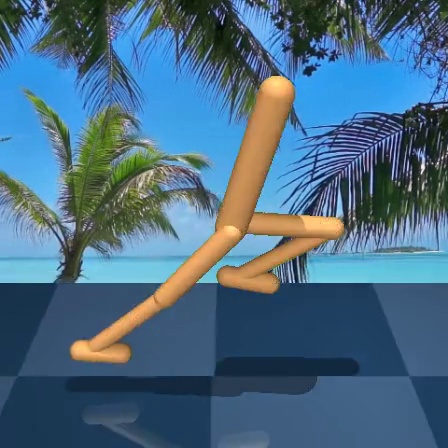}\hspace{-1mm}
    \includegraphics[width=0.09\linewidth]{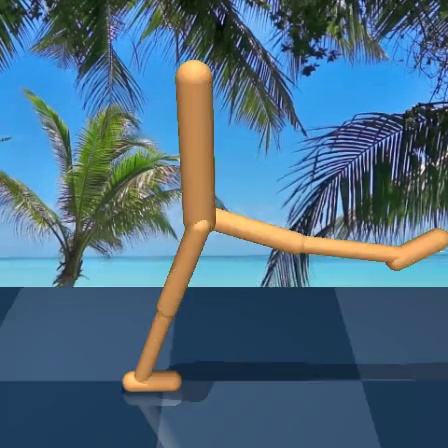}\hspace{-1mm}
    \includegraphics[width=0.09\linewidth]{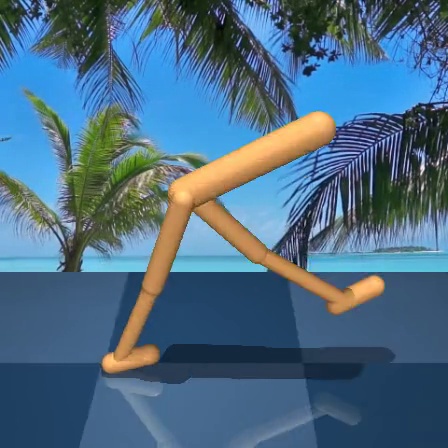}\hspace{-1mm}
    \includegraphics[width=0.09\linewidth]{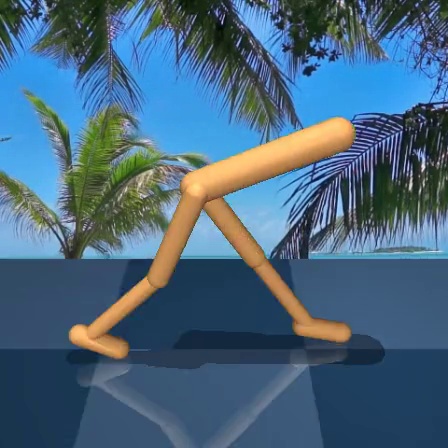}\hspace{-1mm}
    \includegraphics[width=0.09\linewidth]{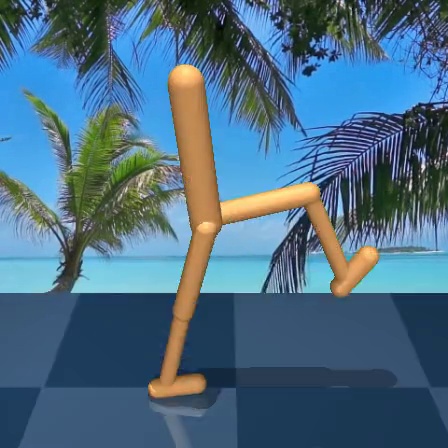}\hspace{-1mm}
    \includegraphics[width=0.09\linewidth]{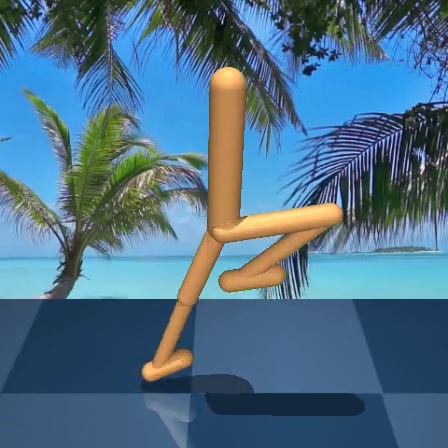}\hspace{-1mm}
    \includegraphics[width=0.09\linewidth]{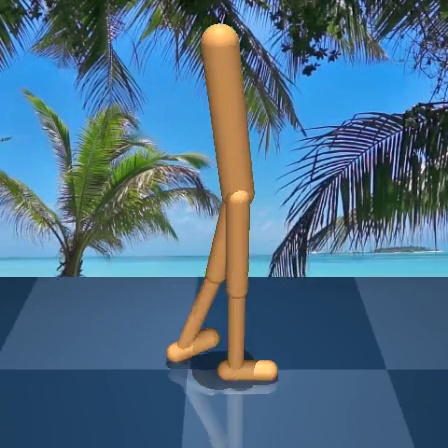}\hspace{-1mm}
    \includegraphics[width=0.09\linewidth]{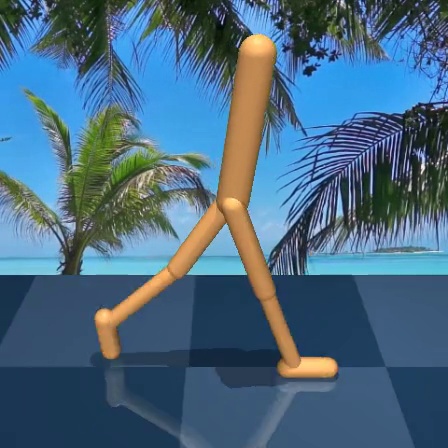}\hspace{-1mm}
    \includegraphics[width=0.09\linewidth]{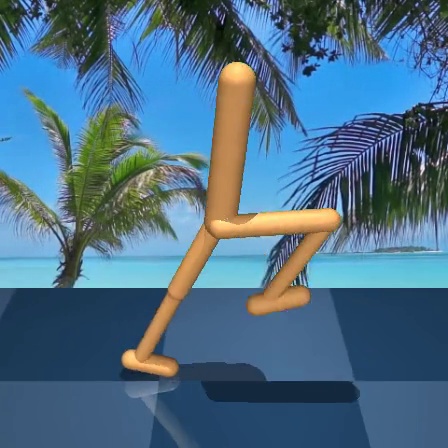}\hspace{-1mm}
    \includegraphics[width=0.09\linewidth]{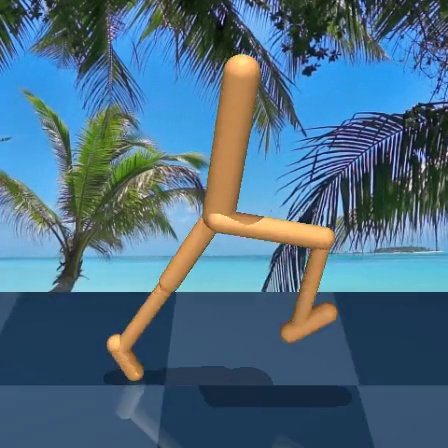}\hspace{-1mm}
    \includegraphics[width=0.09\linewidth]{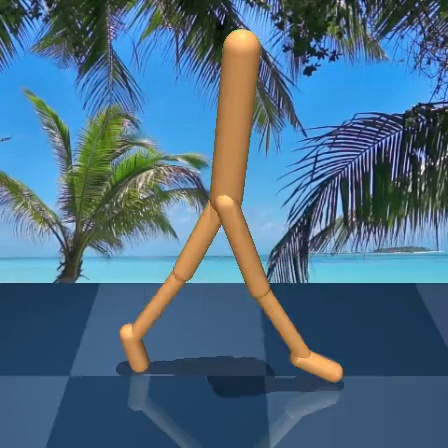}\hspace{-1mm}
    \includegraphics[width=0.09\linewidth]{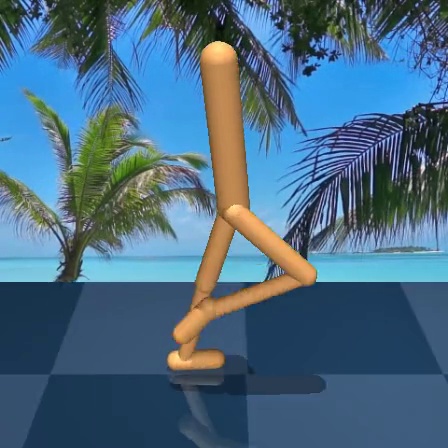}\hspace{-1mm}
    \includegraphics[width=0.09\linewidth]{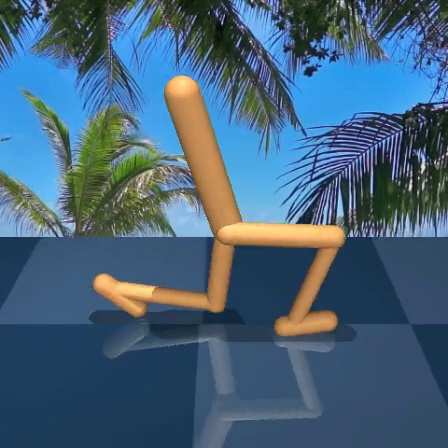}\hspace{-1mm}
    \includegraphics[width=0.09\linewidth]{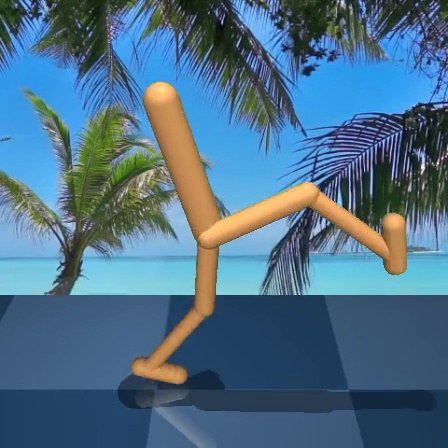}\hspace{-1mm}
    \includegraphics[width=0.09\linewidth]{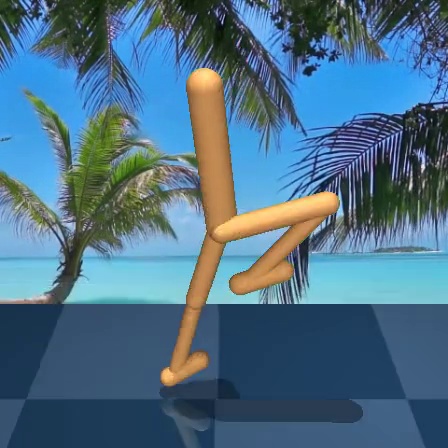}\hspace{-1mm}
    \caption{Example trajectories of the training environment (first row), and DrQ (second row), SVEA (third row), and PIE-G (fourth row) deployed in the walker-walk \emph{video-easy} task with distractors. Results are illustrated by using the models of each algorithm after training 500K environmental steps.}
    \label{fig:trainexample}
\end{figure*}

\section{Broader Experiments}
\label{sec:broaderexperiments}
In this section, we provide wider empirical results to further validate our theoretical results. Our experiments are conducted on DMC-GB, and we consider three settings: \textit{color-hard}, \textit{video-easy}, and \textit{video-hard}. The color-hard tasks require the agent to generalize to color-jittered observations. Video-easy and video-hard tasks, instead, replace the background of the agent with the playing videos. Among them, video-hard tasks are the most challenging, since they contain more complex and fast-switching videos (up to 100 videos), and the reference plane of the ground is also removed. We detail the examples of the training and testing environments in DMC-GB below in Figure \ref{fig:example}.

\begin{figure*}[h]
    \centering
    \subfigure[training]{
    \label{fig:train}
    \includegraphics[scale=0.1]{figures/walker_walk_drq_train_100.jpg}
    }\hspace{3mm}
    \subfigure[color-hard]{
    \label{fig:colorhard}
    \includegraphics[scale=0.1]{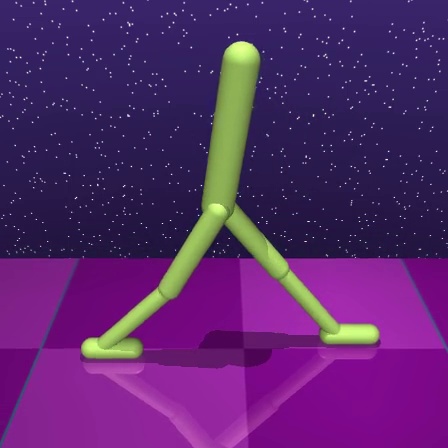}
    \includegraphics[scale=0.1]{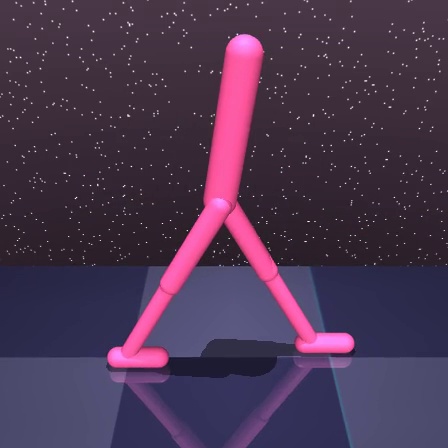}
    }\hspace{3mm}
    \subfigure[video-easy]{
    \label{fig:videoeasy}
    \includegraphics[scale=0.1]{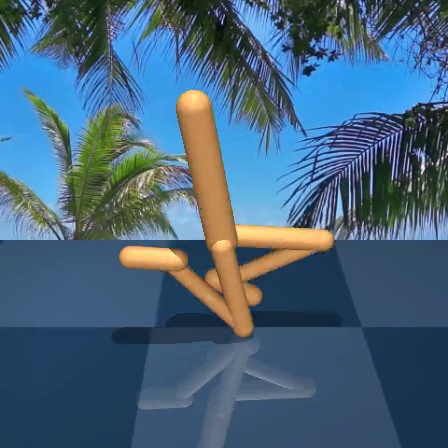}
    \includegraphics[scale=0.1]{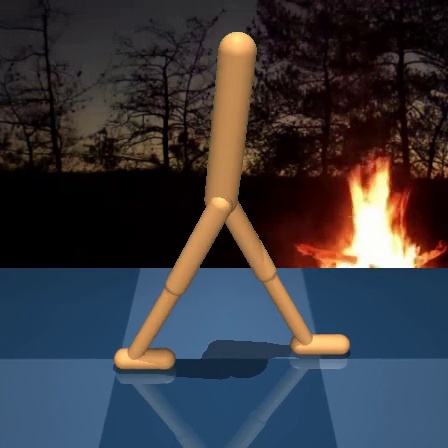}
    }\hspace{3mm}
    \subfigure[video-hard]{
    \label{fig:videohard}
    \includegraphics[scale=0.1]{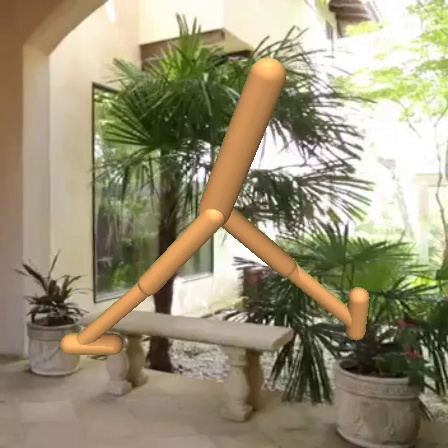}
    \includegraphics[scale=0.1]{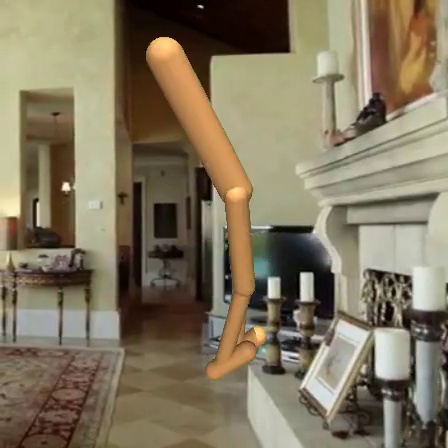}
    }\hspace{3mm}
    \caption{Examples of the clean training environment and testing environments with randomly jittered color and moving video backgrounds in DMC-GB.}
    \label{fig:example}
\end{figure*}

We adopt the implementation of DrQ and SVEA from the DMC-GB \footnote{https://github.com/nicklashansen/dmcontrol-generalization-benchmark} under MIT License, and adopt the implementation of PIE-G from the open-sourced codebase from the authors \footnote{https://anonymous.4open.science/r/PIE-G-EF75/README.md}. We use the default hyperparameters of these algorithms. We evaluate each algorithm across 100 episodes and 5 different random seeds (1-5) for all of the generalization settings. Note that we use SVEA(conv), which leverages a random convolution network for data augmentation, instead of SVEA(overlay), as SVEA(overlay) adopts an outer source of dataset for training and data augmentation, which we think somewhat has an overlap with PIE-G. In practice, SVEA(conv) exhibits similar performance compared with that of SVEA(overlay). The experiments are conducted with PyTorch 1.8 version.

\begin{figure*}[!htb]
    \centering
    \includegraphics[width=0.45\linewidth]{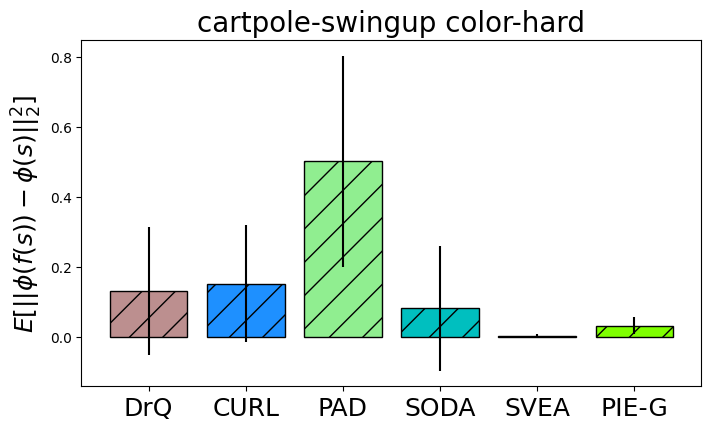}\hspace{-1mm}
    \includegraphics[width=0.45\linewidth]{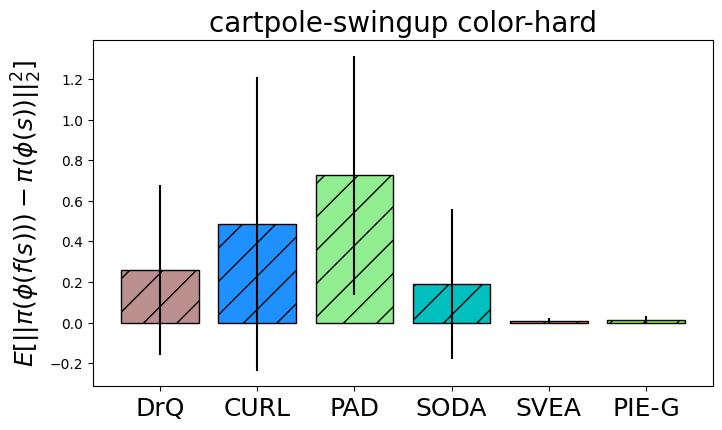}\hspace{-1mm}
    \includegraphics[width=0.45\linewidth]{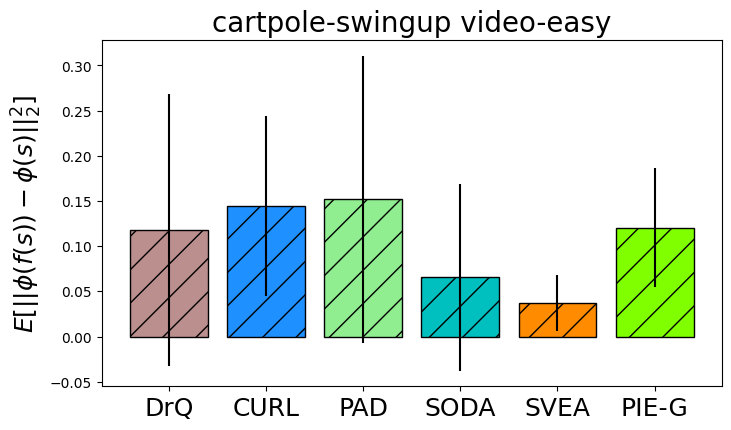}\hspace{-1mm}
    \includegraphics[width=0.45\linewidth]{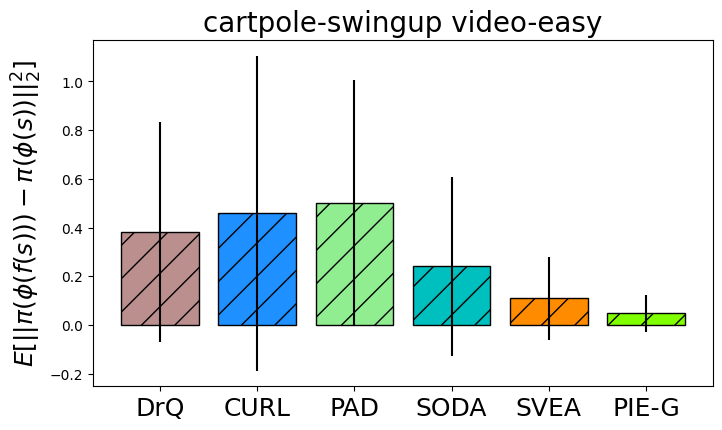}\hspace{-1mm}
    \includegraphics[width=0.45\linewidth]{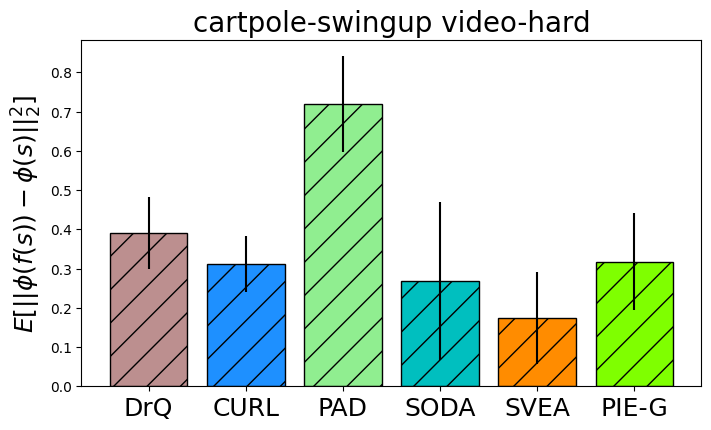}\hspace{-1mm}
    \includegraphics[width=0.45\linewidth]{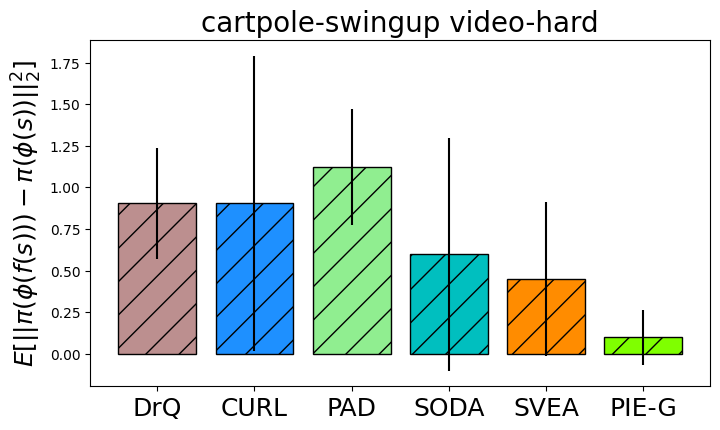}
    \caption{\textbf{Evidence that our theoretical results can explain empirical algorithms.} We present comparison of average representation deviation ($\mathbb{E}[\|\phi(s) - \phi(s^\prime)\|_2^2]$, left column) and average policy deviation ($\mathbb{E}[\|\pi(\phi(s)) - \pi(\phi(s^\prime))\|_2^2]$, right column) of 6 typical methods on color-hard, video-easy, and video-hard settings of cartpole-swingup task from DMC-GB. The results are averaged over the trajectory and across 5 varied random seeds. The error bar denotes the average standard deviation along the trajectory and 5 seeds.}
    \label{fig:barplotcartpoleswingup}
\end{figure*}

In the main text, we mainly showcase the policy deviation and representation deviation of DrQ, CURL, PAD, SODA, SVEA, and PIE-G on the color-hard, video-easy, and video-hard settings of \texttt{walker-walk} and \texttt{finger-spin} tasks. We further present the results on two additional tasks here, \texttt{cartpole-swingup} and \texttt{walker-stand}. It is worth noting that our comparison is valid and reasonable since the sample efficiency and the final performance of these algorithms in the training environments are quite similar. Following the same procedure in the main text, we collect a trajectory in the training environment with the learned policy, and then add distractors (e.g., the background is replaced with moving videos) on it. We then evaluate the policy deviation and representation deviation across 100 episodes (the distractors are different in different episodes). This is repeated with 5 different random seeds. The comparison results on \texttt{cartpole-swingup} is depicted in Figure \ref{fig:barplotcartpoleswingup}, and one can find that the results also match our conclusion in the main text, i.e., both the representation deviation and policy deviation of DrQ, PAD and CURL are larger than those of SODA, SVEA and PIE-G.

\begin{figure*}[!htb]
    \centering
    \includegraphics[width=0.45\linewidth]{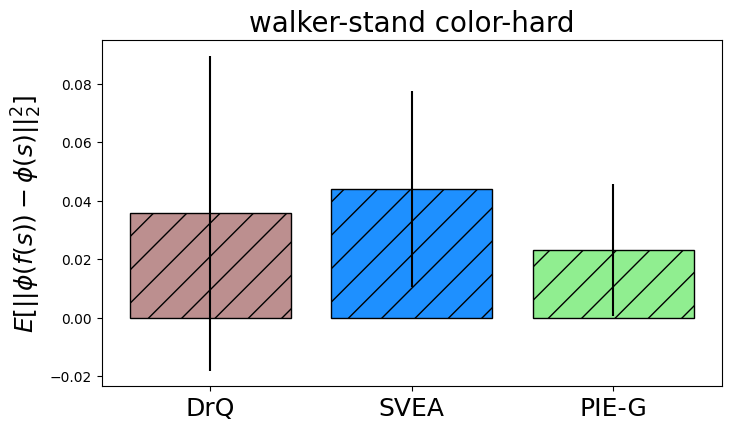}\hspace{-1mm}
    \includegraphics[width=0.45\linewidth]{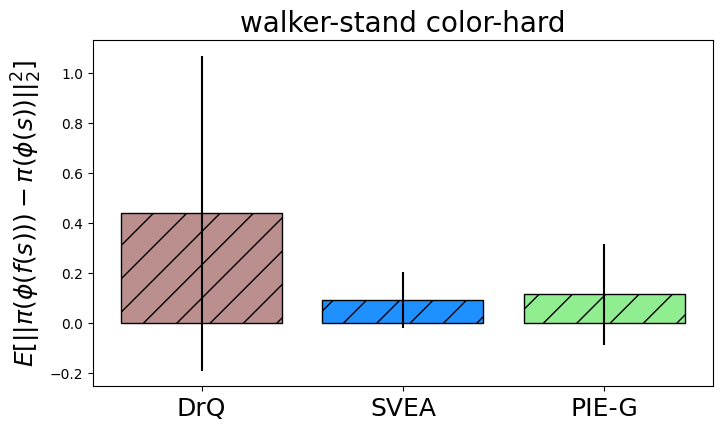}\hspace{-1mm}
    \includegraphics[width=0.45\linewidth]{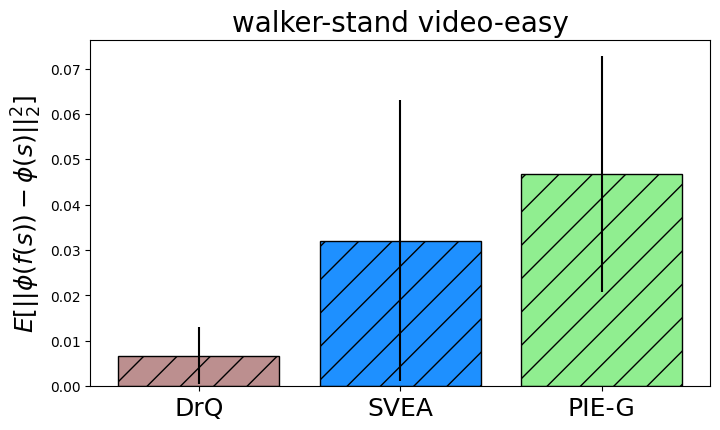}\hspace{-1mm}
    \includegraphics[width=0.45\linewidth]{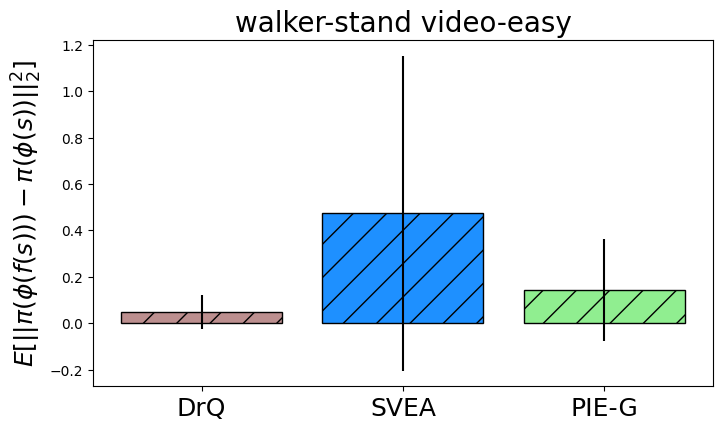}\hspace{-1mm}
    \includegraphics[width=0.45\linewidth]{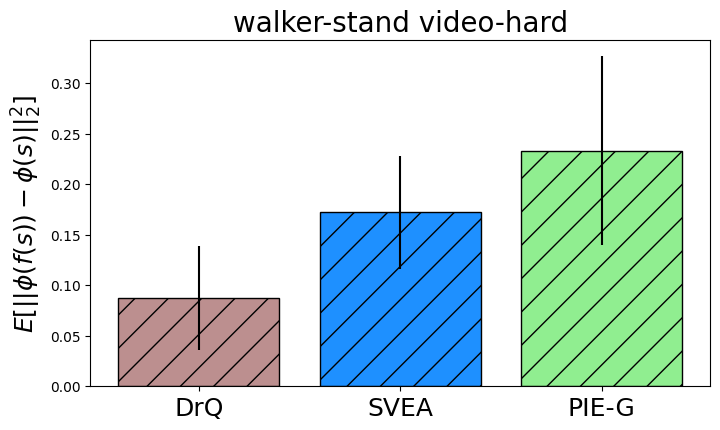}\hspace{-1mm}
    \includegraphics[width=0.45\linewidth]{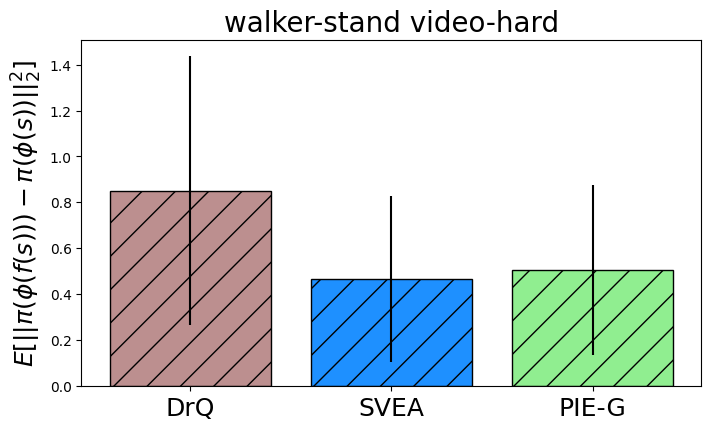}
    \caption{\textbf{Evidence that our theoretical results can explain empirical algorithms.} We present comparison of average representation deviation ($\mathbb{E}[\|\phi(s) - \phi(s^\prime)\|_2^2]$, left column) and average policy deviation ($\mathbb{E}[\|\pi(\phi(s)) - \pi(\phi(s^\prime))\|_2^2]$, right column) of 6 typical methods on color-hard, video-easy, and video-hard settings of walker-stand task from DMC-GB. The results are averaged over the trajectory and across 5 varied random seeds. The error bar denotes the average standard deviation along the trajectory and 5 seeds.}
    \label{fig:barplotwalkerstand}
\end{figure*}

The comparison results \texttt{walker-stand} is shown in Figure \ref{fig:barplotwalkerstand}, where we only compare DrQ against SVEA and PIE-G. We observe that in \texttt{walker-stand} environment, it seems the representation deviation of DrQ is smaller. We would like to clarify here that the reasons lie in the fact that DrQ itself actually already achieves quite good performance on color-hard, video-easy evaluation modes of walker-stand (comparable to SVEA and PIE-G, please refer to \shortcite{Hansen2021StabilizingDQ,Yuan2022PreTrainedIE}), indicating that this task is simple and simple data augmentation techniques utilized in DrQ can aid the capture of important knowledge from the testing environments. This does not indicate that our theory fails. It can be found in Theorem \ref{theo:final} that the generalization gap bound not only depends on the representation deviation $\varrho$, but also correlates with the Lipschitz constant of the policy, i.e., $L_{\pi_1}$. One can see that the policy deviations of DrQ on walker-stand color-hard and video-hard tasks are larger than SVEA and PIE-G, despite that the representation deviations of DrQ are smaller. It is not difficult to conclude that the Lipschitz constant is the largest in DrQ. Similar phenomenon can also be seen in Figure \ref{fig:rationality} where both $\max\frac{\|\pi(\phi(s)) - \pi(\phi(s^\prime))\|_2^2}{\|\phi(s)- \phi(s^\prime)\|_2^2}$ and $\max\frac{\|\pi(\phi(f(s))) - \pi(\phi(f(s^\prime)))\|_2^2}{\|\phi(f(s))- \phi(f(s^\prime))\|_2^2}$ of DrQ are larger than those of PIE-G. For example, on walker-walk test trajectories, the $\max\frac{\|\pi(\phi(f(s))) - \pi(\phi(f(s^\prime)))\|_2^2}{\|\phi(f(s))- \phi(f(s^\prime))\|_2^2}$ gives 24, while PIE-G gives \textbf{7.8}. These further reveal that the policy learned by the DrQ agent is fragile and unstable, while SVEA and PIE-G exhibit better tolerance to distractors during testing on walker-stand color-hard and video-hard tasks. Hence, with a joint influence of the $L_{\pi_1}$ and the representation deviation, the generalization gap of DrQ is larger.

Interestingly, we notice that on the video-easy setting of walker-stand task, DrQ is better than SVEA in terms of both representation deviation and policy deviation. This is because, on this setting, the average test performance of DrQ gives \textbf{873}, while SVEA underperforms it with an average performance of 795. It is also worth noting that one may wonder why on some tasks like walker-walk video-hard, cartpole-swingup video-easy, the representation deviation difference between DrQ and PIE-G is small, and on walker-stand video-hard, PIE-G has a larger representation deviation. This is because the encoders of the two algorithms are trained on \textit{different} domains, leading to different parameters and representation space. It is then understandable that on some tasks, PIE-G has a similar representation deviation as DrQ. Therefore, a better measurement on the generalization gap may be the policy deviation, which is a joint measurement of $L_{\pi_1}\varrho$, and it turns out that PIE-G is better in most of the cases. Moreover, we emphasize that DrQ often has a larger standard deviation on many tasks than SVEA and PIE-G in terms of both policy deviation and representation deviation. Such instability and error can be accumulated during executing actions in the testing environments, and finally result in a bad performance. PIE-G and SVEA, instead, can learn more robust policies and encoders.

To summarize, some special cases (e.g., walker-stand color-hard, walker-stand video-hard) showing that the representation deviation of DrQ is smaller while its generalization gap is larger do not make our theory a failure. As when comparing the generalization gap of different algorithms, \emph{it is more meaningful to compare policy deviation because different algorithms have different encoders and update rules for policy}, and the generalization gap is jointly determined by the smoothness of the policy and the representation deviation. DrQ has poor test performance as it learns an unstable policy (with a large Lipschitz constant). However, by adopting a better data augmentation method and modifying the way of updating $Q$ function, SVEA (which is built upon DrQ) learns a smooth policy and incurs a better generalization ability, which aligns with our theory. Furthermore, the success of PIE-G also strongly supports our theory, since it only replaces the encoder in the vanilla DrQ-v2 with the pre-trained image encoder. We also note that enabling generalizable representation (and smaller representation deviation) across different evaluation scenarios and achieving a smoother policy seems to be strongly connected in practice. Therefore, we believe our claims and insights hold.

\section{Missing Results on Deviations along the Trajectory}
\label{sec:appmissingalongthetrajectory}
\begin{figure*}[!htb]
    \centering
    \includegraphics[width=0.24\linewidth]{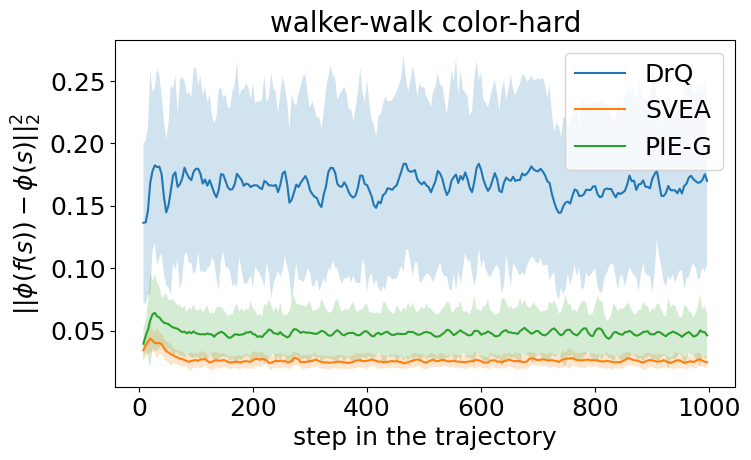}\hspace{-1mm}
    \includegraphics[width=0.24\linewidth]{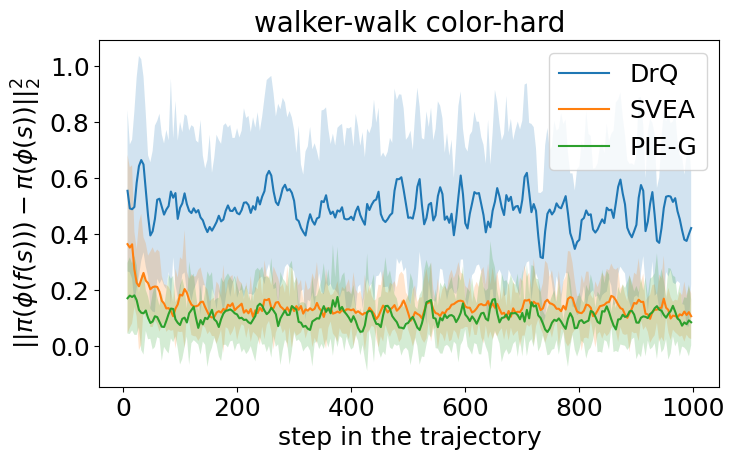}\hspace{-1mm}
    \includegraphics[width=0.24\linewidth]{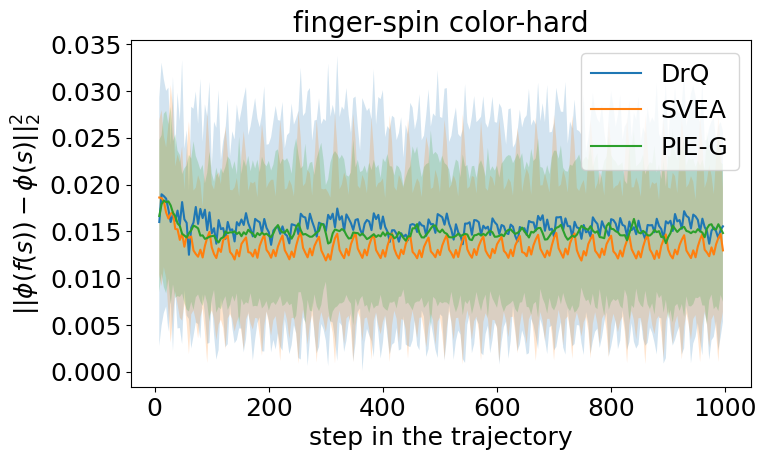}\hspace{-1mm}
    \includegraphics[width=0.24\linewidth]{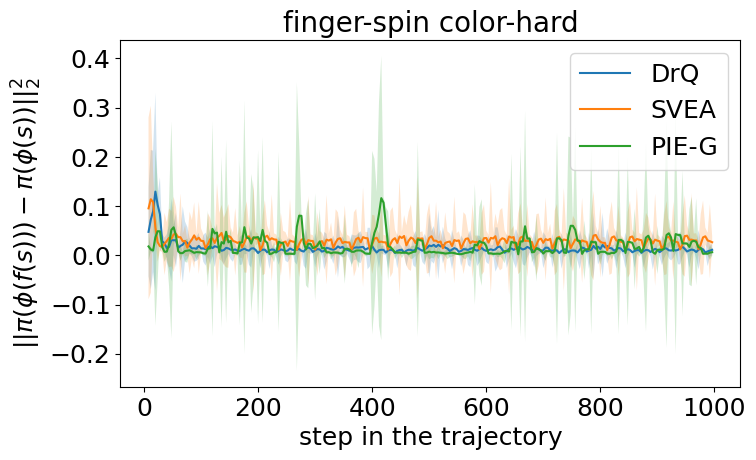}\hspace{-1mm}
    \includegraphics[width=0.24\linewidth]{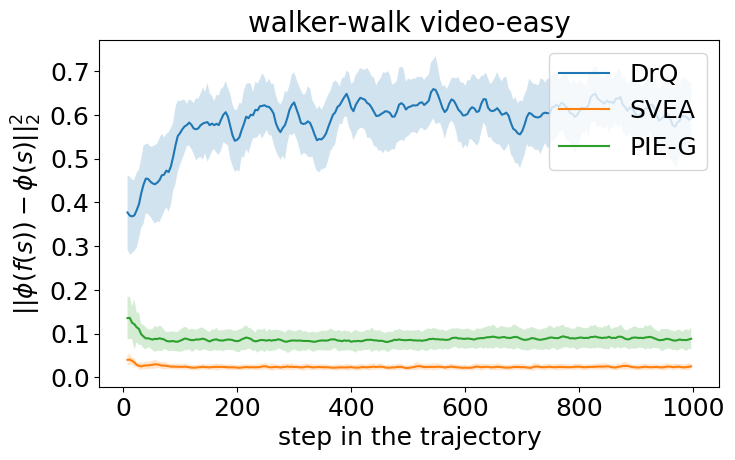}\hspace{-1mm}
    \includegraphics[width=0.24\linewidth]{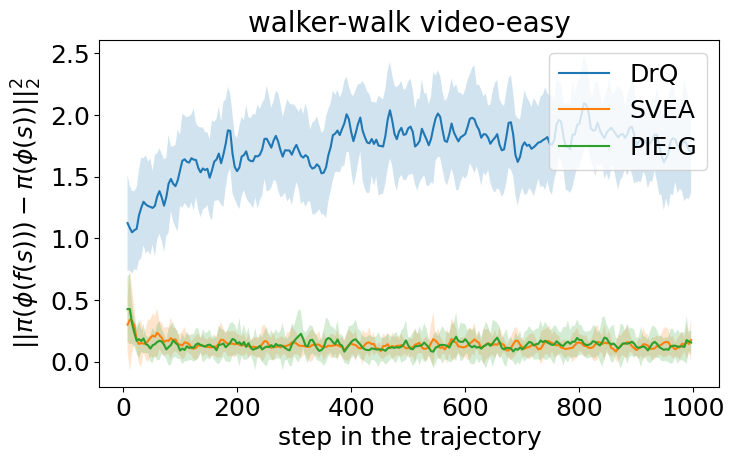}\hspace{-1mm}
    \includegraphics[width=0.24\linewidth]{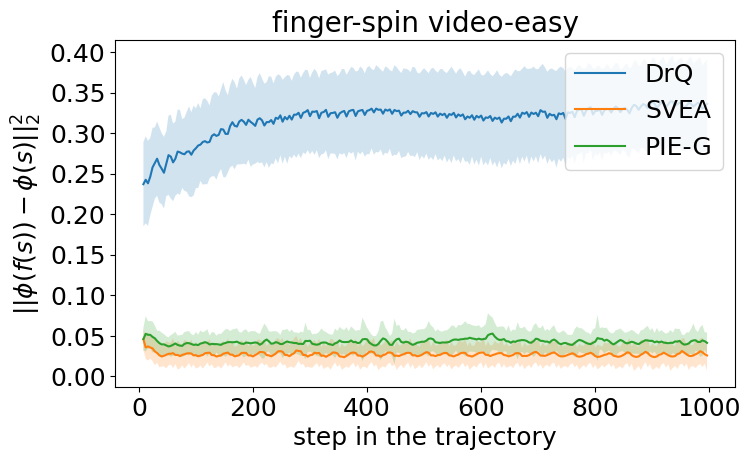}\hspace{-1mm}
    \includegraphics[width=0.24\linewidth]{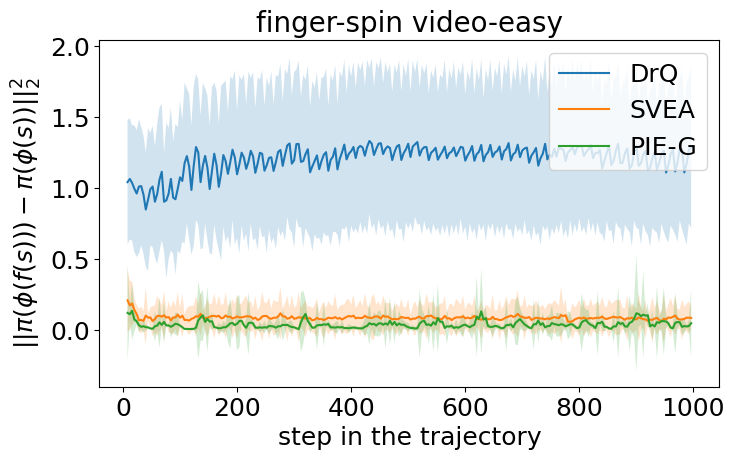}\hspace{-1mm}
    \includegraphics[width=0.24\linewidth]{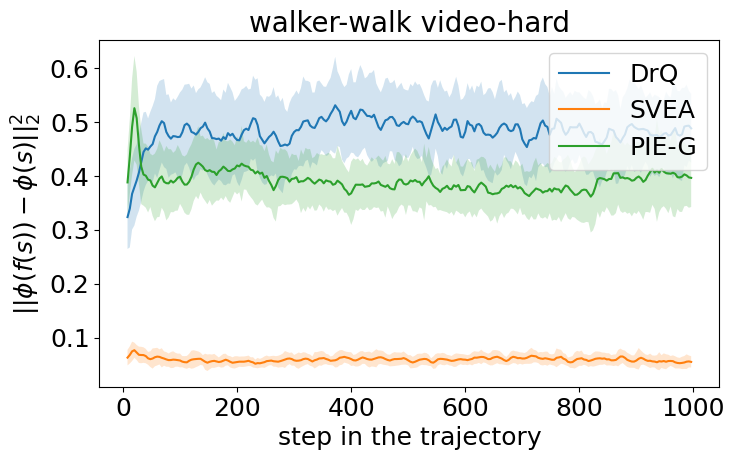}\hspace{-1mm}
    \includegraphics[width=0.24\linewidth]{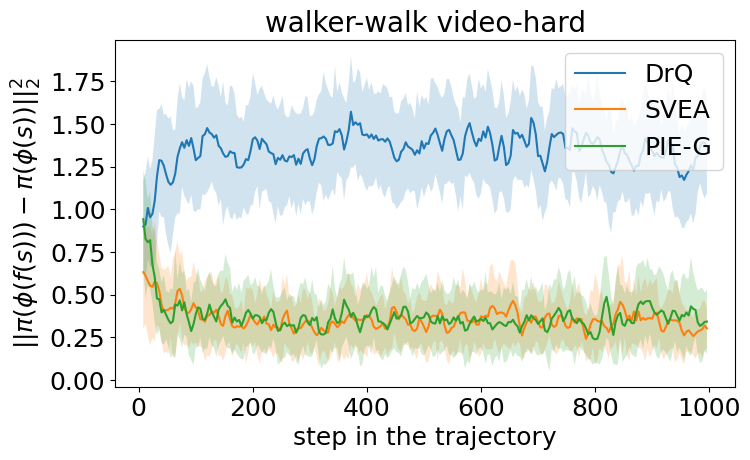}\hspace{-1mm}
    \includegraphics[width=0.24\linewidth]{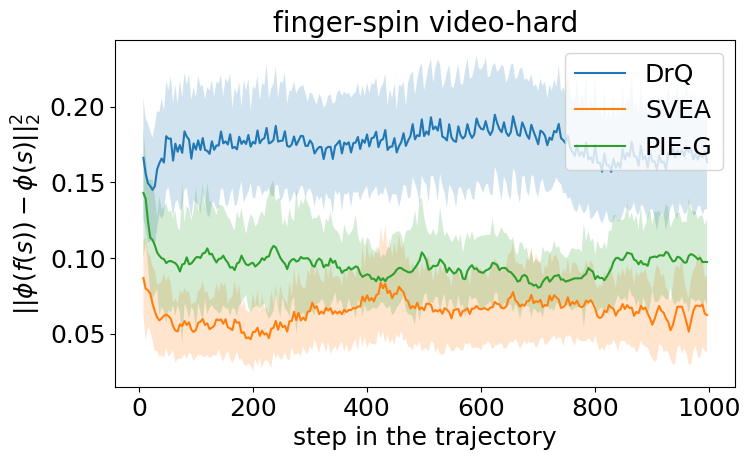}\hspace{-1mm}
    \includegraphics[width=0.24\linewidth]{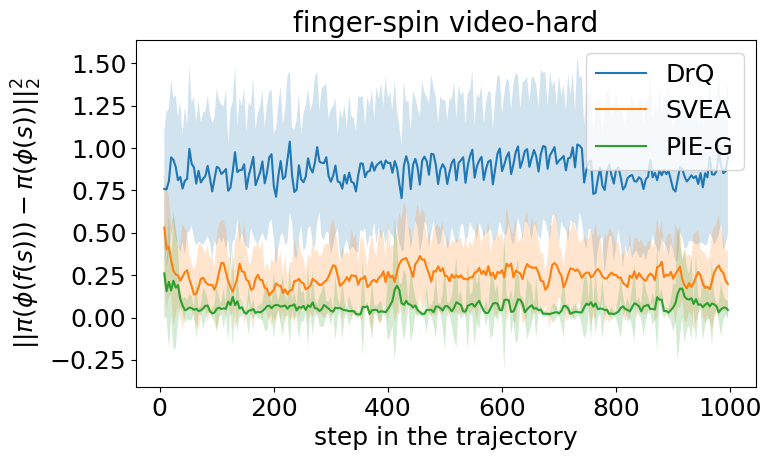}
    \caption{\textbf{Detailed plots of representation deviation and policy deviation comparison.} We show the comparison between SVEA, PIE-G, and DrQ on color-hard, video-easy, and video-hard settings of walker-walk and finger-spin tasks from DMC-GB. The results are averaged over 5 varied random seeds.}
    \label{fig:empiricalevidence}
\end{figure*}

\begin{figure*}[!h]
  \centering
  \includegraphics[width=0.24\linewidth]{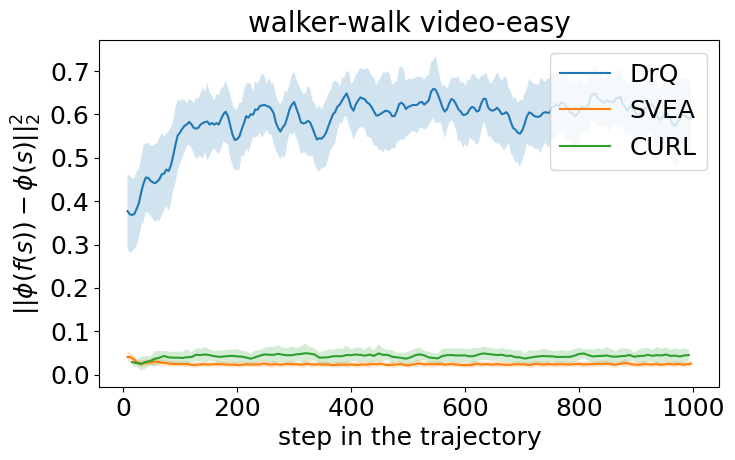}
  \includegraphics[width=0.24\linewidth]{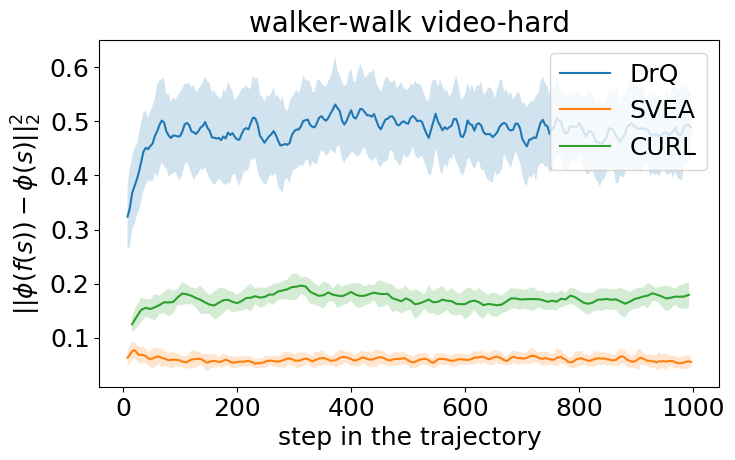}
  \includegraphics[width=0.24\linewidth]{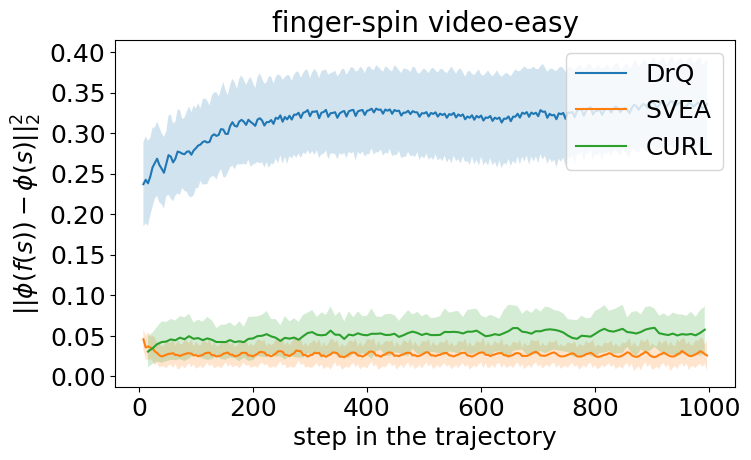}
  \includegraphics[width=0.24\linewidth]{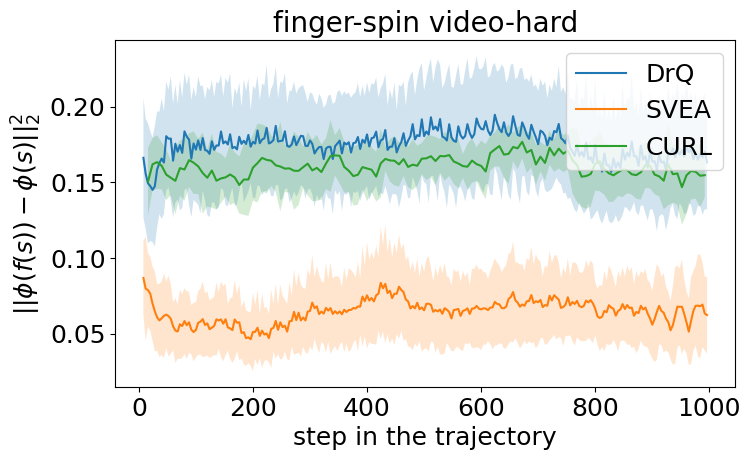}
  \includegraphics[width=0.24\linewidth]{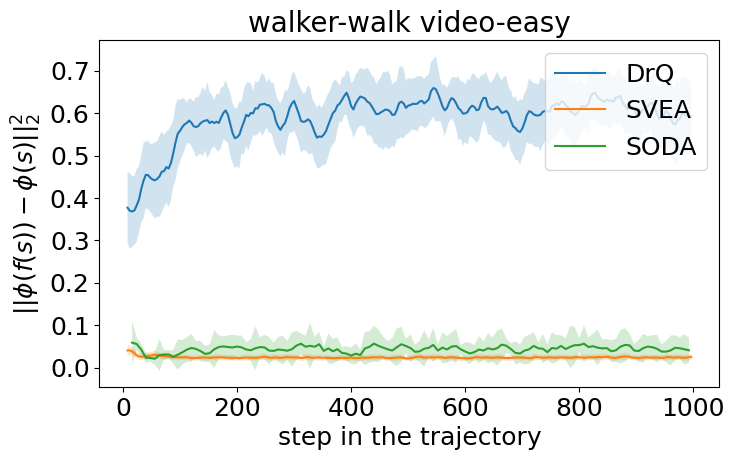}
  \includegraphics[width=0.24\linewidth]{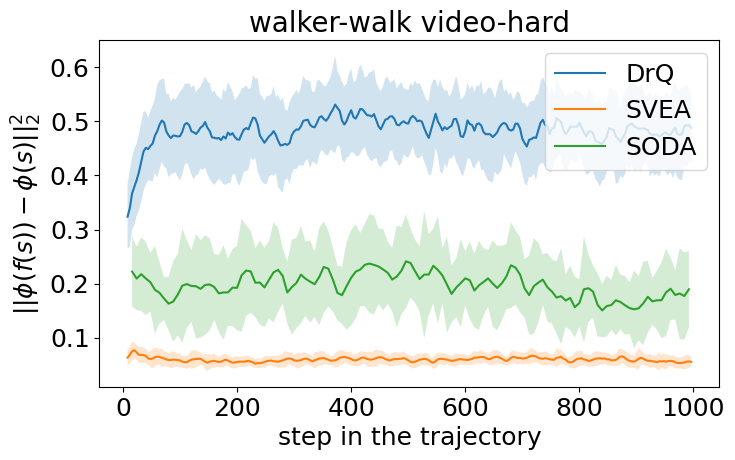}
  \includegraphics[width=0.24\linewidth]{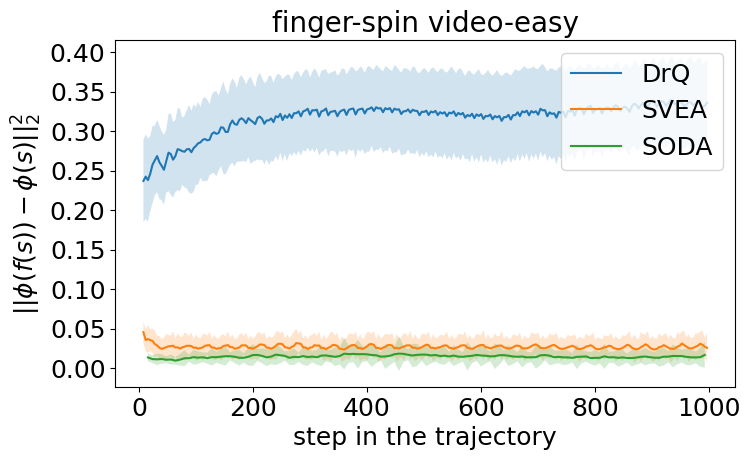}
  \includegraphics[width=0.24\linewidth]{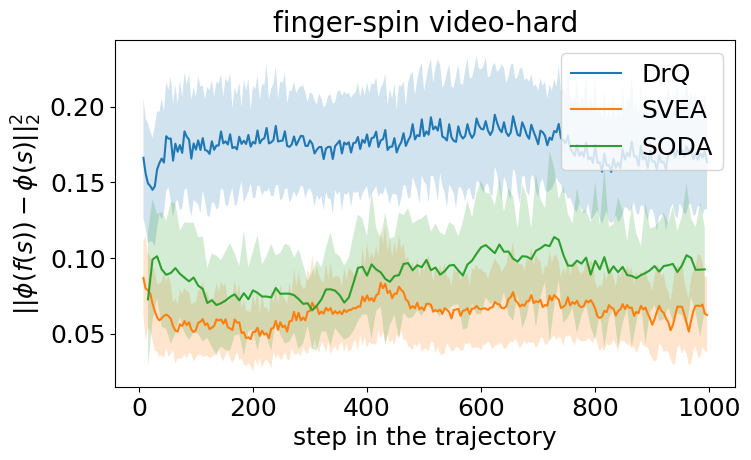}
  \caption{\textbf{Detailed plots of representation deviation comparison.} We show the comparison between DrQ, CURL, SODA, and SVEA on video-easy and video-hard settings of walker-walk and finger-spin tasks from DMC-GB. The results are averaged over 5 varied random seeds.}
  \label{fig:curlsodarep}
\end{figure*}

\begin{figure*}[!h]
  \centering
  \includegraphics[width=0.24\linewidth]{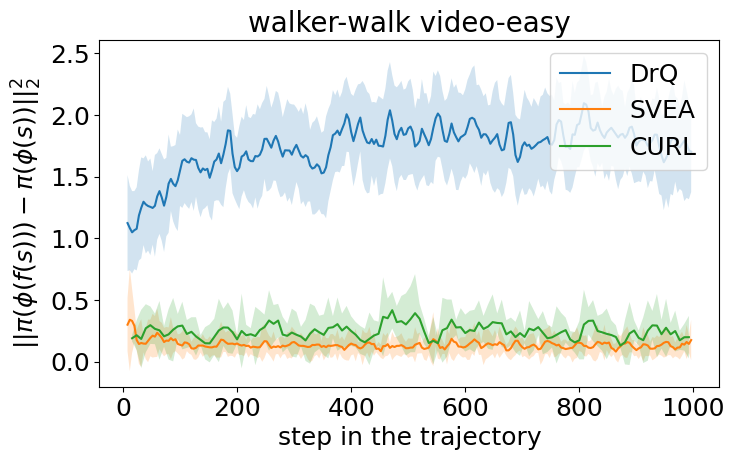}
  \includegraphics[width=0.24\linewidth]{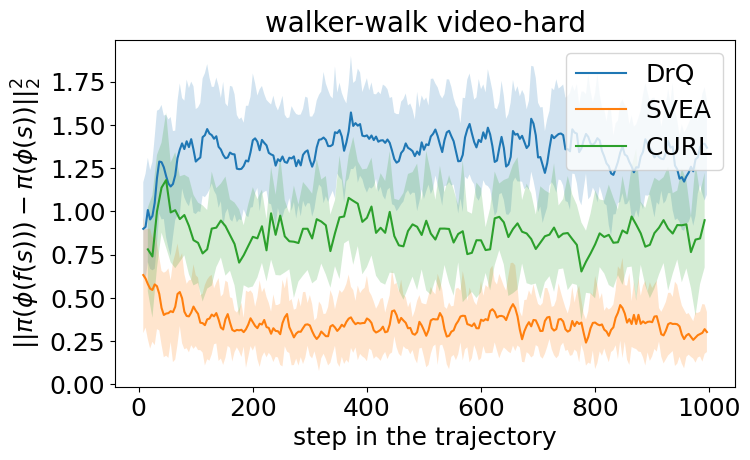}
  \includegraphics[width=0.24\linewidth]{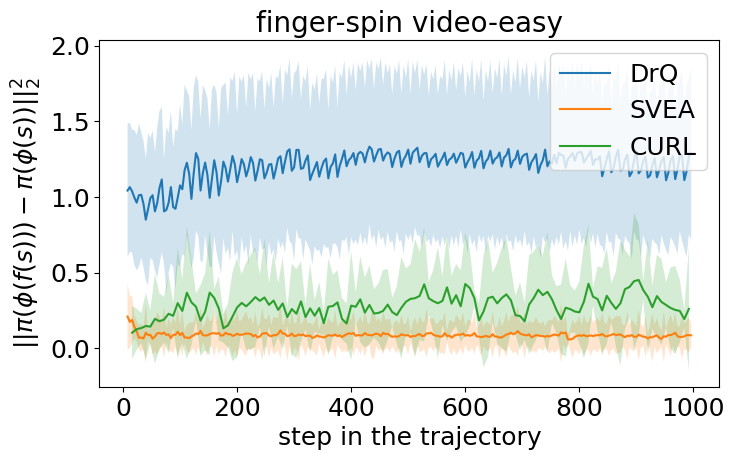}
  \includegraphics[width=0.24\linewidth]{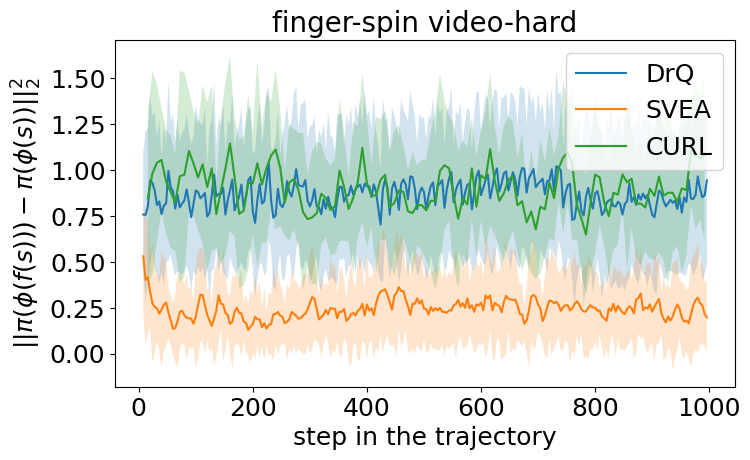}
  \includegraphics[width=0.24\linewidth]{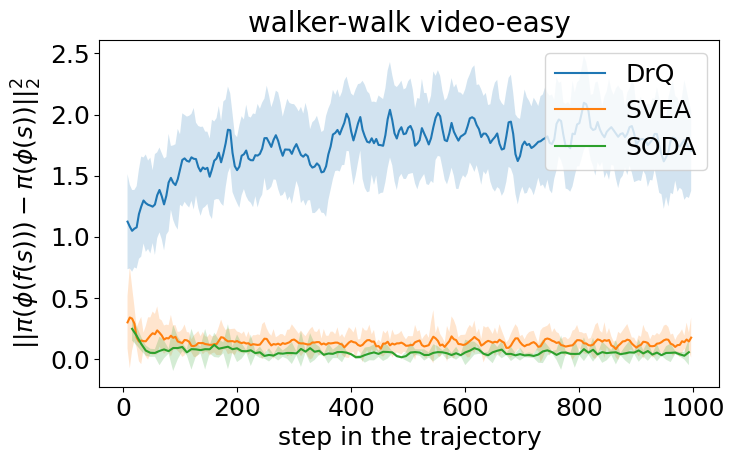}
  \includegraphics[width=0.24\linewidth]{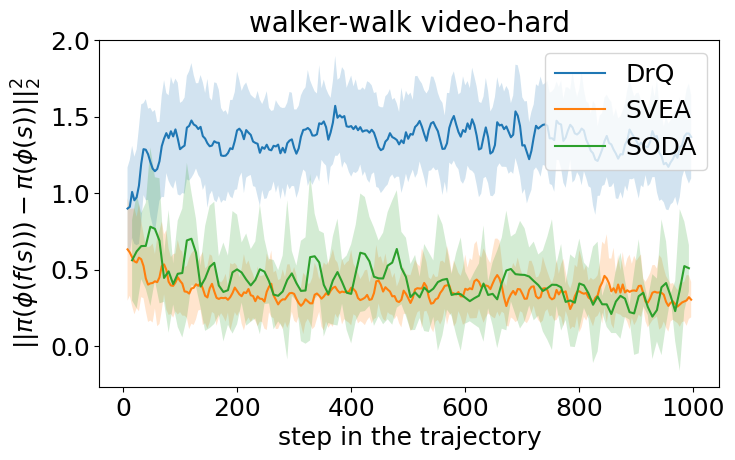}
  \includegraphics[width=0.24\linewidth]{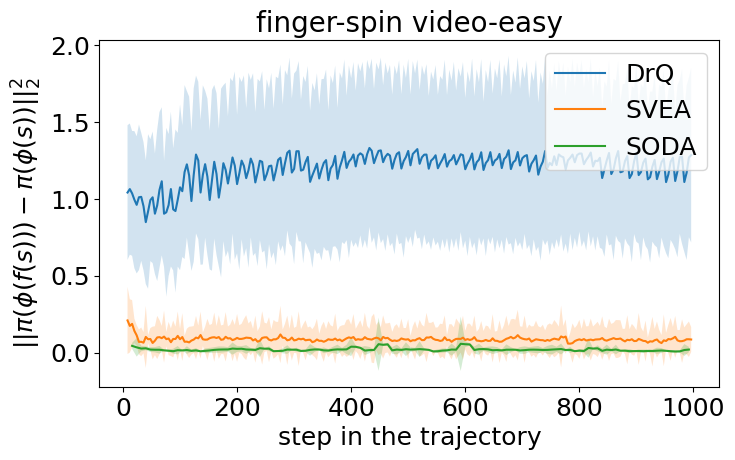}
  \includegraphics[width=0.24\linewidth]{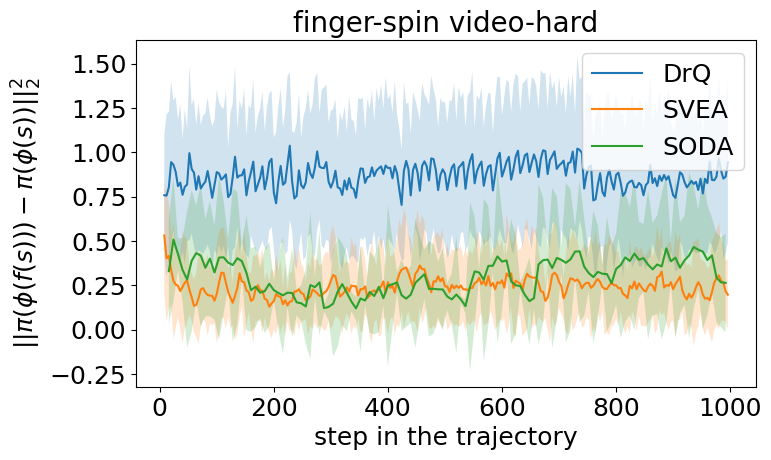}
  \caption{\textbf{Detailed plots of policy deviation comparison.} We show the comparison between DrQ, CURL, SODA, and SVEA on video-easy and video-hard settings of walker-walk and finger-spin tasks from DMC-GB. The results are averaged over 5 varied random seeds.}
  \label{fig:curlsodapolicy}
\end{figure*}

We summarize the comparison results of representation deviation and policy deviation along the evaluation trajectories between DrQ, SVEA and PIE-G in Figure \ref{fig:empiricalevidence}, and the representation deviation comparison between DrQ, CURL, SODA and SVEA in Figure \ref{fig:curlsodarep} and policy deviation comparison in Figure \ref{fig:curlsodapolicy}. The comparison is conducted on two DMC-GB environments, walker-walk and finger-spin.

\section{Compute Infrastructure}

In Table \ref{tab:computing}, we list the compute infrastructure that we use to run all of the algorithms.

\begin{table}[htb]
\caption{Compute infrastructure.}
\label{tab:computing}
\centering
\begin{tabular}{c|c|c}
\toprule
\textbf{CPU}  & \textbf{GPU} & \textbf{Memory} \\
\midrule
AMD EPYC 7452  & RTX3090$\times$8 & 288GB \\
\bottomrule
\end{tabular}
\end{table}


\bibliographystyle{theapa}
\bibliography{sample}

\end{document}